%% file: main.tex
\documentclass[10pt,twocolumn,letterpaper]{article}

 \usepackage[pagenumbers]{iccv} % To force page numbers, e.g. for an arXiv version

\input{preamble}

\usepackage[accsupp]{axessibility}

\definecolor{iccvblue}{rgb}{0.21,0.49,0.74}
\usepackage[pagebackref,breaklinks,colorlinks,allcolors=iccvblue]{hyperref}

\def\paperID{27} % *** Enter the Paper ID here
\def\confName{ICCV}
\def\confYear{2025}

\title{Do It Yourself: Learning Semantic Correspondence from Pseudo-Labels}

\author{Olaf D\"unkel\textsuperscript{1}, Thomas Wimmer\textsuperscript{1,2}, Christian Theobalt\textsuperscript{1,4}, Christian Rupprecht\textsuperscript{3}, Adam Kortylewski\textsuperscript{1,5}\\
{\normalsize \textsuperscript{1}Max Planck Institute for Informatics, Saarland Informatics Campus \quad \textsuperscript{2}ETH Zurich  \quad \textsuperscript{3}University of Oxford} \\ 
{\normalsize\textsuperscript{4}Saarbrücken Research Center for Visual Computing, Interaction and AI  \quad \textsuperscript{5}University of Freiburg}\\
{\small \url{https://genintel.github.io/DIY-SC}}
}

\begin{document}
\maketitle
\input{sec/0_abstract}    
\input{sec/1_intro}
\input{sec/2_related_work}
\input{sec/3_method}

\input{sec/4_experiments}

\input{sec/5_conclusion}

%\clearpage

{
    \small
    \bibliographystyle{ieeenat_fullname}
    \bibliography{main}
}
\clearpage
\input{sec/x_supplementary}

\end{document}

%% file: preamble.tex
\makeatletter
\renewcommand{\paragraph}{%
  \@startsection{paragraph}{4}%
  {\z@}{0.25em}{-1em}%
  {\normalfont\normalsize\bfseries}%
}
\makeatother

\usepackage{comment}
\usepackage[fixed]{fontawesome5} % category icons

\usepackage{graphicx}
\usepackage{tikz}
\usepackage[normalem]{ulem} 

\newcommand{\norm}[1]{\left\lVert#1\right\rVert}

\newcommand{\best}[1]{\textbf{#1}}
\newcommand{\second}[1]{\underline{#1}}

\newcommand{\improvement}[1]{\dashuline{#1}}

\usepackage{colortbl}
\newcommand{\myrowcolour}{\rowcolor[gray]{0.925}}

\newcommand{\oursTab}{{Ours}}
\newcommand{\oursAbbrv}{{DIY-SC}}

\newcommand{\Cow}[1][]{\includegraphics[width=10pt,trim={6cm 9cm 5cm 6cm},clip]{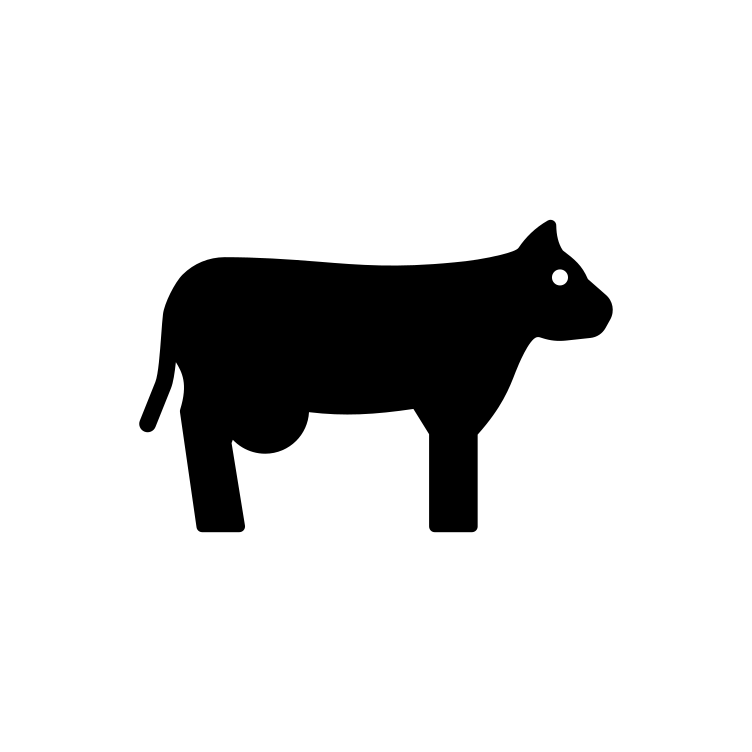}}
\newcommand{\Plant}[1][]{\includegraphics[width=10pt,trim={7cm 6cm 5cm 2cm},clip]{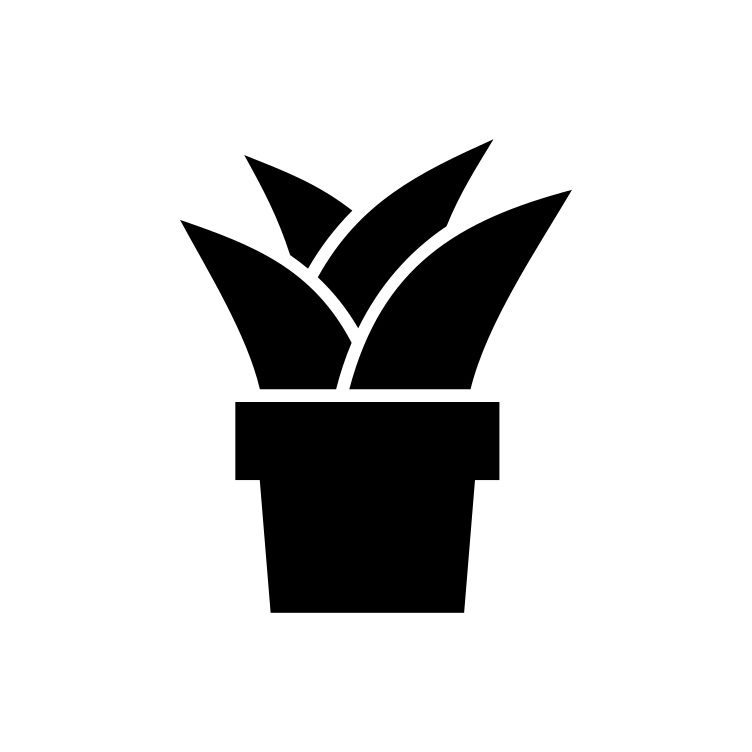}}
\newcommand{\Sheep}[1][]{\includegraphics[width=10pt,trim={6cm 7cm 5cm 6cm},clip]{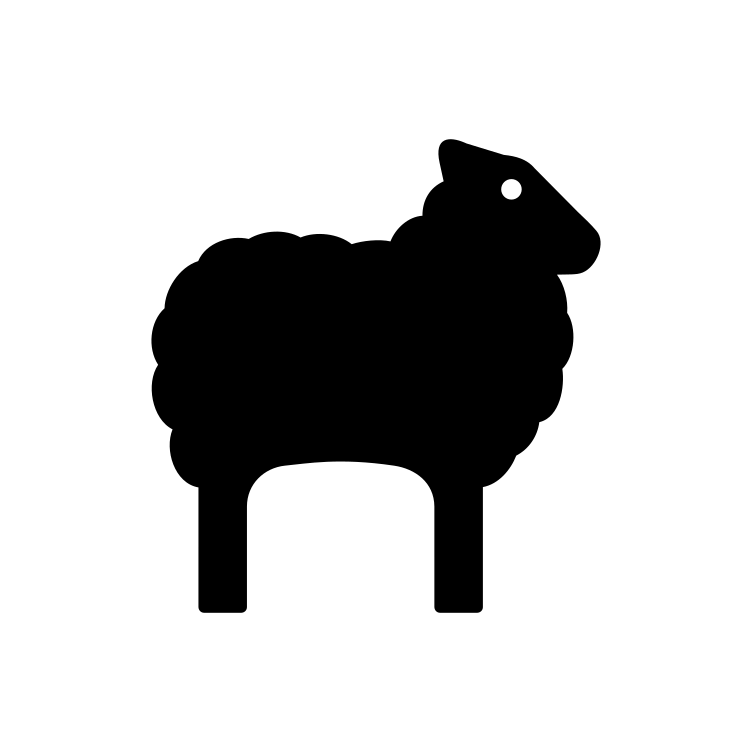}}

%% file: sec/0_abstract.tex
\begin{abstract}
Finding correspondences between semantically similar points across images and object instances is one of the everlasting challenges in computer vision. While large pre-trained vision models have recently been demonstrated as effective priors for semantic matching, they still suffer from ambiguities for symmetric objects or repeated object parts. We propose improving semantic correspondence estimation through 3D-aware pseudo-labeling. Specifically, we train an adapter to refine off-the-shelf features using pseudo-labels obtained via 3D-aware chaining, filtering wrong labels through relaxed cyclic consistency, and 3D spherical prototype mapping constraints. While reducing the need for dataset-specific annotations compared to prior work, we establish a new state-of-the-art on SPair-71k, achieving an absolute gain of over 4\% and of over 7\% compared to methods with similar supervision requirements. The generality of our proposed approach simplifies the extension of training to other data sources, which we demonstrate in our experiments.  
\end{abstract}

%% file: sec/1_intro.tex
\section{Introduction}
\label{sec:intro}

Finding correspondences between images remains a fundamental task in computer vision, having various applications in tracking \citep{doersch2022tap, gao2022aiatrack,karaev2024cotracker3}, mapping and localization \citep{mur2015orb, kokkinos2021point}, affordance understanding \citep{lai2021functional}, pose estimation \citep{xu2022rnnpose}, analysis of characteristic object parts \citep{thai20243, huang2022neural}, image and video generation or editing \citep{hacohen2011non, ofri2023neural, wang2025cove}, 3D representation learning \cite{xiang2024structured,sommer2025common3d,liu2025stable}, or style transfer~\citep{kim2019semantic}.
One remaining challenge in the correspondence estimation field is the task of finding matches across different instances of similar objects, \ie, finding semantic correspondences.
This is a highly semantic task with a certain amount of ambiguity, especially for man-made objects.

Recently, foundation model features have demonstrated surprisingly high zero-shot performance for this task~\citep{tang2023emergent,zhang2023tale,zhang2024telling}.
However, these features still have various weaknesses for finding correspondences, such as ambiguities for similar object parts~\citep{mariotti2024improving} or symmetric objects~\citep{zhang2024telling}. % noisiness (?).
In addition, supervised methods~\citep{zhang2024telling} indicate that there are more effective strategies to combine foundational features for solving semantic correspondence than simple concatenation~\cite{zhang2023tale} or weighted averaging~\cite{mariotti2024improving}. 
Recent works \cite{luo2023diffusion,zhang2024telling,xue2025matcha} train models in a supervised manner using manual keypoint annotations and achieve strong performance when evaluated on the same dataset.
However, as the required manual annotations are scarce and difficult to obtain, this strategy is not scalable to larger, more diverse datasets. 
It thus remains an open challenge to determine the best way to extract and enhance knowledge encoded in foundation models for finding semantic correspondences \textit{without} relying on labor-intensive keypoint supervision.

\begin{figure}[t]
    \centering
    \includegraphics[width=\linewidth]{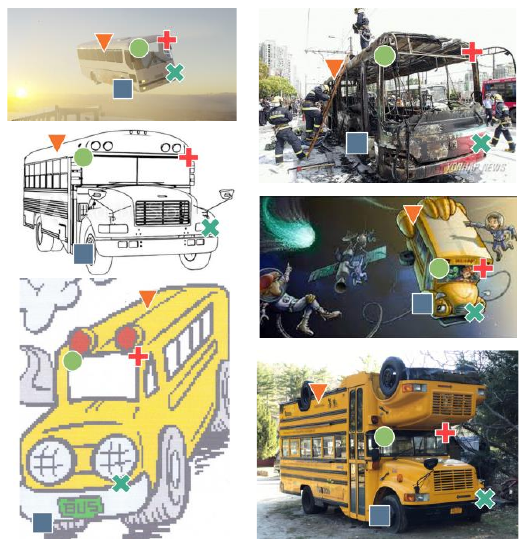}
    \caption{Our method, DIY-SC, is able to find semantic correspondences even for extreme appearance and shape changes. All matches are computed from the keypoints in the top-left image.}
    \label{fig:teaser}
    \vspace{-0.5cm}
\end{figure}

Recent works reduce the necessary level of supervision for learning semantic correspondences by regularizing with 3D information~\citep{mariotti2024improving} or injecting information about label definitions~\citep{zhang2024telling, fundel2024distillation}.
However, they either require careful tuning of weighting factors~\citep{mariotti2024improving} or rely on access to dataset-specific labeling conventions~\citep{zhang2024telling}.

\begin{figure}
    \centering
    \includegraphics[width=.8\columnwidth]{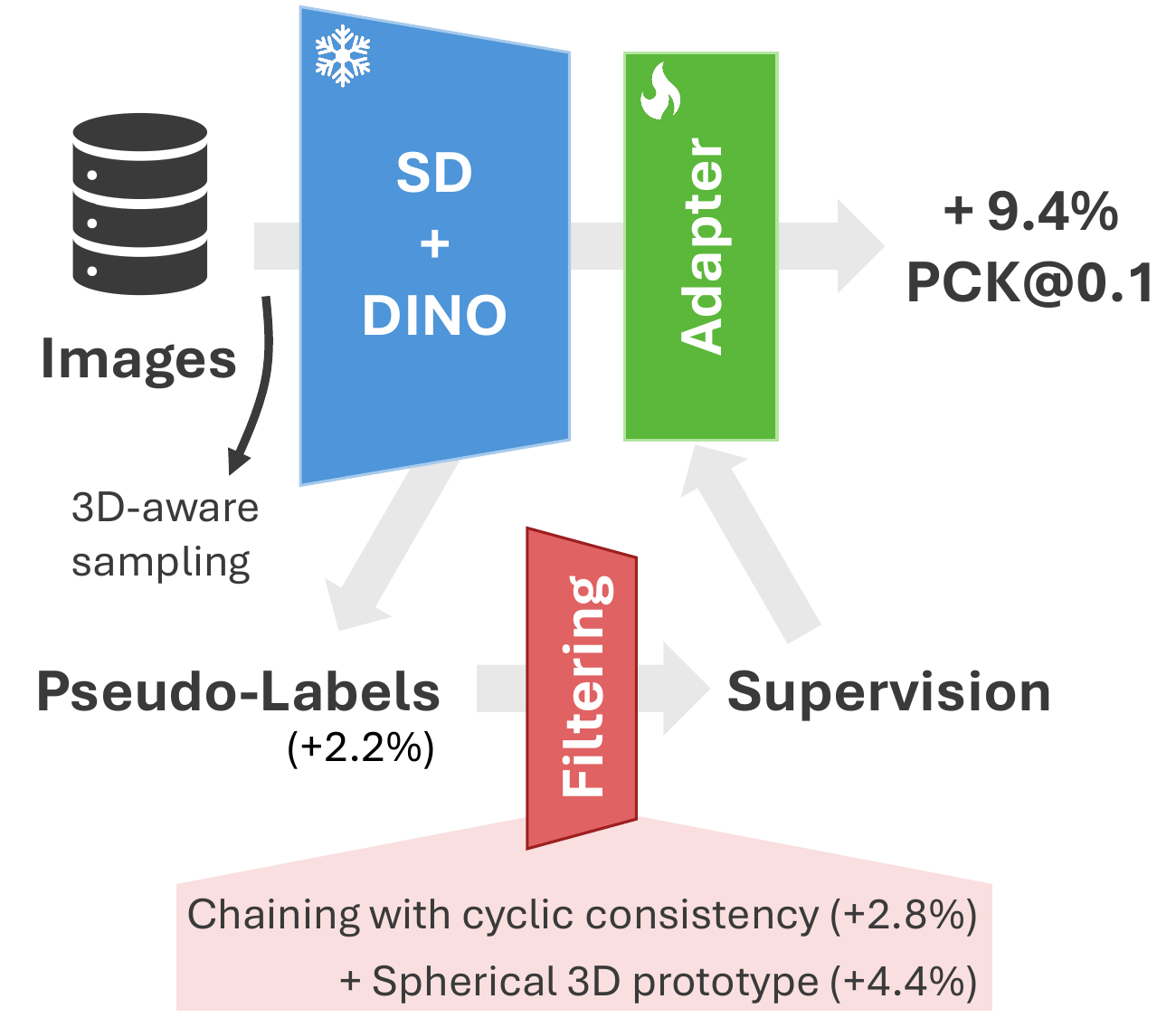}
    \caption{\textbf{Self-training using pseudo-labels.} Using foundational features, we generate pseudo-labels, which are subsequently filtered and used as supervision signals to train a light-weight adapter that refines the features for semantic matching.}
    \label{fig:method-overview-teaser}
    \vspace{-.5cm}
\end{figure}

We address both limitations in our work by learning features that significantly outperform the previous SOTA without requiring manual keypoint label definitions:
Our work demonstrates that training with self-generated labels, \ie,~pseudo-labels, where label quality is improved through weak 3D-aware supervision, is surprisingly effective for improving the semantic correspondence performance, also generalizing to heavy appearance changes (\cref{fig:teaser}).
To achieve this, we train a lightweight adapter supervised by the pseudo-labels, which effectively refines the foundation model's features, as shown in \cref{fig:method-overview-teaser}. 
Specifically, we pursue a zero-shot approach~\citep{zhang2023tale} for generating pseudo-labels for image pairs exhibiting moderate viewpoint variation, a strategy that has been demonstrated to be effective in such contexts but fails for larger viewpoint variations~\citep{zhang2024telling}.
We compose these labels over multiple image pairs to acquire higher label quality for harder correspondence pairs, \ie, with larger viewpoint variance.
We then employ a spherical prototype as a weak geometric prior~\citep{mariotti2024improving} to reject wrong matches, which addresses the inherent challenge of \textit{matchability} and reduces feature ambiguity.

In contrast to previous weakly supervised methods~\citep{zhang2024telling, fundel2024distillation}, our approach does not rely on dataset-specific keypoint definitions, reducing the barriers to applying it to other datasets.
To demonstrate this, we also train a model on the recent ImageNet-3D dataset \cite{ma_imagenet3d_2024}.
Pre-training on this larger dataset notably also improves the performance on SPair-71k~\citep{min2019spair}, demonstrating the effectiveness of our more generalizable approach.

To summarize our contributions:
\begin{itemize}
    \item We show that pseudo-labeling is effective for learning better semantic correspondence features, and we demonstrate that the quality of pseudo-labels can be improved through 3D-aware chaining, relaxed cyclic-consistency constraints, and the integration of a weak geometric prior.
    \item We show that our strategy is scalable to a larger dataset, further improving the model's performance.
    \item Finally, we set a new SOTA on SPair-71k with weak supervision, outperforming the previous best model by 4.5 absolute points.
\end{itemize} % TODO

%% file: sec/2_related_work.tex
\section{Related Work}
\label{sec:related_work}

\textbf{Semantic correspondence.}
Semantic matching, \ie., finding correspondences between \textit{different} instances of the same object class, is more challenging than geometric matching due to potential variations in appearance and shape. The scarcity and ambiguity of manually annotated data pose an extra challenge to learning-based methods for this task~\citep{truong2021warp,zhang2025semantic}.

While early works relied on hand-crafted descriptors~\citep{lowe2004distinctive, liu2010sift}, deep learning enabled learning better-suited feature extractors~\citep{yi2016lift, kim2017dctm, novotny2017anchornet,rocco2018end} and direct semantic correspondence detection networks~\citep{rocco2017convolutional, han2017scnet, kim2019semantic}.
Due to the limited ground truth data, techniques emerged that rely on weak supervision  \citep{lan2021discobox,chen2020show,zhang2023self,truong2022probabilistic}, leverage warp supervision and cycle consistency losses~\citep{zhou2016learning,truong2021warp, truong2022probabilistic}, or rely on increasing label supervision with pseudo-labels~\citep{kim2022semi,li2021probabilistic,huang2023weakly}.

With the advent of so-called foundation models, the semantic matching literature has come full circle: Recent studies demonstrated that features obtained from such models can be utilized for detecting semantic correspondences in a zero-shot manner~\citep{aberman2018neural, amir2021deep, zhang2023tale, hedlin2023unsupervised, tang2023emergent, cheng2024zero}.
More specifically, DINO features~\citep{caron2021emerging, oquab2023dinov2} have been shown to exhibit strong coarse semantic awareness useful for handling larger, cross-instance variations~\citep{amir2021deep, zhang2023tale, zhang2024telling, suri2024lift, fundel2024distillation, xue2025matcha}, while features from diffusion models~\citep{rombach2022high,stracke2024cleandift} can complement them, resulting in better performance~\citep{hedlin2023unsupervised, tang2023emergent, zhang2023tale, zhang2024telling, mariotti2024improving, li2024sd4match, fundel2024distillation, xue2025matcha}.

While a simple nearest-neighbor search in feature space proves to be a powerful zero-shot method for correspondence estimation, prior work also uncovered systematic problems of this approach, \eg, in the disambiguation of symmetric object parts~\citep{luo2023diffusion, zhang2024telling, mariotti2024improving, li2024sd4match, wimmer2024back,sommer2025common3d}. Therefore, several works have proposed appending adapter modules that are fine-tuned using supervision with ground-truth correspondences~\citep{zhang2024telling, xue2025matcha}.
While reducing the labeling requirements, methods that build a joint atlas for objects in multiple images usually require larger quantities of input images and perform test-time optimization~\citep{gupta2023asic, ofri2023neural}. 
\citet{zhang2024telling} use keypoint-specific information to disambiguate left-right symmetries at test time. 
While their strategy is simple, not all symmetries can be solved by image transformations, e.g., flipping, and the required keypoint-specific information is generally not available.
\citet{fundel2024distillation} proposed fine-tuning distilled foundational features for geometric matching using 3D ground truth data.
This approach is, however, inherently limited to the same object instance, resulting in suboptimal results for cross-instance matching.  
\citet{mariotti2024improving} proposed learning a map from features to the surface of a sphere using 3D pose information as a weak supervision signal.
While this method successfully resolves most symmetries for simple objects, such as cars, a spherical prior does not accurately capture objects with more complex topologies.

Our work also aims to improve the performance of semantic matching by refining foundational features.
In contrast to \citet{zhang2024telling}, we do not require dataset-specific keypoint definitions since our work only relies on weak supervision that is available at scale.
Similar to \citet{mariotti2024improving}, we leverage 3D information to reduce ambiguities. 
However, our work does not suffer from performance drops for objects with more complex topology and thus does not require tuning of weighting factors

\textbf{Pseudo-labeling.}
\citet{lee2013pseudo} proposed pseudo-labeling, also referred to as self-training, as a technique for semi-supervised training of neural networks, where a small initial set of labels is propagated to a larger set of unlabeled instances.
Various methods have been developed based on this concept, demonstrating the generality and scalability of this approach \citep{asano2019self, xie2020self, caron2021emerging, karaev2024cotracker3, hamilton2022unsupervised, attaiki2022ncp}, sometimes in conjunction with knowledge distillation from other pre-trained models. 
% \citep{gupta2023asic} proposed to use sparse pseudo-labels obtained from DINO in the context of semantic matching. 
% Their proposed method refines these correspondences in a test-time optimization to align multiple images to common atlas.
Previous works have expanded manual keypoint annotations and demonstrated improved semantic correspondence performance \cite{kim2022semi,li2021probabilistic,huang2023weakly}.
In our work, we propose to extract pseudo-labels using off-the-shelf foundational features (at training time), which we use to fine-tune a light-weight adapter for the task of semantic correspondence estimation. 

%% file: sec/3_method.tex
\section{Method}
\label{sec:method}
\begin{figure*}[t]
  \centering
   \includegraphics[width=\linewidth]{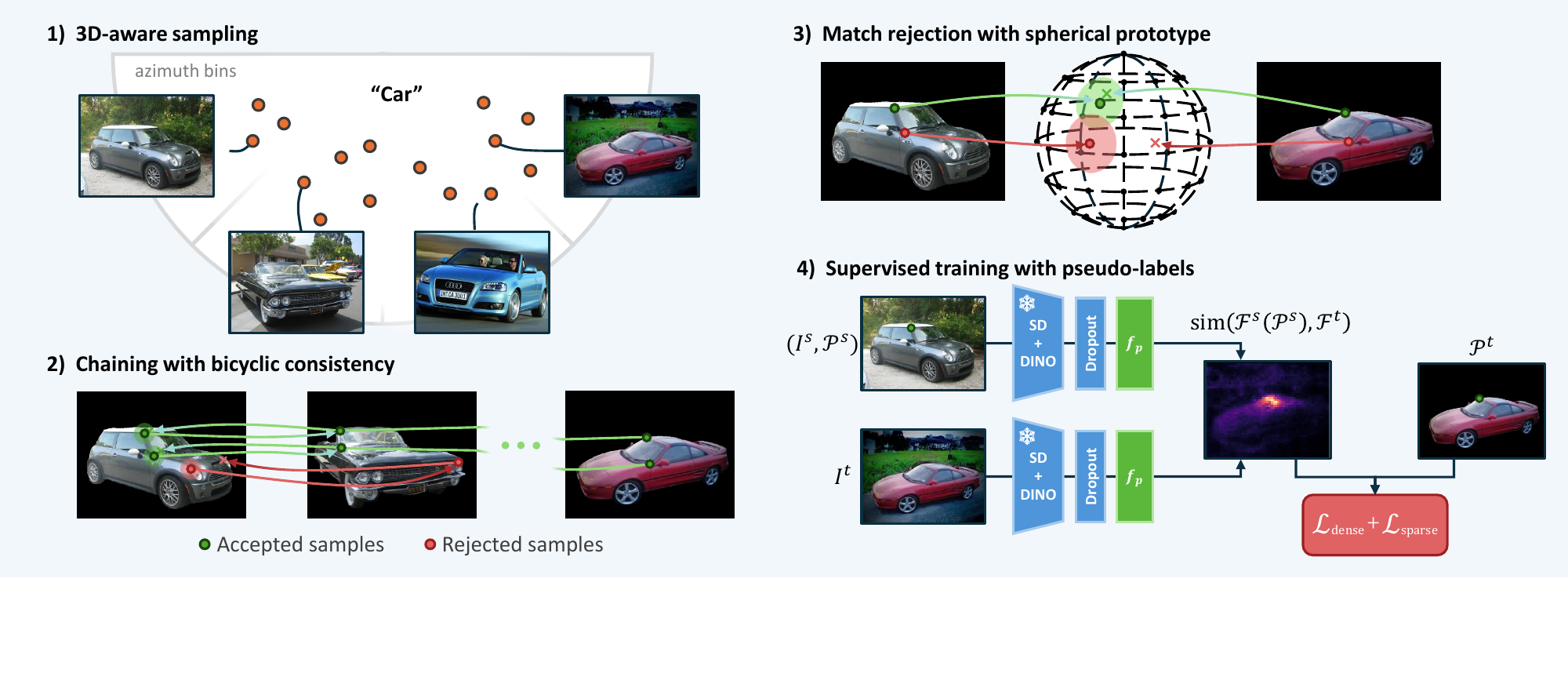}
   \vspace{-1.7cm}
   \caption{\textbf{Method overview.} We use azimuth information to sample image pairs for which higher zero-shot performance can be expected (1). We then chain the pairwise predictions to get correspondences for larger viewpoint changes, where we reject matches that do not fulfill a relaxed cyclic consistency constraint (2). We further filter pseudo-labels by rejecting pairs that can not be mapped to a similar location on a 3D spherical prototype (3). Finally, we use the resulting pseudo-labels to train an adapter $f_p$ in a supervised manner (4).}
   \label{fig:method}
   \vspace{-.2cm}
\end{figure*}

Foundational features show strong semantic awareness for zero-shot matching across instances.
However, while the high feature similarity of similar object parts is reasonable from a semantic perspective, distinguishing them is crucial for accurate semantic correspondences. 
The performance gap between supervised and unsupervised semantic matching methods \cite{zhang2024telling} shows that a learned combination and refinement of features outperforms a simple concatenation or averaging of multiple foundational features.
As we aim to supervise a refinement stage without relying on human annotations, we draw on the effectiveness of pseudo-labeling for other tasks~\citep{karaev2024cotracker3,kage2024review,attaiki2022ncp} and hypothesize its potential for improving features for semantic correspondence.

In this section, we first formalize the semantic correspondence task (\cref{sec:met:problem}).
Then, we introduce our two-stage method:
First, pseudo-labels are generated via a zero-shot matching strategy and subsequently filtered (\cref{sec:met:pseudo-labels}).
These pseudo-labels are then used to supervise the training of a light-weight adapter $f_p$ (\cref{sec:met:adapter}).
Finally, we detail how our method can be scaled to a larger dataset in \cref{sec:met:scaling}. 

\subsection{Problem Definition} \label{sec:met:problem}
The problem of finding semantic correspondences is defined as follows:
Given a source image $I^s$, a target image $I^t$, and a query point $p_i^s \in \mathbb{R}^2$ on the object in the source image, the task is to find the target point $p_i^t \in \mathbb{R}^2$  on the target image that localizes the object part that semantically corresponds to the queried object part.
One strategy to solve this task is to find the nearest neighbor in feature space via the cosine similarity $\text{sim}(\cdot)$ of the feature maps $\mathcal{F}^\text{s}$ and $\mathcal{F}^\text{t}$:
\begin{equation} \label{eq:nn_feat}
    p_i^t =\text{NN}^{s\rightarrow t}(p_i^s)=\arg\max_{q_i^t \in I^t} \text{sim}\left( \mathcal{F}^s(p_i^s),\mathcal{F}^t(q_i^t \right)).
\end{equation}
However, in contrast to zero-shot strategies that solely apply pre-trained feature extractors~\citep{zhang2023tale}, we aim to improve the features such that the refined features are better suited for finding semantic correspondences.
Since the features $\mathcal{F}^\text{DINO}$~\citep{oquab2023dinov2} and $ \mathcal{F}^\text{SD}$~\citep{rombach2022high} generalize well due to their large training corpus, we learn a light-weight adapter $\mathcal{F}=f_\text{p}(\tilde{\mathcal{F}})$ on top of the concatenated features $\tilde{\mathcal{F}}=\left[\mathcal{F}^\text{DINO},\mathcal{F}^\text{SD}\right]$, as proposed by \citet{zhang2024telling}.

\subsection{Generation of High-Quality Pseudo-Labels} \label{sec:met:pseudo-labels}
Given a pair of images $(I^s,I^t)$ with their feature maps $(\mathcal{F}^s,\mathcal{F}^t)$ and the instance masks $(\mathcal{M}^s,\mathcal{M}^t)$, the goal is to find matches $(p_i^s, p_i^t)$ that can be used as pseudo-labels for supervising the training of the adapter $f_p(\cdot)$.
A na\"ive strategy would apply the nearest-neighbor (NN) computation (\cref{eq:nn_feat}) of all points on the mask:
\begin{equation}
    p^t_i=\text{NN}^{s\rightarrow t}(p^s_i)
\end{equation}
with $p_i^s\in\mathcal{M}^s$ and $p_i^t \in \mathcal{M}^t$.
However, due to viewpoint variations, topological variations, or occlusions, not all points on the mask $\mathcal{M}^s$ can be mapped to the target image. This challenge is commonly referred to as the \textit{matchability} or \textit{visibility} problem \cite{truong2021warp}.
We show that training an adapter with these na\"ively obtained pseudo-labels already improves results (see \ref{tab:ablations}).
However, as self-training heavily relies on the quality of the pseudo-labels, we aim to improve their quality by rejecting wrong matches.
Pseudo-labels should be removed if no matching is possible (\textit{matchability}) or if points are matched wrongly because of erroneous zero-shot correspondences, \eg,~because of \textit{feature ambiguity} of object parts with a similar appearance.

Our work addresses the challenge of removing wrong labels in the following ways:
(1) We perform 3D-aware sampling of image pairs along a chain $(I^s,I^t)$ where image pairs have similar viewpoints and more accurate pseudo-labels are expected.
(2) We enforce cyclic consistency for the pseudo-labels along the chain to reduce the number of spurious matches.
(3) We reject wrong pseudo-labels by integrating a canonicalized spherical 3D prior.

\begin{table}
  \centering
  \resizebox{\columnwidth}{!}{
  \begin{tabular}{@{}l|@{\hskip 0.2cm}c@{\hskip 0.2cm}c@{\hskip 0.2cm}c@{\hskip 0.2cm}c@{\hskip 0.2cm}c@{}}
    \toprule
    $\Delta v$ & $[0^\circ, 45^\circ)$&$(0^\circ,90^\circ)$&$(45^\circ, 135^\circ)$&$(90^\circ, 180^\circ)$&$(135^\circ, 225^\circ)$ \\
    \midrule
    PCK@0.1 & 75.9& 68.1&  57.2& 54.0& 52.6\\
    \bottomrule
  \end{tabular}
  }
  \caption{\textbf{Zero-shot performance (SD+DINO) degrades for larger viewpoint changes.}
  The SPair-71k dataset has $8$ azimuth bins with an equidistant size of $45^{\circ}$. We compute the ratio of correctly corresponded keypoints (PCK@0.1 \textit{per-img}) for increasing viewpoint changes averaged over all categories.
  % \textcolor{blue}{Class specific is in suppl.}
  }
  \label{tab:analysis_delta_vp}
  \vspace{-.3cm}
\end{table}

\paragraph{(1) 3D-aware image pair sampling.}
The zero-shot approach, which utilizes features from Stable Diffusion \cite{rombach2022high} and DINOv2 \cite{oquab2023dinov2}, performs well for objects that do not undergo significant changes in their topology and appearance, and are presented from a similar viewpoint. However, the performance clearly degrades for larger viewpoint changes, as studied by \citet{zhang2024telling}. %where perfomance drop is 10p for SD+DINO.
We study the sensitivity of the semantic correspondence performance for varying viewpoint differences using the azimuth angle annotations of the object-centric semantic correspondence dataset SPair-71k \cite{min2019spair} in \cref{tab:analysis_delta_vp}. 
We find that the zero-shot performance decreases for larger viewpoint changes, \ie, if the viewpoint's azimuth angle $\phi$ varies by more than $45^{\circ}$.

This observation motivates the generation of pseudo-labels for image pairs with a similar viewpoint.
However, more challenging image pairs where the viewpoint heavily deviates are also required to learn a strong point matcher.

To account for this, we propose propagating matches through a $K$-tuple $(I_1,...,I_K)$, where each pair $(I_k,I_{k+1})$ represents an image pair of the same category with a small but non-zero viewpoint variation (\cref{fig:method}, (1)), such that $d_\text{circ}(\phi_k,\phi_{k+1})<90^{\circ}$ with the spherical distance $d_\text{circ}(\cdot)$, which is equivalent to one azimuth bin in the SPair-71K dataset. 
The matches $\mathcal{K}^k=\{ (p^k_i , p^{k+1}_i ): p^k_i\in \mathcal{P}^{k}\subseteq \mathcal{M}^k,p^{k+1}_i\in \mathcal{P}^{k+1}\subseteq \mathcal{M}^{k+1}\}$
%\begin{equation}
%\mathcal{K}^k=\{ (p^k_i , p^{k+1}_i ): p^k_i\in \mathcal{P}^{k}\subseteq \mathcal{M}^k,p^{k+1}_i\in \mathcal{P}^{k+1}\subseteq \mathcal{M}^{k+1}\}
%\end{equation}
are propagated via recursive application of the NN operator for the subsequent pairs:
\begin{align}
        \mathcal{P}^{k+1}&=\text{NN}^{k\rightarrow k+1}(\mathcal{P}^{k})\\
                &=\text{NN}^{k\rightarrow k+1}(\text{NN}^{k-1\rightarrow k}(\mathcal{P}^{k-1}))=\cdots \notag\\
                \mathcal{P}^1 &= \mathcal{M}^1.
\end{align}
Through this strategy, each considered image pair is expected to have a higher pseudo-label quality than the na\"ive strategy, where pseudo-labels are generated for tuples with potentially very different poses $(I_1, I_k)$. 

\paragraph{(2) Cyclic consistency of propagated pseudo-labels.}
While the aforementioned strategy can improve the quality of matches, it does not necessarily prevent spurious matches that can occur due to feature ambiguities, \eg, because of similar appearance.
In order to reduce such spurious matches or matching of incompatible regions, we apply a cyclic consistency constraint for the matched points, as proposed by \citet{aberman2018neural}.
A match $(p^s_i, p^t_i)$ is cyclic consistent if the following holds:
\begin{equation}
    \text{NN}^{s \rightarrow t}(p_i^s)=p^t_i \quad \text{and} \quad \text{NN}^{t \rightarrow s}(p^t_i)=p^s_i,
    % p^s_i = \text{NN}^{t \rightarrow s}\left(\underbrace{\text{NN}^{s \rightarrow t}(p^s_i)}_{=p^t_i}\right)
\end{equation}
with the nearest neighbor operator $\text{NN}$ in feature space, as introduced in \cref{eq:nn_feat}.

We observe that this cyclic consistency constraint rejects many keypoint tuples $(p_i^s, p_i^t)$, as the zero-shot matching approach has difficulties mapping back to the exact location of $p_i^s$. 
For this reason, we relax the cyclic consistency constraint, where we do not enforce exact consistency but allow a small deviation from the source point:
\begin{equation}\label{eq:soft_cyclic_constraint}
\begin{split}
    p^t_i=\text{NN}^{s\rightarrow t}(p^s_i),  \quad  \hat{p}^s_i = \text{NN}^{t\rightarrow s}(p^t_i), \\ 
    \text{and} \quad \norm{\hat{p}^s_i-p^s_i}_2<r_\text{max}
    \end{split}
\end{equation}
with the rejection radius $r_{max}$, see \cref{fig:method} (2).
We iteratively apply this relaxed cyclic consistency constraint to all matches $\mathcal{K}^k$, which filters out incompatible pairs and keeps matches capturing concepts that could be propagated through the set of $K$ images
We choose $K=4$ in our experiments since this considers the full range of viewpoint variations of $180^{\circ}=4\cdot 45^{\circ}$, given that the azimuth annotations for SPair-71k are only coarse bins of $45^{\circ}$.

\paragraph{(3) Rejection of wrong pseudo-labels with a canonicalized spherical 3D prior.}
Although our chaining strategy removes spurious matches, wrong matches might still occur, in particular for cases of left-/right-ambiguity and repeated object parts, as, \eg, illustrated in \cref{fig:method}, where simple nearest-neighbor search on foundational features fails.

Therefore, we aim to specifically remove such wrong matches by leveraging a canonicalized object-centric spherical 3D prior that captures objects in an aligned coordinate system.
\citet{mariotti2024improving} propose a \textit{spherical mapper} that maps DINOv2 feature patches $\text{x}_i^\text{DINO}=\mathcal{F}^\text{DINO}(p_i)$ to points $\psi_i\in\mathcal{S}^2$ on a canonicalized sphere $\psi_i=f_s(\text{x}^\text{DINO}_i)$ by using only coarse view point information and instance masks during training.
The key advantage of this strategy is that it naturally assigns object parts to regions on a spherical prototype across different instances and categories, which allows rejecting wrong samples in case they are in different areas of the sphere, \eg, when considering visually similar wheels on the left or the right side of a car.

\citet{mariotti2024improving} demonstrate that this approach significantly improves the performance on SPair-71k when combining their spherical features with the concatenated features of SD and DINO via a weighted average.
However, as this approach modulates the original feature similarity, the performance deteriorates for categories that are not well represented by a sphere, \eg, non-rigid categories.
In contrast, our approach uses the spherical mapper only for removing potentially wrong pseudo-labels. That filter should have a high true positive rate while still keeping a sufficiently high true negative rate: Supervising with few correct matches is more desirable than supervising with many matches that are partially systematically wrong, as we show in our experiments.
For this purpose, we compute the spherical points $\mathcal{P}^s$ and $\mathcal{P}^t$ for all matches:
\begin{equation}
    \Psi^s=f_s(\mathcal{F}^\text{DINO}(\mathcal{P}^s)) \quad \text{and} \quad \Psi^t=f_s(\mathcal{F}^\text{DINO}(\mathcal{P}^t)).
\end{equation} 
Then, we reject all matches $(p^s_i,p^t_i)$ where
\[
\text{sim}(\psi ^s_i ,\psi ^t_i) < \theta_\text{th}
\]
with the threshold $\theta_\text{th}<0.15\cdot\pi$ that is selected to allow disambiguating left and right and potentially repeated object parts, such as the wheels of a car.
While this thresholding may also remove correct matches for classes that are not well-represented by a rigid object prototype, we observe that the number of matches is still sufficiently large to serve as a dense supervision signal.
Most importantly, only removing pseudo-labels does not modulate the original zero-shot matches.
This is in contrast to regularizing with spherical mapper features, which deteriorates the localization accuracy of the original foundational features \citep{mariotti2024improving}.

\subsection{Supervised Training with Pseudo-Labels} \label{sec:met:adapter}

After having generated the pseudo-labels, we train the adapter $f_p(\cdot)$ using supervised training with the following two losses:
First, we train a CLIP \cite{radford2021learning}-inspired sparse contrastive loss as proposed by \citet{luo2023diffusion}:
\begin{equation}
    \mathcal{L}_{\text{sparse}} = CL\bigl(\mathcal{F}^s(\mathcal{P}^s), \mathcal{F}^t(\mathcal{P}^t)\bigr).
\end{equation}
This loss maximizes feature similarity for corresponding points while minimizing similarity to non-matching points.
Second, we also supervise with the dense loss
\begin{equation}
\begin{split}
    \mathcal{L}_{\text{dense}} = \sum \bigl\|\hat{p}_i^t - \bigl(p_i^t + \epsilon\bigr)\bigr\|_2  \quad \text{with}  \\
    \hat{p}_i^t=\text{WindowSoftArgmax}( \mathcal{F}^s\bigl(p_i^s\bigr)^\top \mathcal{F}^t),
    \end{split}
\end{equation}
with a window soft-argmax \citep{zhang2024telling,kim2019semantic} and Gaussian noise samples $\epsilon$.
The dense loss is particularly important for us since it propagates the gradient also to areas in the feature map without labels, \ie, unmatched areas.
This learned combination and refinement of foundational features is a decisive improvement over previously proposed simple concatenation~\citep{zhang2023tale, zhang2024telling} or weighted averaging of feature similarities of a weak geometric regularizer~\citep{mariotti2024improving}.

\begin{table*}[t]
    \centering
    \resizebox{\textwidth}{!}{
    \begin{tabular}{lc|rrrrrrrrrrrrrrrrrr|r}
            & & \faIcon{plane} & \faIcon{bicycle} & \faIcon{crow} & \faIcon{ship} & \faIcon{wine-bottle} & \faIcon{bus} & \faIcon{car} & \faIcon{cat} & \faIcon{chair} & \Cow  & \faIcon{dog} & \faIcon{horse} & \faIcon{motorcycle} & \faIcon{walking} & \Plant & \Sheep & \faIcon{train} & \faIcon{tv} & avg\\ 
        \midrule
        \myrowcolour
        ASIC & \citep{gupta2023asic} &57.9 &25.2 &68.1& 24.7& 35.4 &28.4 &30.9 &54.8 &21.6 &45.0 &47.2& 39.9& 26.2& 48.8 &14.5 &24.5 &49.0 &24.6 &36.9\\
        DINOv2 & \cite{oquab2023dinov2} 
            & 72.7 & 62.0 & 85.2 & 41.3 & 40.4 & 52.3 & 51.5 & 71.1 & 36.2 & 67.1 & 64.6 & 67.6 & 61.0 & 68.2 & 30.7 & 62.0 & 54.3 & 24.2 & 55.6 \\
        \myrowcolour
        DIFT & \citep{tang2023emergent} 
            & 63.5 & 54.5 & 80.8 & 34.5 & 46.2 & 52.7 & 48.3 & 77.7 & 39.0 & 76.0 & 54.9 & 61.3 & 53.3 & 46.0 & 57.8 & 57.1 & 71.1 & 63.4 & 57.7 \\
        SD + DINOv2 & \cite{zhang2023tale} 
            & 73.0 & 64.1 & 86.4 & 40.7 & 52.9 & 55.0 & 53.8 & 78.6 & 45.5 & 77.3 & 64.7 & 69.7 & 63.3 & 69.2 & 58.4 & 67.6 & 66.2 & 53.5 & 64.0 \\
        \myrowcolour
        DistillDIFT* (U.S.) & \citep{fundel2024distillation}
            & 74.6 & 60.4 & 88.7 & 42.5 & 53.5 & 55.0 & 54.6 & 80.8 & 42.7 & 78.6 & 72.0 & 71.4 & 62.2 & 70.7 & 53.1 & 68.6 & 65.2 & 61.6 & 65.1\\
        \midrule
        SphMap$^\dag$ & \citep{mariotti2024improving}
            &75.3 & 63.8 & 87.7 & \second{48.2} & 50.9 & \second{74.9} & \second{71.1} & 81.7 & 47.3 & 81.6 & 66.9 & 73.1 & 65.4 & 61.8 & 55.5 & 70.2 & 75.0 & 58.5 & 67.8 \\
        \myrowcolour
        TLR & \citep{zhang2024telling}
            & \second{78.0} & \second{66.4} & \second{90.2} & 44.5 & \best{60.1} & 66.6 & 60.8 & 82.7 & \best{53.2} & 82.3 & 69.5 & \second{75.1} & \second{66.1} & 71.7 & \second{58.9} & \second{71.6} & \second{83.8} & 55.5 & 69.6 \\
        DistillDIFT* (W.S.) & \citep{fundel2024distillation}
            & \best{78.2} & 63.8 & 90.1 & 45.0 & 54.6 & 68.0 & 63.7 & \second{83.2} & 49.3 & \second{82.6} & \best{74.5} & 73.8 & 63.5 & \second{72.0} & 56.2 & 71.0 & \best{86.2} & \second{66.5} & \second{70.6} \\
        \myrowcolour
         \oursTab\ &
            & 77.2 & \best{69.1} & \best{90.8} & \best{54.2} & \second{57.9} & \best{83.7} & \best{77.5} & \best{86.5} & \second{53.1} & \best{86.7} & \second{73.1} & \best{78.5} & \best{72.5} & \best{74.0} & \best{73.5} & \best{76.0} & 77.2 & \best{69.5} & \best{74.4} \\
        \midrule
        \midrule
        SphMap-$\mathcal{S}^2$ (IN3D) &
            & 74.3 & 60.9 & 82.8 & 49.4 & 50.4 & 76.2 & 73.3 & 71.7 & 47.9 & 73.0 & 53.4 & 68.9 & 69.4 & 50.9 & 34.8 & 54.4 & 62.5 & 57.1 & 61.9\\
        \myrowcolour
        \oursTab\ (IN3D) &
            & 75.9 & 68.7 & 90.1 & 55.2 & 56.1 & 82.9 & 76.5 & 82.7 & 55.4 & 83.6 & 71.2 & 75.2 & 71.3 & 64.3 & 56.2 & 69.8 & 78.6 & 61.2 & 71.2 \\
        \oursTab\ (IN3D $\rightarrow$ SPair) &
            & \improvement{77.6} & \improvement{70.3} & \improvement{91.0} & 53.6 & \improvement{58.7} & \improvement{84.8} & \improvement{80.3} & 86.1 & \improvement{54.1} & \improvement{87.1} & \improvement{74.0} & \improvement{79.2} & 72.0 & \improvement{75.4} & 71.9 & \improvement{76.4} & \improvement{77.4} & \improvement{71.9} & \improvement{75.1} \\
        \bottomrule
    \end{tabular}
    }
     \caption{\textbf{Per-category PCK@0.1 scores (\textit{per-keypoint}) on SPair-71k}. \best{Best} and \second{second best} are highlighted. When pre-training on ImageNet3D~\citep{ma_imagenet3d_2024}, results are \improvement{improved}. $^\dag$SphMap avg is slightly higher than in the original paper where they used macro averaging. We trained using their code. *DistillDIFT was evaluated using their trained checkpoints.
     SphMap and DIY-SC require 3D pose annotations.
     }
    \label{tab:Spair_all_cats}
    \vspace{-.3cm}
\end{table*}

\subsection{Scaling to a Larger Dataset} \label{sec:met:scaling}
The comparatively small size of the SPair-71k dataset, with only 18 categories and 1,800 images in total, motivates scaling to a larger dataset with more diverse classes and a greater number of images.
The spherical mapper~\citep{mariotti2024improving} is potentially applicable to larger object-centric datasets with 3D annotations, but the simple geometric prior limits the improvement when applied via weighted average, as proposed by \citet{mariotti2024improving}.
The geometric-aware flipping used by the current SOTA methods~\citep{fundel2024distillation,zhang2024telling} is not applicable to other datasets since it requires access to the keypoint label definitions, only presented in correspondence datasets.

In contrast, our work relies solely on categories, masks, and 3D object poses as a weak supervision signal, which are available on a larger scale.
Therefore, our method can be scaled to larger datasets, leading to generalizable semantic correspondence matching by training on objects that are not present in the SPair-71k dataset.
We demonstrate this by training our whole pipeline on the recently proposed ImageNet-3D \cite{ma_imagenet3d_2024} dataset, which has around 86k images.
We train the spherical mapper similarly to \citet{mariotti2024improving}, where we use masks acquired from SAM~\citep{ravi2024sam}.
As more accurate 3D pose labels are present in the ImageNet-3D dataset, we do not supervise the viewpoint direction only using the bin of the azimuth angle, as proposed by \citet{mariotti2024improving} for the SPair-71k dataset.
Instead, we extend their formulation by computing the correlation coefficient on the two-sphere $\mathcal{S}^2$ with the pose annotation:
\begin{equation}\label{eq:sphere2_sph_map}
    \mathcal{L}_{\mathcal{S}^2} =\sum_{\mathbf{I},\mathbf{I'}}\norm{ \psi_\mathbf{I}\cdot \psi_\mathbf{I'}  - \mu(f_S(\mathbf{I}))\cdot \mu(f_S(\mathbf{I'}))}^2,
\end{equation}
with $\psi_\mathbf{I}\in\mathcal{S}^2$ and $\psi_\mathbf{I'}\in\mathcal{S}^2$ computed from the 3D pose annotations of the objects in the two images.
This not only supervises the azimuth viewpoint information (yaw) but also the pitch of an object and, therefore, improves canonicalization.
We provide more details in the supplementary.

This strategy enables us to scale our proposed method to more classes and images from a significantly larger dataset, ultimately leading to improved generalization.

%% file: sec/4_experiments.tex
\begin{table}
  \centering
  \resizebox{\columnwidth}{!}{
  \begin{tabular}{lcllcccc}
    \toprule
    &&   \multicolumn{3}{c}{SPair-71k}& \multicolumn{3}{c}{AP-10k (PCK@0.1)}\\
        \cmidrule(lrr){3-5}\cmidrule(llr){6-8}
        Models &&  0.1 & 0.05 & 0.01 &  I.S & C.S. & C.F.\\
    \midrule
    \myrowcolour
    SD + DINOv2 & \citep{zhang2023tale} &  59.9 & 44.7 & 7.9 & 62.9 & 59.3 & 48.3\\
    DistillDIFT* (U.S.) & \citep{fundel2024distillation} &  60.8& 45.4&8.0&  -&-&-\\
    \midrule
    \myrowcolour
    SphMap$^\dag$ & \citep{mariotti2024improving} &  64.4& 48.2&8.4&  65.4&63.1&51.0\\
    TLR & \citep{zhang2024telling} &  65.4& 49.1 &\second{9.9}&  68.7& 64.6& 52.7\\
    %TLR (without flip) & -& -& -& -& -& -&  -&-&-\\
    \myrowcolour
    DistillDIFT* (W.S.) & \citep{fundel2024distillation} &  65.3 & 49.8& 8.9 & -&-&-\\
    \oursTab\  &&    \best{71.6} & \best{53.8} &\best{10.1}& \second{70.6}& \second{69.1}& \second{57.8}\\
    \myrowcolour
    \oursTab\ (DINOv2)  &&    \second{70.6}& \second{51.1}&9.0& \best{71.2}& \best{69.8}& \best{58.3}\\
    \midrule
    \midrule
    %\myrowcolour
    TLR (sup)& \citep{zhang2024telling} &  82.9 & 72.6 & 21.6 & 70.1 & 68.3 & 58.4 \\ % 87.5 & 85.8 & 78.4 \\
    \bottomrule
  \end{tabular}
  }
  \caption{
  \textbf{Results for different PCK levels (\textit{per-image}) on the SPair-71k and the AP-10k dataset.} Reproduced supervised results for the model \cite{zhang2024telling} solely trained on SPair-71k. 
  Results for AP-10K are intra-species (I.S.), cross-species (C.S.), and cross-family (C.F.), as introduced in \cite{zhang2024telling}.
  $^\dag$ and * as in \cref{tab:Spair_all_cats}.
  DINOv2: Only application and refinement of DINOv2 features.
  }
  \label{tab:per-image-pck}
  \vspace{-.6cm}
\end{table}

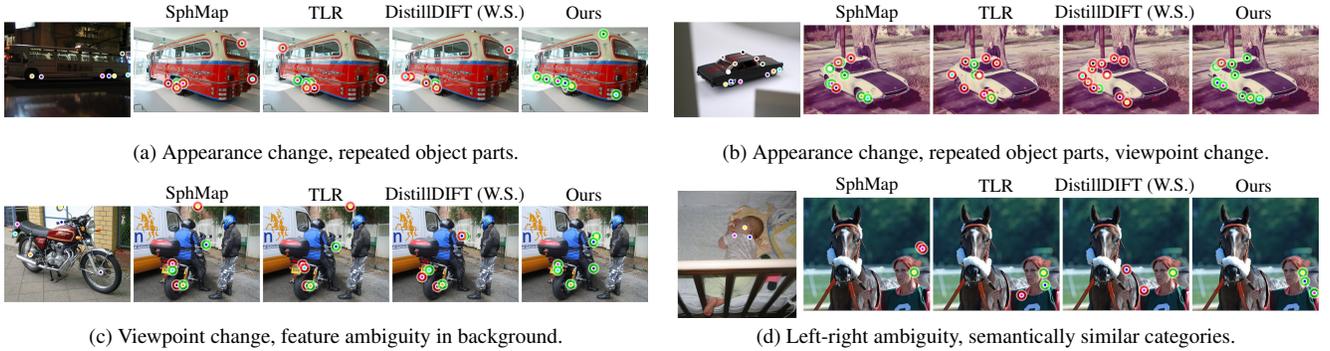
\begin{figure*}[t]
    \centering
    \begin{subfigure}[b]{0.49\linewidth}
        \centering
        \begin{tikzpicture}
            \node[anchor=north west, inner sep=0] (img) at (0,0) {\includegraphics[width=\textwidth]{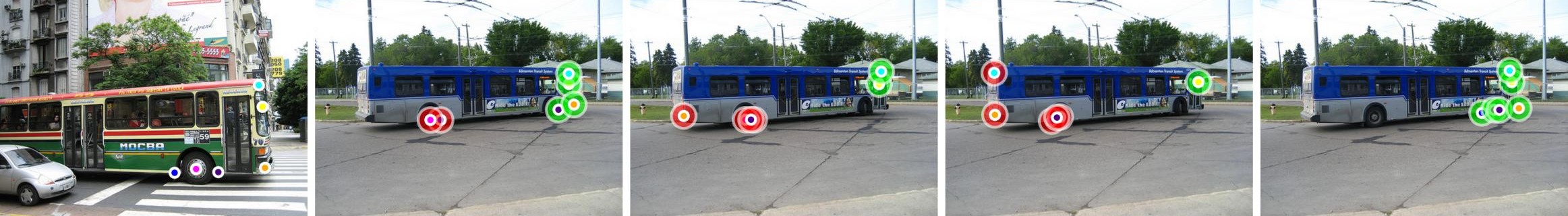}};
            \begin{scope}[shift={(img.south west)}, x={(img.south east)}, y={(img.north west)}]
                \node[align=center] at (0.3,1.10) {\scriptsize{SphMap}};
                \node[align=center] at (0.5,1.10) {\scriptsize{TLR}};
                \node[align=center] at (0.7,1.10) {\scriptsize{DistillDIFT (W.S.)}};
                \node[align=center] at (0.9,1.10) {\scriptsize{Ours}};
            \end{scope}
        \end{tikzpicture}
        \caption{Appearance change, repeated object parts.}
    \end{subfigure}\hfill
    \begin{subfigure}[b]{0.49\linewidth}
        \centering
        \begin{tikzpicture}
            \node[anchor=north west, inner sep=0] (img) at (0,0) {\includegraphics[width=\textwidth]{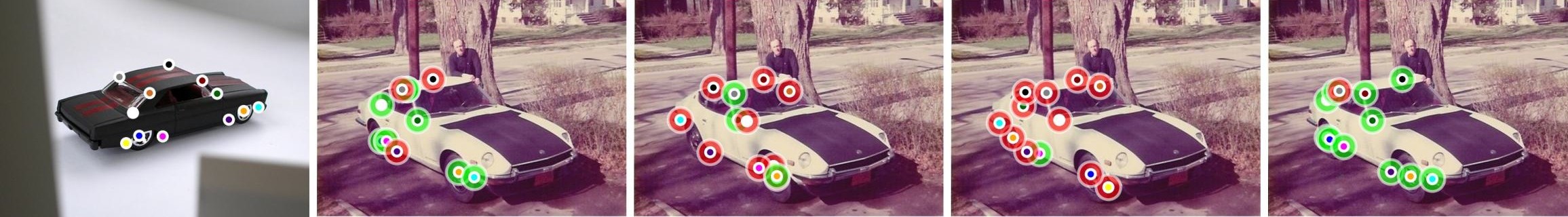}};
            \begin{scope}[shift={(img.south west)}, x={(img.south east)}, y={(img.north west)}]
                \node[align=center] at (0.3,1.10) {\scriptsize{SphMap}};
                \node[align=center] at (0.5,1.10) {\scriptsize{TLR}};
                \node[align=center] at (0.7,1.10) {\scriptsize{DistillDIFT (W.S.)}};
                \node[align=center] at (0.9,1.10) {\scriptsize{Ours}};
            \end{scope}
        \end{tikzpicture}
        \caption{Appearance change, repeated object parts, viewpoint change.}
    \end{subfigure}\hfill
    \begin{subfigure}[b]{0.49\linewidth}
        \centering
        \begin{tikzpicture}
            \node[anchor=north west, inner sep=0] (img) at (0,0) {\includegraphics[width=\textwidth]{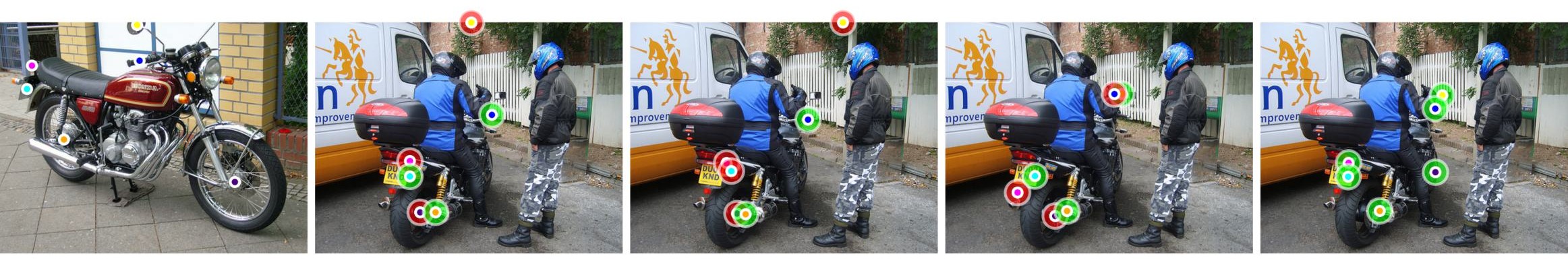}};
            %\begin{scope}[shift={(img.south west)}, x={(img.south east)}, y={(img.north west)}]
            %    \node[align=center] at (0.3,0.97) {\scriptsize{SphMap}};
            %    \node[align=center] at (0.5,0.97) {\scriptsize{TLR}};
            %    \node[align=center] at (0.7,0.97) {\scriptsize{DistillDIFT (W.S.)}};
            %    \node[align=center] at (0.9,0.97) {\scriptsize{Ours}};
            %\end{scope}
        \end{tikzpicture}
        \caption{Viewpoint change, feature ambiguity in background.}
    \end{subfigure}\hfill
    \begin{subfigure}[b]{0.49\linewidth}
        \centering
        \begin{tikzpicture}
            \node[anchor=north west, inner sep=0] (img) at (0,0) {\includegraphics[width=\textwidth]{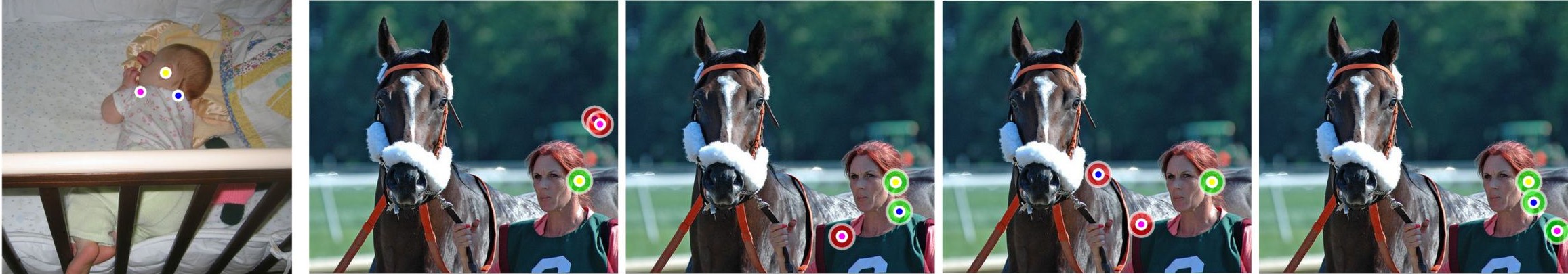}};
            %\begin{scope}[shift={(img.south west)}, x={(img.south east)}, y={(img.north west)}]
            %    \node[align=center] at (0.3,1.10) {\scriptsize{SphMap}};
            %    \node[align=center] at (0.5,1.10) {\scriptsize{TLR}};
            %    \node[align=center] at (0.7,1.10) {\scriptsize{DistillDIFT (W.S.)}};
            %    \node[align=center] at (0.9,1.10) {\scriptsize{Ours}};
            %\end{scope}
        \end{tikzpicture}
        \caption{Left-right ambiguity, semantically similar categories.}
    \end{subfigure}
    \caption{\textbf{Qualitative results for challenging examples.} We present four challenging examples of the SPair-71K test dataset where current SOTA models mostly fail.
    Correct matches that lie within the PCK$_\text{bbox}@0.1$ radius of the ground-truth label are color-coded in green, while incorrectly matched keypoints are depicted in red.
    }
    \label{fig:qual_examples}
    \vspace{-.3cm}
\end{figure*}

\section{Experiments}
\label{sec:experiments}

\paragraph{Details on pseudo-label generation.}
Using our sampling strategy, we generate 30k pseudo-labeled image pairs per category from SPair-71k.
We use a maximum of 50 randomly sampled keypoints for supervised training for each image pair.
Following \citet{zhang2023tale}, we resize the input images to $960^2$ and $840^2$ to extract the SD and DINOv2 features, resulting in feature maps of resolution $60 \times 60$.
We choose the rejection radius $r_\text{max}$ for the relaxed bicyclic consistency constraint such that a deviation of one feature patch is allowed.

\paragraph{Details on adapter training.}
Similar to \citet{zhang2024telling}, we use four bottleneck layers with 5M parameters for the adapter. 
The model is trained using the AdamW optimizer with a weight decay rate $0.001$, a learning rate of $5\cdot 10^{-3}$, and the one-cycle scheduler for 200k steps.

\paragraph{Datasets.}
Recent works~\citep{tang2023emergent,zhang2023tale,zhang2024telling,mariotti2024improving} find SPair-71k \cite{min2019spair} to be the most challenging and informative semantic matching benchmark due to its larger size and more challenging matches, containing images from $18$ different categories with $600$ to $900$ image pairs per category.
We also evaluate the methods on the recently proposed AP-10k \citep{yu2021ap} semantic correspondence dataset \citep{zhang2024telling} without training on it.

\paragraph{Metrics.}
We follow the standard settings~\citep{gupta2023asic,huang2022learning,zhang2024telling,xue2025matcha,mariotti2024improving,min2019spair} and evaluate the semantic correspondence performance via the Percentage of Correct Keypoints (PCK).
It is defined by computing the ratio of correctly predicted matched keypoints that lie within a radius of $R=\alpha \cdot\max (h,w)$ around the correct ground truth match, where we report the \textit{bbox} variant with $h$ and $w$ referring to the height and width of the bounding box of the considered object, respectively.
Since various prior works report it differently, we report both the PCK when averaged over keypoints (\textit{per kpt}) and when averaged over images (\textit{per img}).

\paragraph{Prior work.}
We compare our method with recent zero-shot and weakly-supervised strategies.
\textit{\mbox{DIFT}}~\citep{tang2023emergent}, \textit{\mbox{DINOv2}}~\citep{oquab2023dinov2}, and \textit{\mbox{SD + DINOv2}}~\citep{zhang2023tale} extract features from foundation models and perform nearest-neighbor matching in feature space. \textit{\mbox{DistillDIFT (U.S.)}}~\citep{fundel2024distillation} is further fine-tuned for correspondence detection on 3D data. \textit{ASIC}~\citep{gupta2023asic} uses DINO features to establish coarse correspondences and refines them at test-time.
Spherical mapper (\textit{\mbox{SphMap}})~\citep{mariotti2024improving} uses class labels, object masks, and 3D pose information as weak supervision signals during training, similar to us. We note, however, that we do not use any of this information at test time.
Telling Left from Right~(\textit{\mbox{TLR}})~\citep{zhang2024telling} and \textit{\mbox{DistillDIFT (W.S.)}}~\citep{fundel2024distillation} make use of the keypoint label definitions to flip images at test time.
We compare to zero-shot approaches and the spherical mapper \citep{mariotti2024improving} trained on ImageNet-3D with our extended 2-sphere configuration.

\subsection{Experimental Results}

\paragraph{Quantitative results on SPair-71k.}
The category-specific results in \cref{tab:Spair_all_cats} show that our model achieves strong gains across nearly all categories, setting a new SOTA on SPair-71k and demonstrating the effectiveness of our pseudo-labeling strategy.
Improvements are especially large for categories that are symmetric or contain repeated object parts, \eg, $+15.7\%$ for bus or $+14.0\%$ for car.
While \citet{mariotti2024improving} already improved performance for such cases, our approach outperforms theirs by effectively combining SD and DINOv2 features through pseudo-label supervision. 
Unlike their method, we retain original features and simply reject matches that strongly deviate from a spherical prototype.

All in all, we clearly outperform all previous methods on all PCK-levels (\cref{tab:Spair_all_cats} and \cref{tab:per-image-pck}), including methods that use the same weak supervision as we do~\cite{mariotti2024improving} and methods that include dataset-specific information about the definition of the keypoint labels~\citep{zhang2024telling, fundel2024distillation}.
Surprisingly, training an adapter only on top of DINOv2 features with the same pseudo-labels results in competitive semantic correspondence performance compared to the adapter on top of SD+DINOv2 features, even outperforming this strategy for the animal AP-10K dataset. 
This demonstrates that DINOv2 features achieve competitive semantic correspondence performance if refined effectively.

\begin{table}
  \centering
  \resizebox{\columnwidth}{!}{
  \begin{tabular}{lcllcccc}
    \toprule
    &&   \multicolumn{3}{c}{SPair-71k}& \multicolumn{3}{c}{AP-10k (PCK@0.1)}\\
        \cmidrule(lrr){3-5}\cmidrule(llr){6-8}
        Models &&  0.1 & 0.05 & 0.01 &  I.S & C.S. & C.F.\\
    \midrule
    \myrowcolour
    \oursTab\ (SPair) &&    {71.6} & {53.8} &{10.1}& {70.6} & {69.1} & {57.8}\\
    SphMap-$\mathcal{S}^2$ (IN3D) &&  58.5& 44.0&7.7&  56.6&53.2&38.9\\ %exp_002_in3d_spherical_cc__tlr_sample_0_sph_True_wsph_0.2_kpt_True_onlydino_False
    \myrowcolour
    %\oursTab\ (SphMap-\mathcal{S}) (IN3D) &&  68.0 & 51.1 &9.8&  65.9&64.7&52.5\\
    \oursTab\  (IN3D) &&  68.0 & 51.1 &9.8&  67.8&65.8&53.3\\
    \oursTab\  (IN3D $\rightarrow$ SPair) && \best{72.2} & \best{54.6} & \best{10.6} & \best{71.1} & \best{69.4} & \best{58.1}\\
    \bottomrule
  \end{tabular}
  }
  \caption{
  \textbf{Effect of pre-training on ImageNet-3D (PCK \textit{per img}).}
  SphMap-$\mathcal{S}^2$ represents the weighted application of our 2-sphere-based spherical mapper, evaluated following \citet{mariotti2024improving}.
  }
  \vspace{-.6cm}
  \label{tab:per-image-pck-imagenet3d}
\end{table}

\paragraph{Evaluation on AP-10k.}
Additionally, we also evaluate on the AP-10k dataset and report the results in \cref{tab:per-image-pck}.
We find that, despite not being trained on this dataset and only having a few animal categories in the SPair-71k training dataset, we outperform the current SOTA, the geometry-aware approach by \citet{zhang2024telling}, showcasing the generalization capabilities of our method.
We additionally compare to a fully-supervised model~\citep{zhang2024telling} trained on SPair-71k. We observe a severe performance drop of this model ($- 17.4\%$ PCK@0.1 for I.S.) compared to fully-supervised training on AP-10k~\citep[Tab.3]{zhang2024telling}.
On the other hand, our approach does not suffer such a performance drop and, thus, achieves comparable results to a model trained with manual annotations.
We hypothesize that this is due to the denser and more diverse pseudo-labels compared to the fixed set of manually annotated keypoints that are being overfitted in supervised training.

\paragraph{Scaling to a larger dataset.}
Our weakly-supervised training strategy allows scaling the training to larger object-centric datasets that do not include keypoint annotations.
While pre-training improves the model performance on SPair-71k and AP-10k (\cref{tab:per-image-pck} and \cref{tab:per-image-pck-imagenet3d}), we surprisingly even outperform the previous SOTA on SPair-71k despite not having trained on the SPair-71k dataset, showcasing strong generalization capabilities.
This is in contrast to the na\"ive weighted application of the ImageNet-3D spherical mapper, where the results significantly deteriorate, potentially explained by the fact that this model does not generalize well to the unseen categories in SPair-71k.
We also study the impact of automatically acquired 3D pose labels in the supplementary.

\paragraph{Qualitative results.}
We present qualitative results in \cref{fig:qual_examples} comparing our model to current SOTA models on four challenging examples of the SPair-71k test dataset.
%, uncovering their fundamental weaknesses and supporting our proposed strategy.
While the spherical mapper effectively addresses cases of symmetries and repeated object parts for rigid objects, it comes with the inherent challenge of finding the correct weighting factor for interpolating between foundational and processed spherical feature similarities, where the optimal weight depends on the considered category and sample.
For example, while the weighting is too low in (a), the weighting is too large in (d) since the ill-defined background features of SphMap result in a match with an object in the background.
Additionally, its effectiveness is limited to categories that are well represented by a spherical prototype, \eg, a potential reason for the motorbike failure case in (c).
TLR and DistillDIFT both suffer from the fact that simply flipping an object does not resolve all ambiguities, such as for categories with repeated object parts (a,b), rotated objects (b), or cases with left-/right ambiguities (d).
DistillDIFT is less prone to incorrect matches caused by background clutter (c), whereas TLR and SphMap suffer from the fact that they rely on the potentially noisy nearest neighbor of SD and DINOv2 features.
However, DistillDIFT still fails to disambiguate the human and the horse in (d), and it suffers from left-right ambiguity if the flipping is not effective (d).
Furthermore, as visible in (b), DistillDIFT's performance deteriorates for pairs with heavy appearance change, which may be attributed to the multi-view fine-tuning on 3D instances, which does not generalize well to cross-instance correspondences.
In contrast, our method identifies high-quality semantic correspondences for these challenging examples, as it successfully resolves feature ambiguities and does not rely on the weighting of two similarity terms.

\subsection{Ablations}
We thoroughly ablate all components of our method and present the main results in \cref{tab:ablations}.
First, supervision with na\"ively generated pseudo-labels already outperforms the zero-shot approach.
This is due to the fact that learning the light-weight adapter naturally stabilizes feature maps for semantic correspondence estimation by merging the complementary information captured by SD and DINOv2 features. Constraining pseudo-labels to the mask of an object further enhances the performance.
We also observe that the relaxed cyclic consistency constraint outperforms both unfiltered pseudo-labels and exact cyclic consistency filtering.
The performance further increases when employing our chaining strategy that improves the pseudo-label quality because (1) NN matching is only performed for easier image pairs, \ie, a similar object pose, and (2) the bicyclic consistency constraint reduces spurious matches.
Adding the sphere-based rejection strategy substantially improves all model instantiations by reducing wrong matches for objects with symmetries and repeated object parts that make up a significant ratio of the SPair-71k dataset.
This rejection also clearly enhances the naive pseudo-labeling strategy, combining all modules to achieve the best performance.

\begin{table}
\centering
\resizebox{\columnwidth}{!}{
\begin{tabular}{
c@{\hskip 0.1cm}
|c@{\hskip 0.1cm}
|c@{\hskip 0.1cm}
|c@{\hskip 0.1cm}
|c@{\hskip 0.1cm}
|c}
\toprule
pseudo & cyc. cons. & relaxed c.c. & chaining & sph. rej. & PCK@0.1 \\
\myrowcolour
\midrule
       &            &              &          &           & 65.0      \\
\checkmark      &            &              &          &           & 67.2      \\
\myrowcolour
\checkmark      & \checkmark          &              &          &           & 66.9      \\
\checkmark      &            & \checkmark            &          &           & 68.4      \\
\myrowcolour
\checkmark      &            & \checkmark            & \checkmark        &           & 70.0      \\
\checkmark      &            &            &          & \checkmark         & 72.9      \\
\myrowcolour
\checkmark      &            & \checkmark            & \checkmark        & \checkmark         & \best{74.4}      \\
\bottomrule
\end{tabular}
}
\caption{\textbf{Ablations on SPair-71k.} All introduced components bring a significant improvement. The baseline is evaluated using the SD+DINO zero-shot approach with window soft argmax.}
\label{tab:ablations}
\vspace{-.6cm}
\end{table}

\paragraph{Failure Cases.}
While our method outperforms previous methods, it still fails for some challenging cases of SPair-71K, such as images containing objects that are upside-down.
We show visual examples and discuss failure cases of our method in the supplementary.

%% file: sec/5_conclusion.tex
\section{Conclusion}
\label{sec:conclusion}
We presented \oursAbbrv, a method for finding semantic correspondences with weak supervision.
We demonstrated the effectiveness of training with pseudo-labels for semantic correspondence:
By generating labels using a zero-shot approach with foundational features and carefully designing methods for rejecting wrong matches, we provide a strong supervision signal that can be used to train an adapter that refines foundational features.  
Our experiments showed improvements of \textbf{+4.5\%} and \textbf{+6.8\%} over the previous SOTA for PCK@0.1 on SPair-71k, measured per-keypoint and per-image, respectively.
Our ablation studies have shown that the proposed chaining and filtering steps, as a means of enhancing the quality of pseudo-labels, improve the corresponding performance significantly.

\paragraph{Limitations and future work.}
Although our weak supervision strategy scales to larger datasets, reducing the supervision requirements is of importance to further scale the approach. See the supplementary for a preliminary feasibility analysis. 
Furthermore, while the match rejection with the spherical mapper shows significant improvements, it is less effective for objects that are not represented well by a sphere, which motivates extending this strategy.
We hope that the proposed improved correspondence features can enhance tasks that rely on semantic features.

\section*{Acknowledgments}
Adam Kortylewski acknowledges support via his Emmy Noether Research Group funded by the German Research Foundation (DFG) under Grant No. 468670075. 
Thomas Wimmer is supported through the Max Planck ETH Center for Learning Systems.

%% file: sec/x_supplementary.tex
%%%%%%%%%%%%%%%%%%%%%%%%%%%%%%%%%%%%%%%%%%%%%%%%%%%%%%%%%%%%
\clearpage
\setcounter{page}{1}
 {
   \newpage
       \twocolumn[
        \centering
        \Large
        \textbf{\thetitle}\\
        \vspace{0.5em}Supplementary Material \\
        \vspace{1.5em}
       ] %< twocolumn
   }

\appendix

%\section{Appendix} \label{sec:supplementary}

\section{Details on ImageNet-3D Trained Model} \label{sec:suppl-imagenet3d}

In this subsection, we present more details about the model that was trained on ImageNet-3D, including more details about the improved spherical mapper.

\paragraph{Pose Conversion in ImageNet-3D.}
We reformulate the loss objective for taking into account viewpoint information as presented in \cref{eq:sphere2_sph_map}.
We acquire the needed labels in the following way:
Given the rotation matrix $R$ presenting the 3D pose in the ImageNet-3D dataset~\citep{ma_imagenet3d_2024}, we compute the corresponding coordinate on the 2-sphere $\psi=\left[ \theta, \phi \right] \in \mathcal{S}^2$ as follows:
\[
\begin{aligned}\label{eq:rot_to_sph}
\begin{pmatrix} x \\ y \\ z \end{pmatrix} &= R\,\begin{pmatrix} 0 \\ 0 \\ 1 \end{pmatrix},\\[1ex]
\end{aligned}
\]
\[\theta = \arccos(z),\quad \phi = \operatorname{atan2}(y, x)\]

\paragraph{Experimental Details}
For training the spherical mapper, we use the same hyperparameter as \citet{mariotti2024improving} and we train for 200 epochs on the ImageNet-3D dataset.
The generation of the pseudo-labels and training of the adapter follow the same hyperparameters as our presented model for SPair-71k.
We train on this larger dataset for 400k steps.

\section{Discussion of Failure Cases} \label{sec:suppl-failure-cases}

While we show significant improvements for most object categories, our method does not improve results for all classes compared to the SOTA.
Our approach performs worse when objects are vertically flipped, \eg, for the airplane category or bicycles, as presented in the challenging example in \cref{fig:fail_bicycle}.
In such cases, a limiting factor is that the polar angle is not available for 3D-aware sampling and training of the spherical mapper.
Our method also fails for heavy perspective changes \cref{fig:fail_horse}.

\begin{table}[htb!]
  \centering
  \begin{tabular}{@{}cc@{}}
    \subcaptionbox{Our model can fail for objects that are up-side down.\label{fig:fail_bicycle}}{%
      \includegraphics[width=0.45\linewidth]{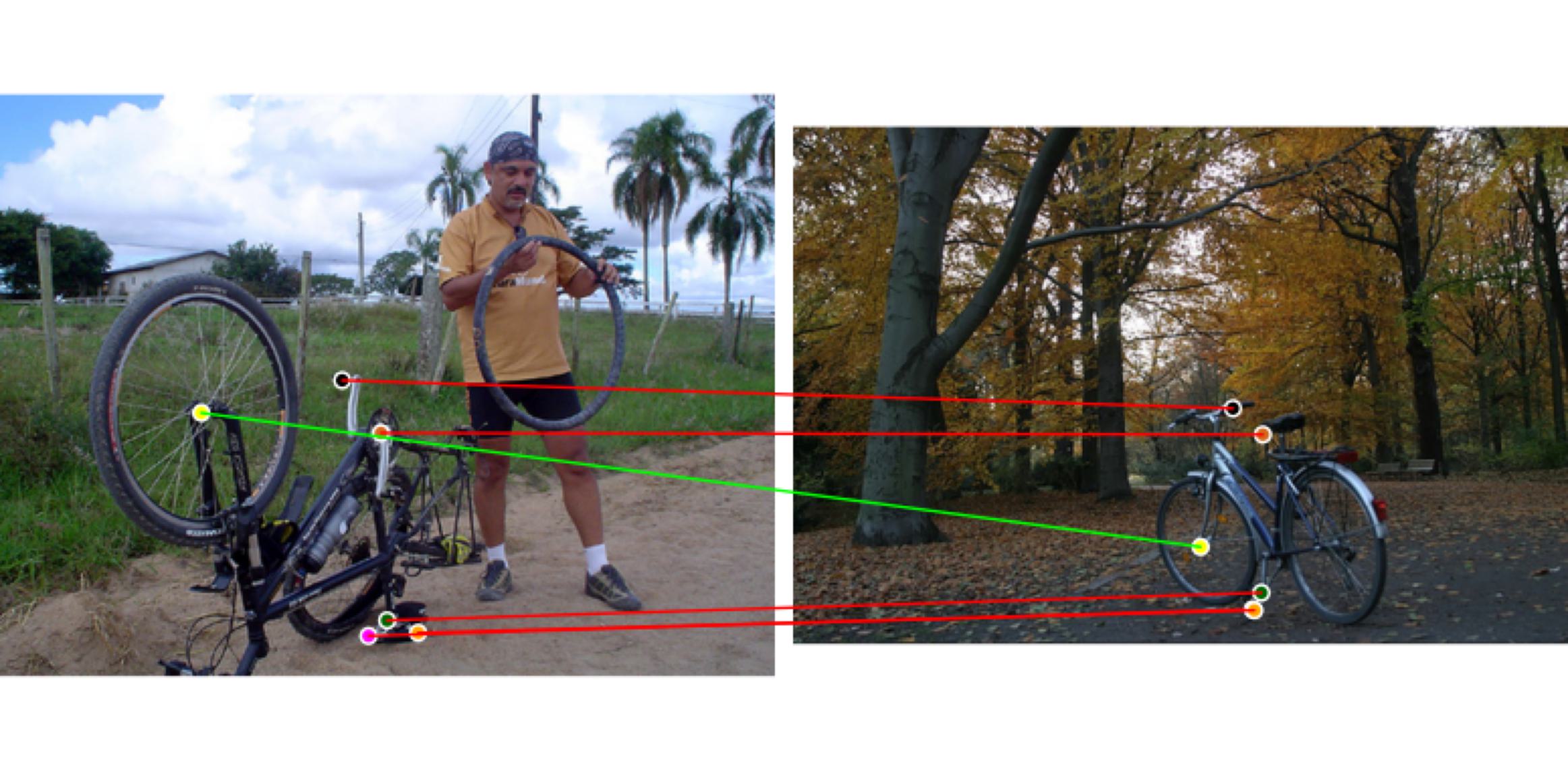}
    } &
    \subcaptionbox{Matching objects under heavy scale changes is challenging.\label{fig:fail_horse}}{%
      \includegraphics[width=0.45\linewidth]{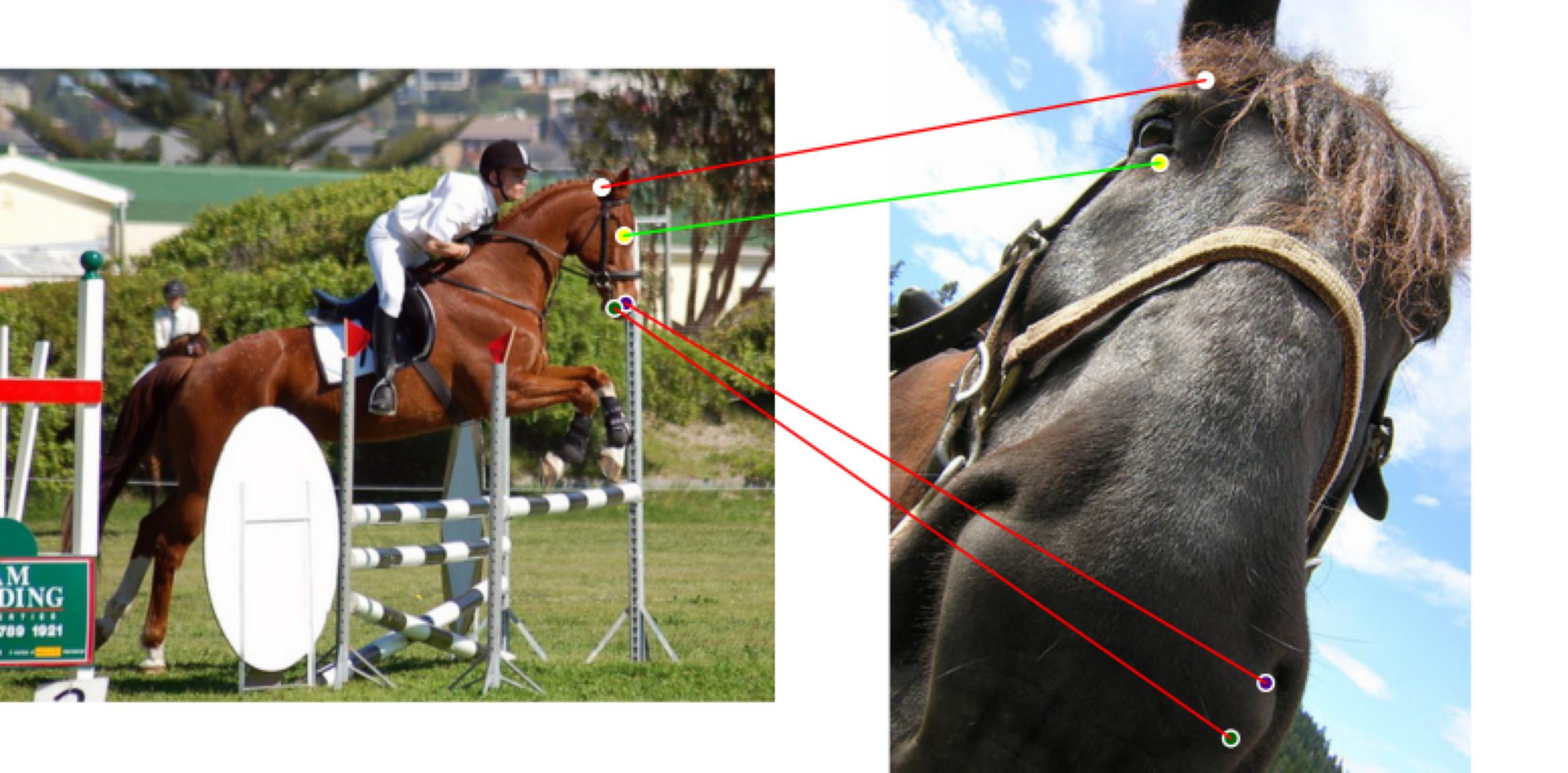}
    } \\
  \end{tabular}
  \caption{\textbf{Examples of failure cases.}}
  \label{tab:fail_examples}
  \vspace{-10pt}
\end{table}

\section{Training with Less Supervision} \label{sec:no_pose_label}
To explore the scalability of our proposed strategy to larger datasets without 3D pose annotations or masks, we study how the performance deteriorates for SPair-71k when not having access to the viewpoint annotation.
For this, we extract pose information using Orient-Anything~\citep{wang2024orient-rebuttal} and extract object masks using \texttt{\href{https://github.com/danielgatis/rembg}{rembg}} for unsupervised foreground extraction.
Using SAM masks, which are of slightly higher quality but not completely unsupervised (although feasible through, e.g., using GroundingSAM), and the automatically extracted poses, the PCK@0.1 \textit{per-img} of our method drops to 69.6\% compared to when using GT pose annotations (71.6\%). 
Using \texttt{rembg} masks, it drops further to 68.0\%, which is expected but still around 2.7p better than the previous best weakly supervised method~\citep{fundel2024distillation}.
While [47] uses dataset-specific information about the keypoint label convention, our approach only requires class labels.% that can be automatically acquired easily.
We report the results in Tab.~\ref{tab:unannotated-demo}.

\begin{table}[h]
    \centering
    \begin{tabular}{l c c}
        \toprule
        3D pose label      & mask label      & PCK@0.1 \\
        \midrule
        Ground Truth     & SAM              & 71.6         \\
        Orient-Anything  & SAM              & 69.6         \\
        Orient-Anything  & \texttt{rembg}   & 68.0         \\
        \bottomrule
    \end{tabular}
    \caption{\textbf{PCK@0.1 \textit{per-image} on SPair-71k without ground-truth viewpoint annotations.}}
    \label{tab:unannotated-demo}
    \vspace{-10pt}
\end{table}

\section{Pre-Training with Pseudo-Labels}
We explore whether pre-training with our pseudo-labels also improves the supervised performance.
For this purpose, we fine-tune the adapter with ground truth labels of the SPair-71k dataset, which improves the supervised performance from 82.9\% \citep{zhang2024telling} to 83.5\% (PCK@0.1 \textit{per-img}).

\section{Pseudo-Label Generation without SD}
When training a refiner of DINOv2 features with pseudo-labels that are acquired only from DINOv2 features, i.e., not from SD+DINOv2 as in the main paper, the performance drops to $67.17\%$ (\textit{per-img}) and to $70.29\%$ (\textit{per-kpt}), which is still on par with recent SOTA models.

\newpage

\begin{table}[t!]
\centering
\begin{tabular}{cccc}
\toprule
Channels & PCK@0.1 & PCK@0.05 & PCK@0.01 \\
\midrule
128   & 74.08 & 56.28 & 11.26 \\
384   & 74.39 & 56.87 & 11.53 \\
768   & 74.43 & 56.76 & 11.22 \\
1536  & 74.63 & 57.13 & 11.60 \\
\bottomrule
\end{tabular}
\caption{\textbf{PCK metrics for varying numbers of feature channels.}}
\label{tab:per_kpt_micro}
\vspace{-10pt}
\end{table}

\section{Ablation of Number of Feature Channels}
Learning a refining module allows reducing the number of channels of the features used for nearest neighbor computation.
The performance does not heavily drop with a lower number of channels (\cref{tab:per_kpt_micro}), which might be a valuable option for memory-constrained applications.

\section{More Qualitative Results}
We show detected correspondences for uncurated image pairs from the SPair-71k test dataset for DistillDIFT, TLR, SphMap, and our method in \cref{fig:uncurated_matches_1} and \cref{fig:uncurated_matches_2}.

\input{sec/qualiative_examples}

%% file: sec/qualiative_examples.tex
\begin{figure*}[htbp]
  \centering
  \begin{tabular}{cccc}
  \multicolumn{1}{c}{DistillDIFT} & \multicolumn{1}{c}{TLR} & \multicolumn{1}{c}{SphMap} & \multicolumn{1}{c}{Ours} \\[5pt]
    \includegraphics[width=0.22\textwidth]{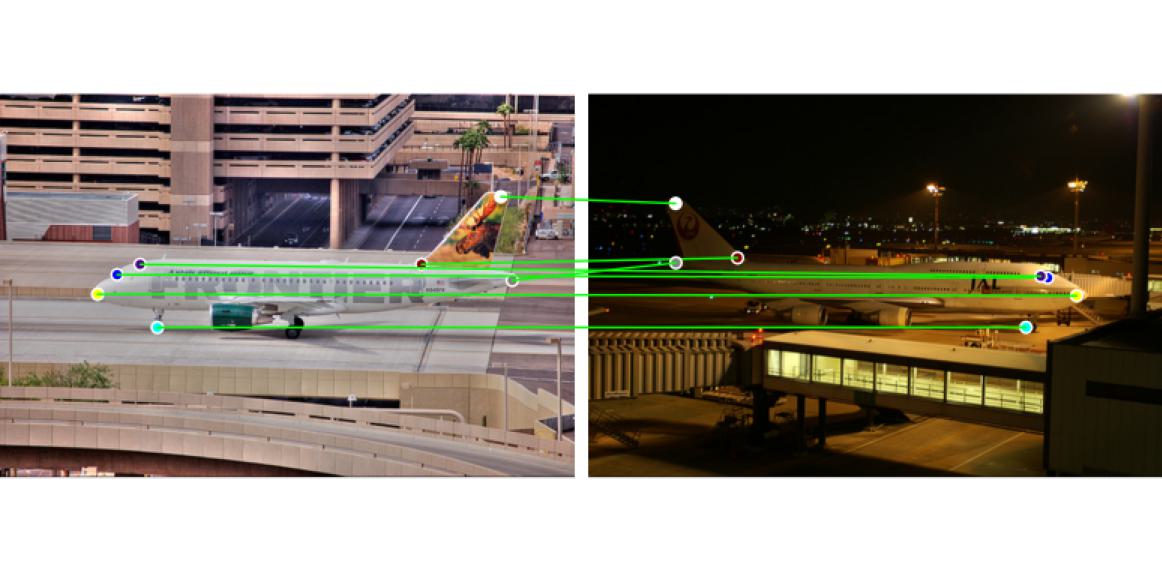} & 
    \includegraphics[width=0.22\textwidth]{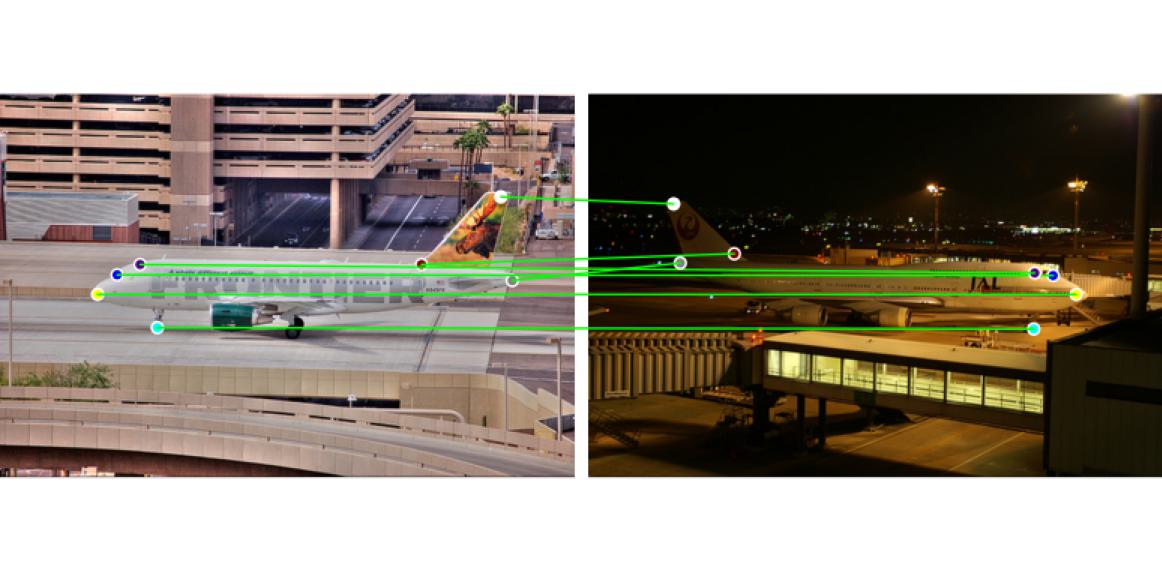} & 
    \includegraphics[width=0.22\textwidth]{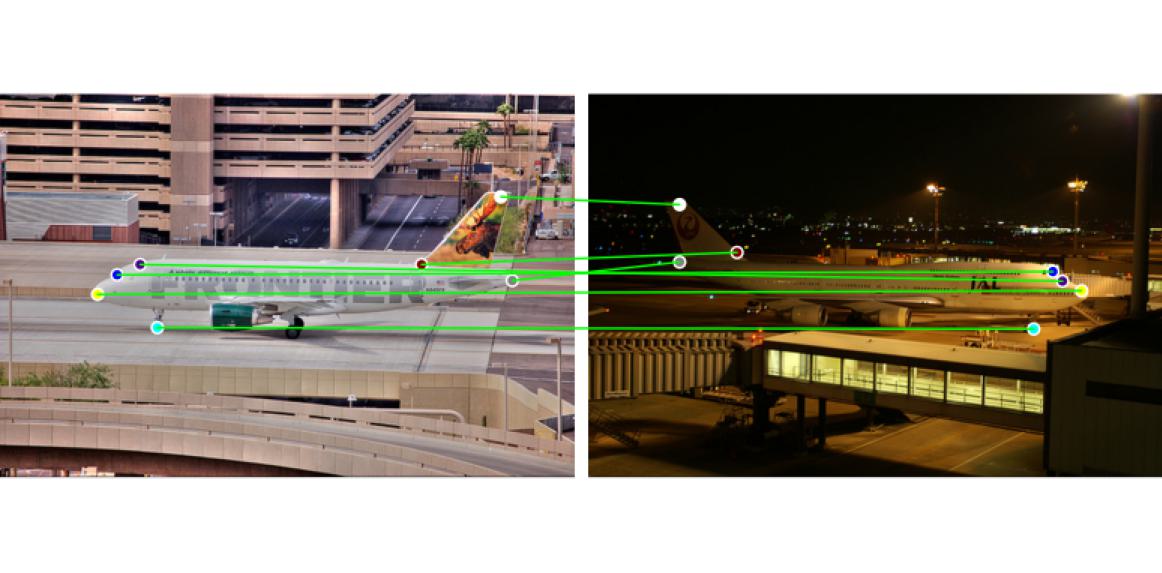} & 
    \includegraphics[width=0.22\textwidth]{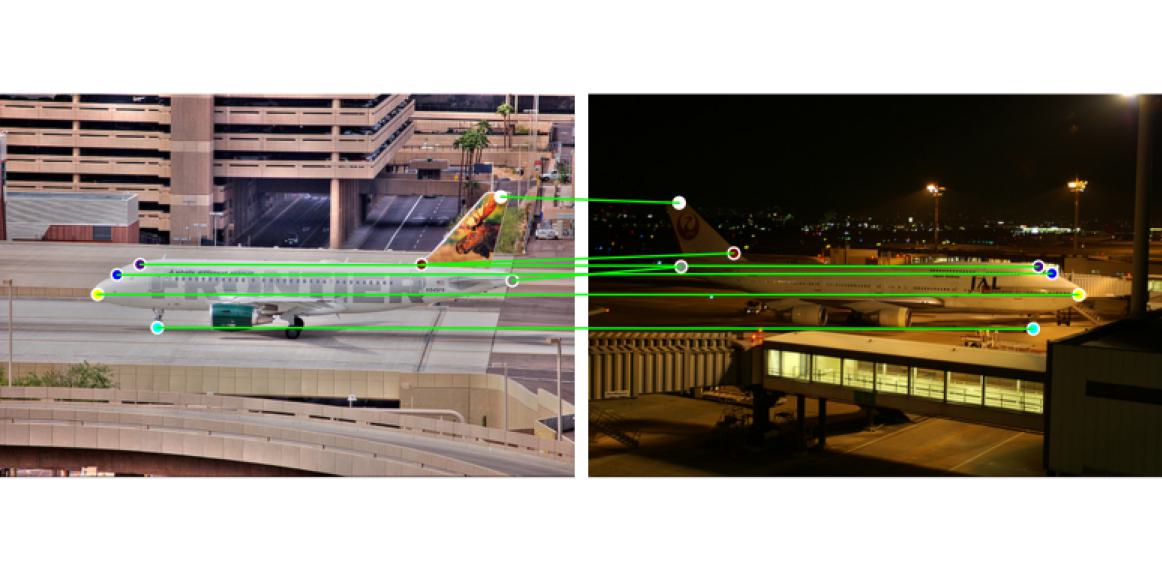} \\
    \includegraphics[width=0.22\textwidth]{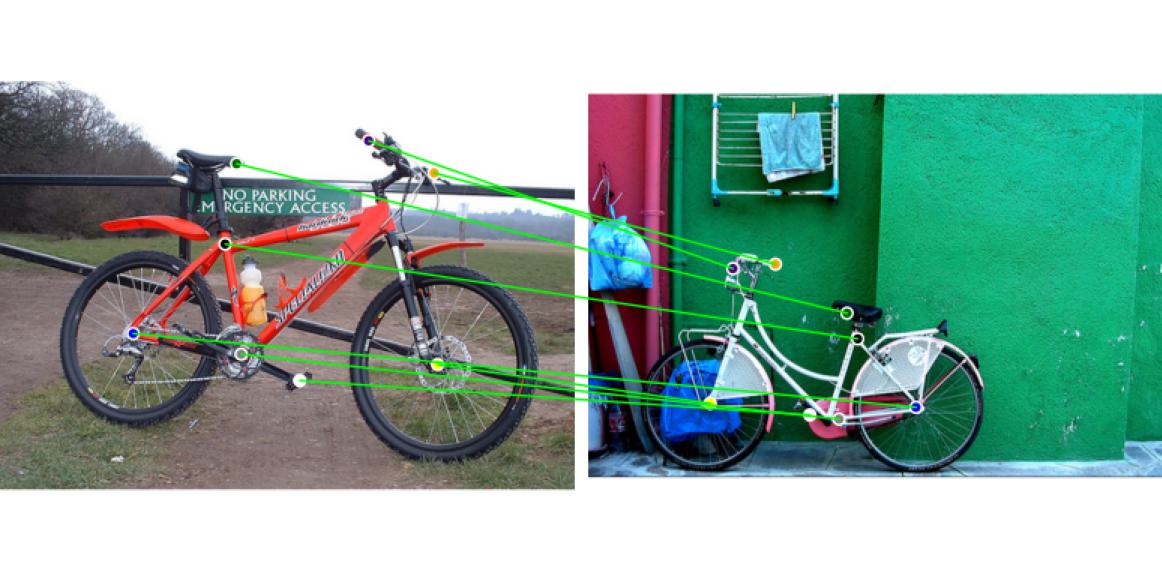} & 
    \includegraphics[width=0.22\textwidth]{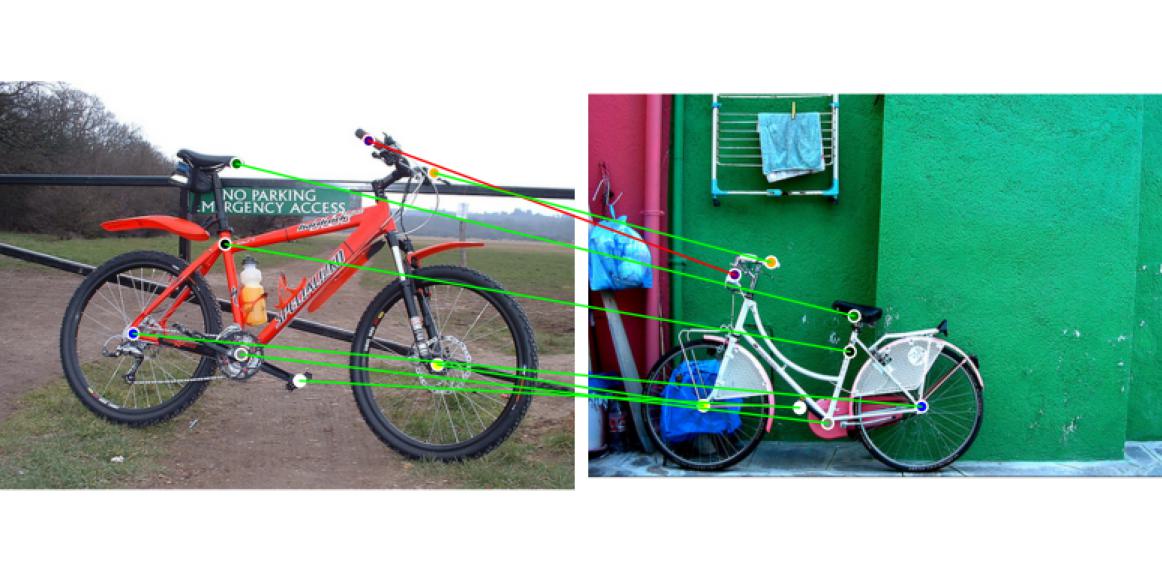} & 
    \includegraphics[width=0.22\textwidth]{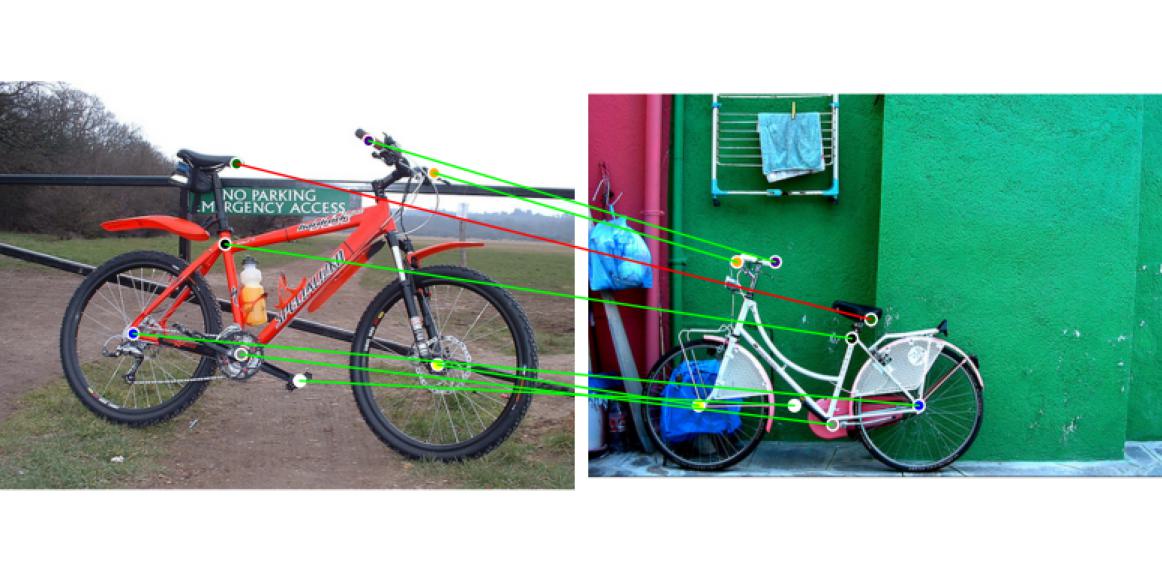} & 
    \includegraphics[width=0.22\textwidth]{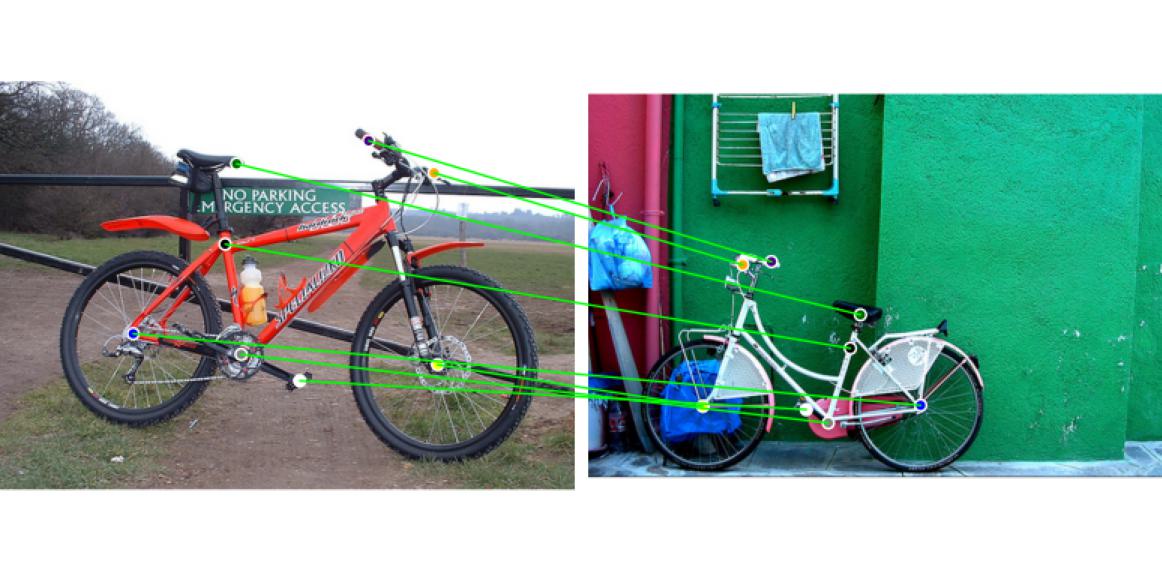} \\
    \includegraphics[width=0.22\textwidth]{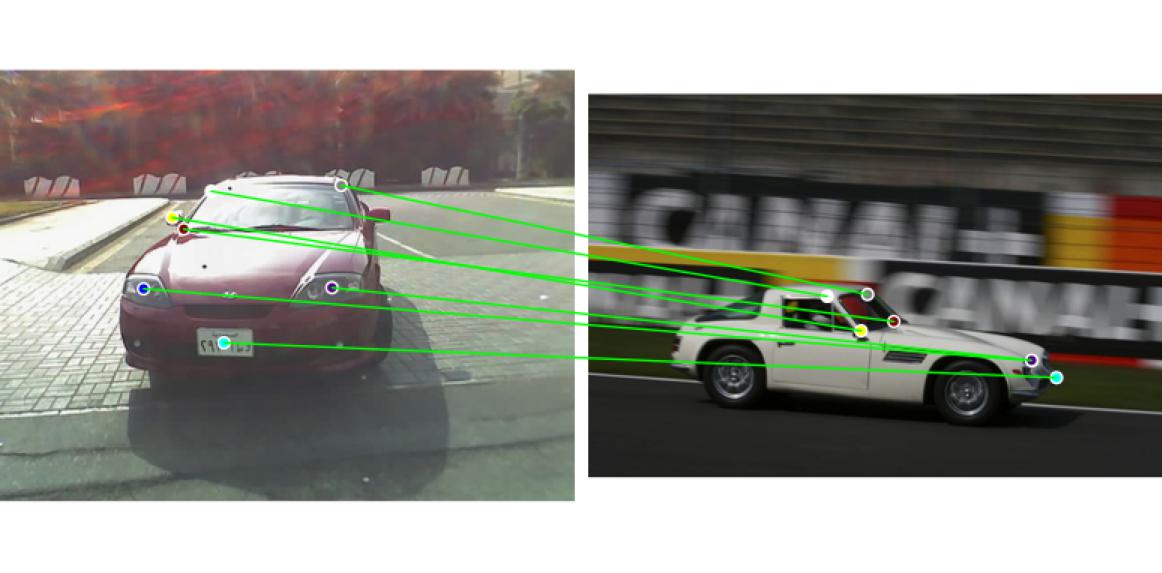} & 
    \includegraphics[width=0.22\textwidth]{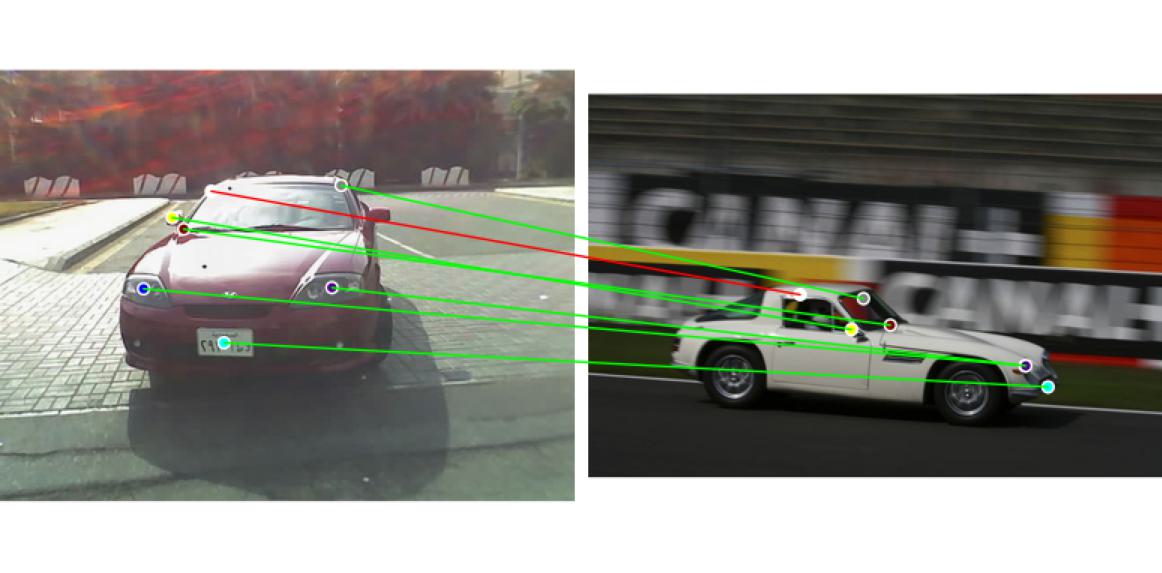} & 
    \includegraphics[width=0.22\textwidth]{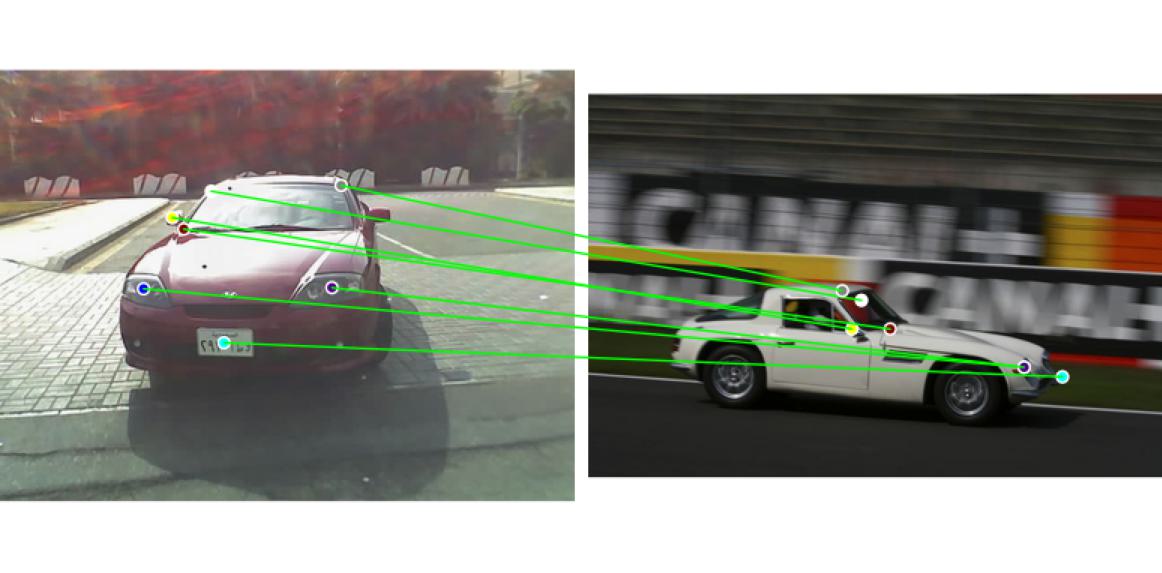} & 
    \includegraphics[width=0.22\textwidth]{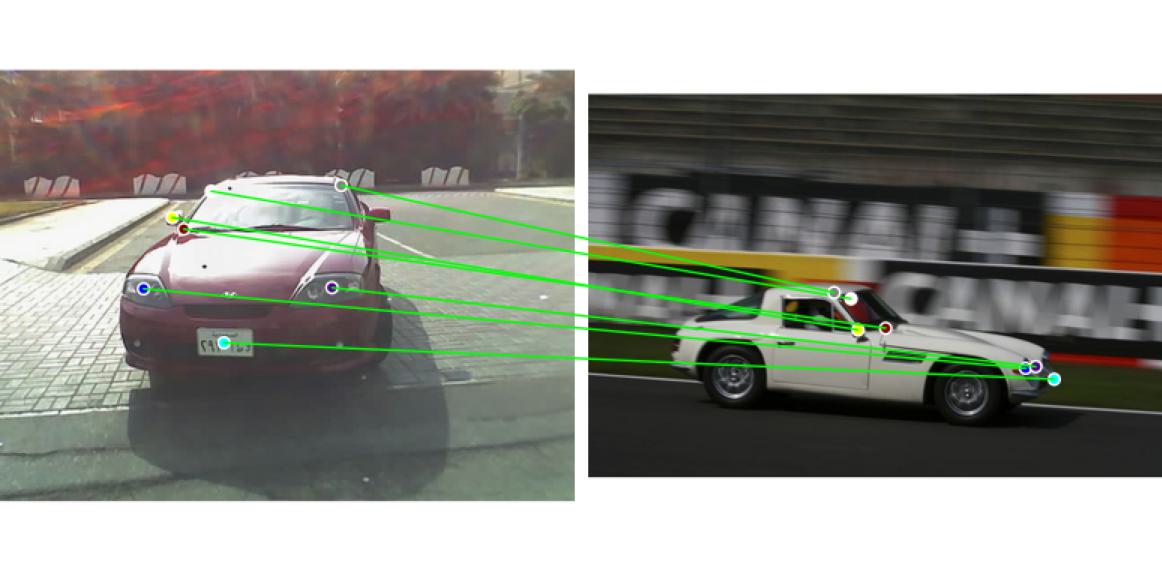} \\
    \includegraphics[width=0.22\textwidth]{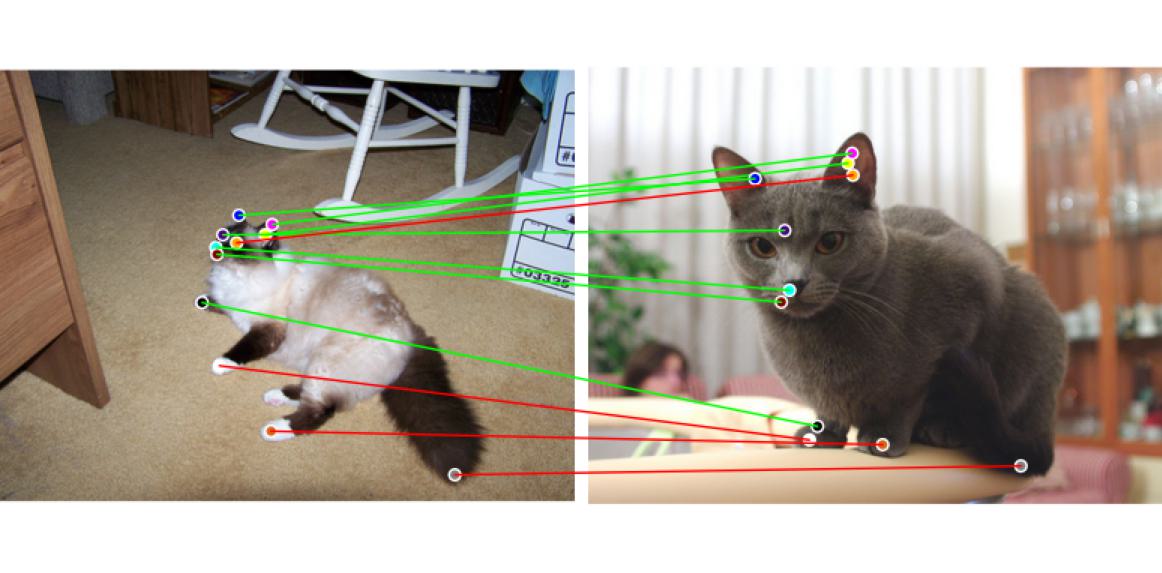} & 
    \includegraphics[width=0.22\textwidth]{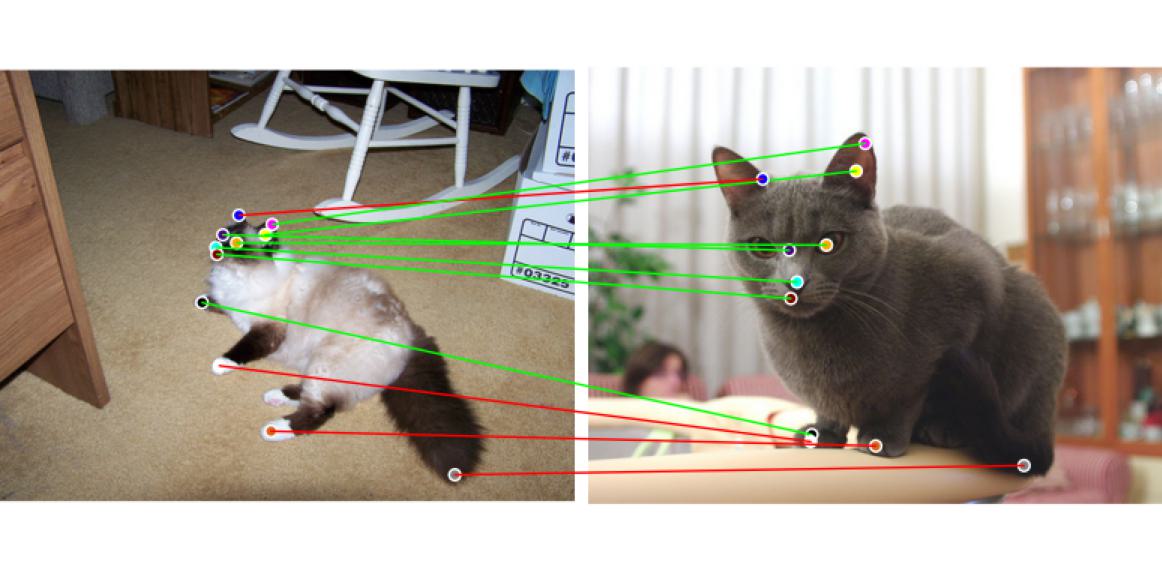} & 
    \includegraphics[width=0.22\textwidth]{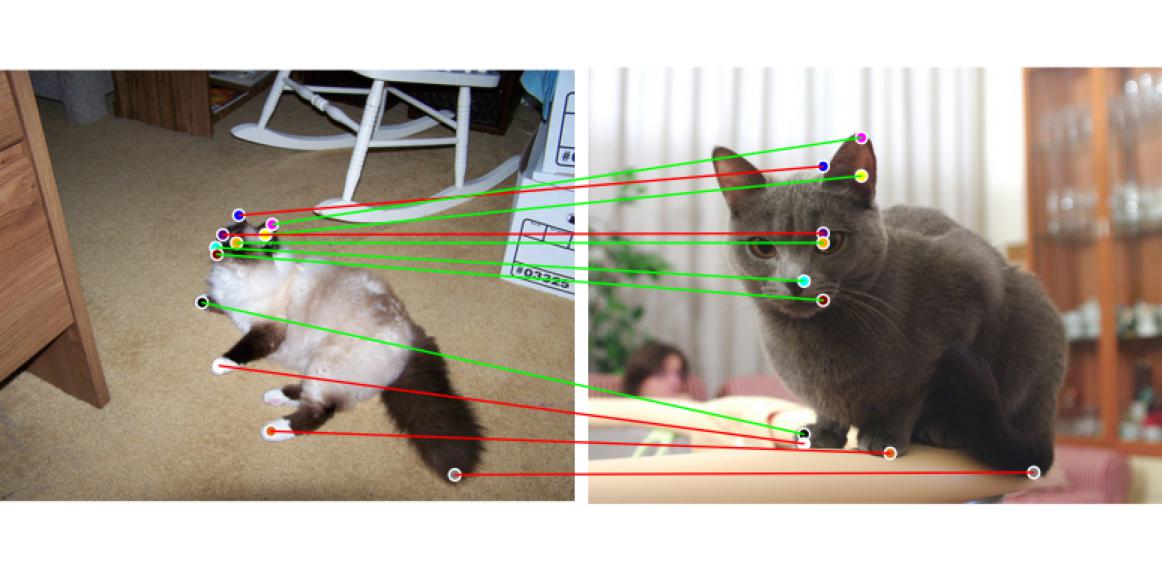} & 
    \includegraphics[width=0.22\textwidth]{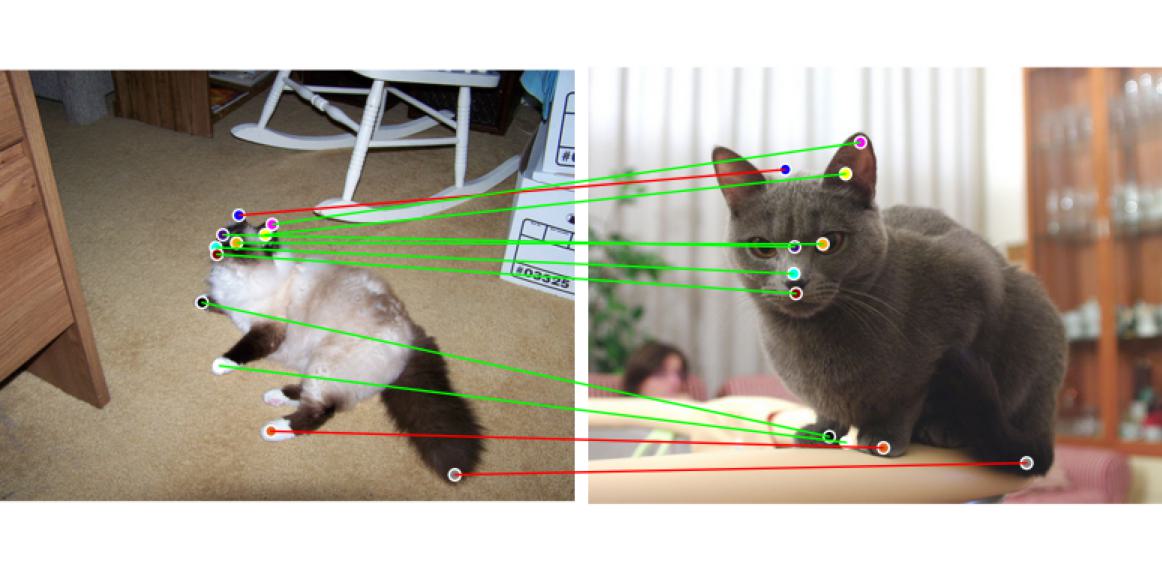} \\
    \includegraphics[width=0.22\textwidth]{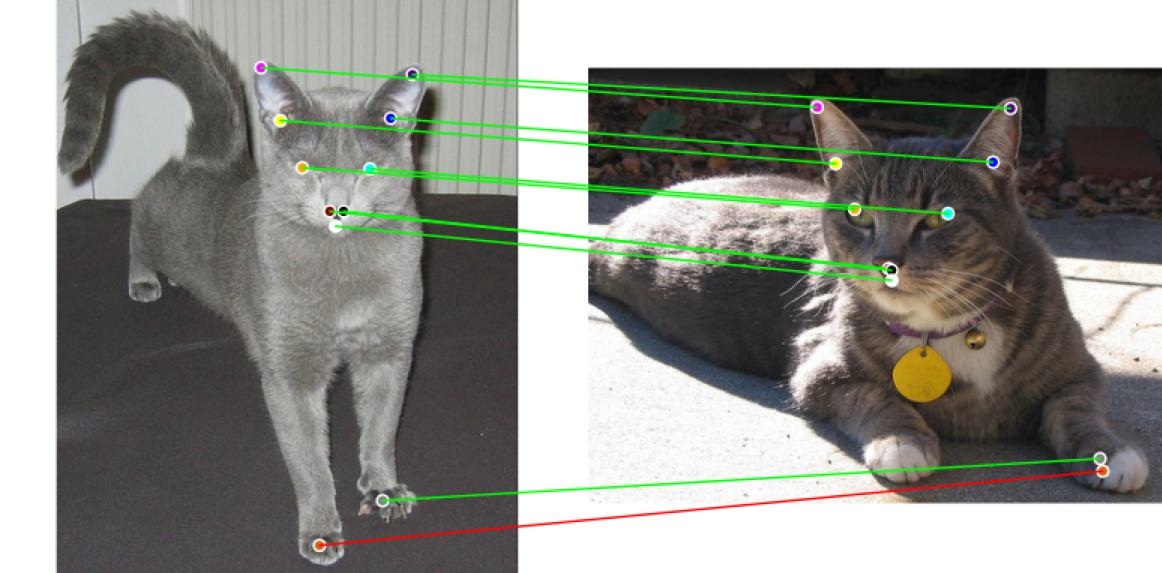} & 
    \includegraphics[width=0.22\textwidth]{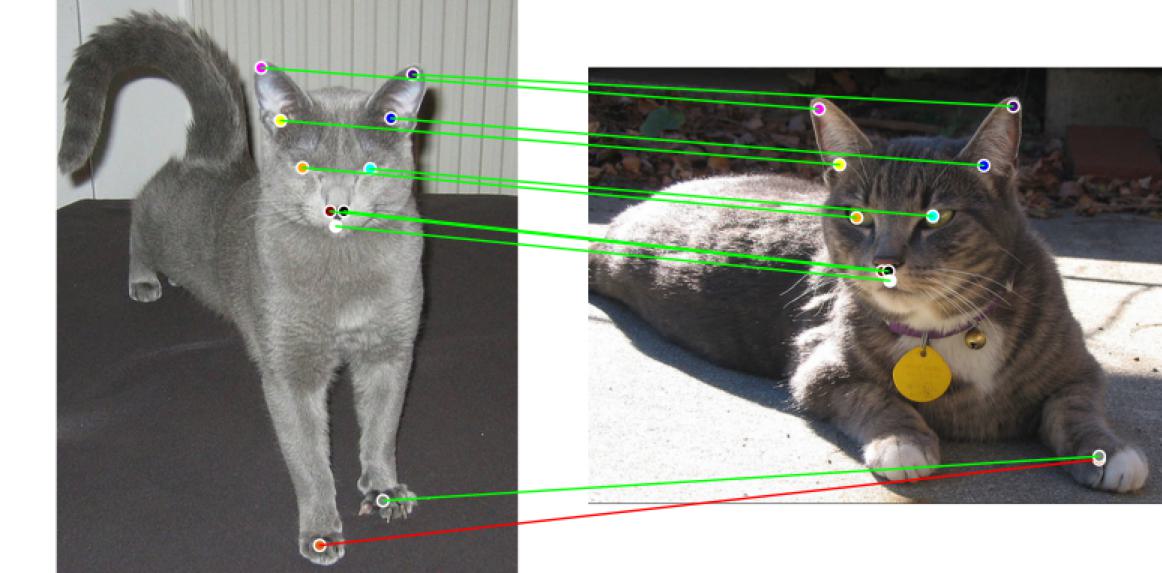} & 
    \includegraphics[width=0.22\textwidth]{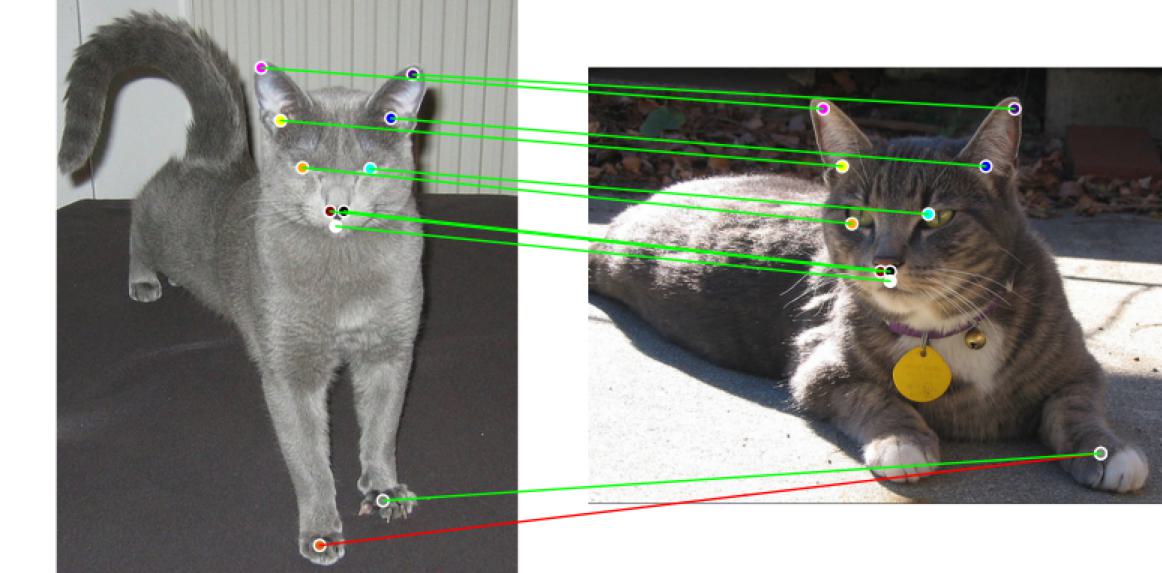} & 
    \includegraphics[width=0.22\textwidth]{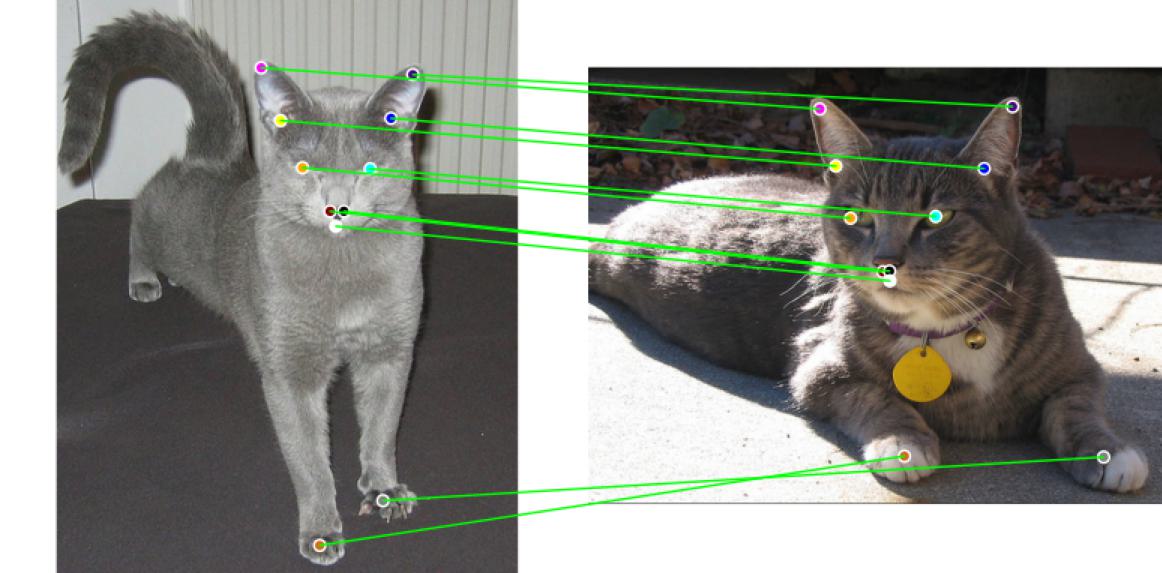} \\
    \includegraphics[width=0.22\textwidth]{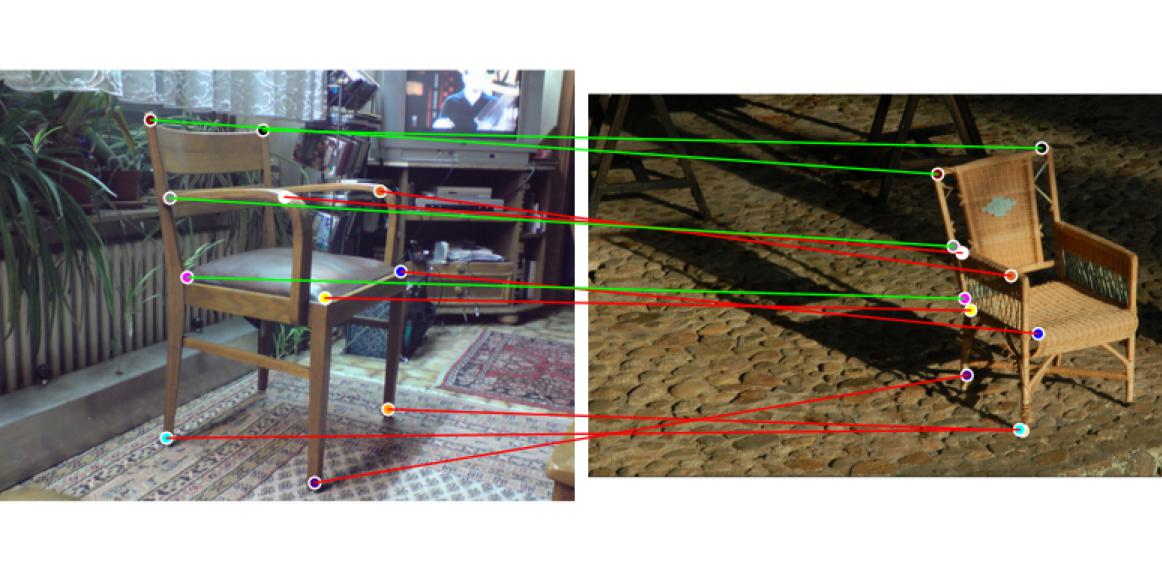} & 
    \includegraphics[width=0.22\textwidth]{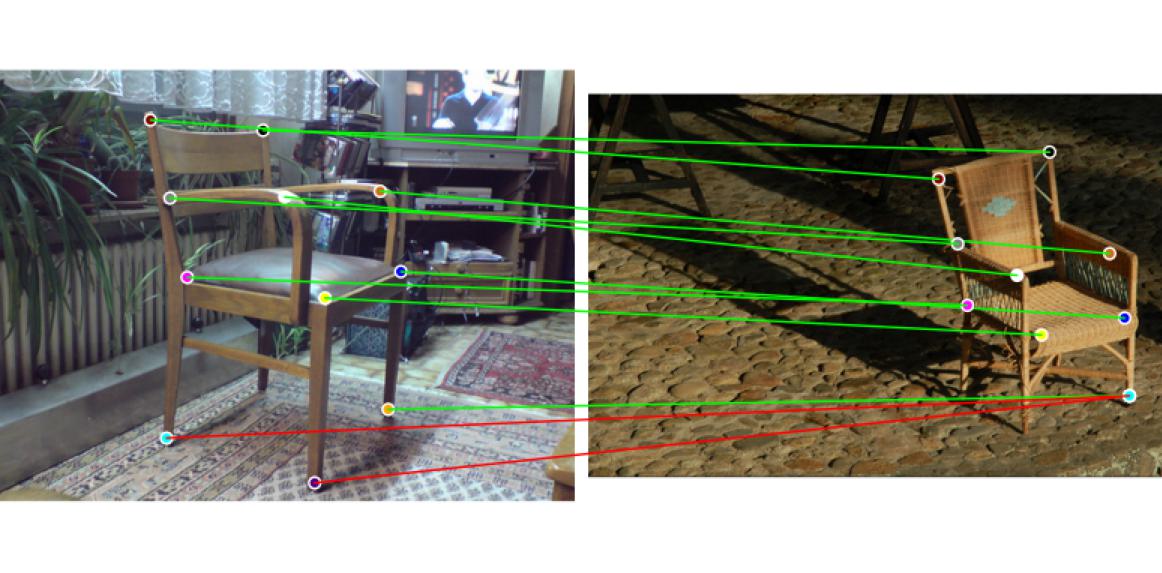} & 
    \includegraphics[width=0.22\textwidth]{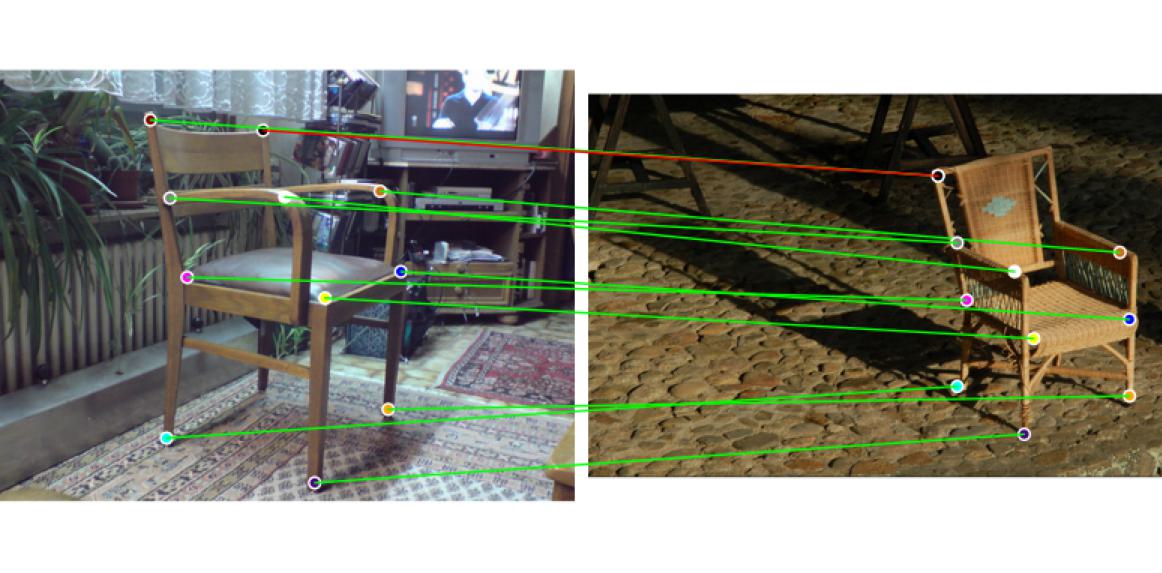} & 
    \includegraphics[width=0.22\textwidth]{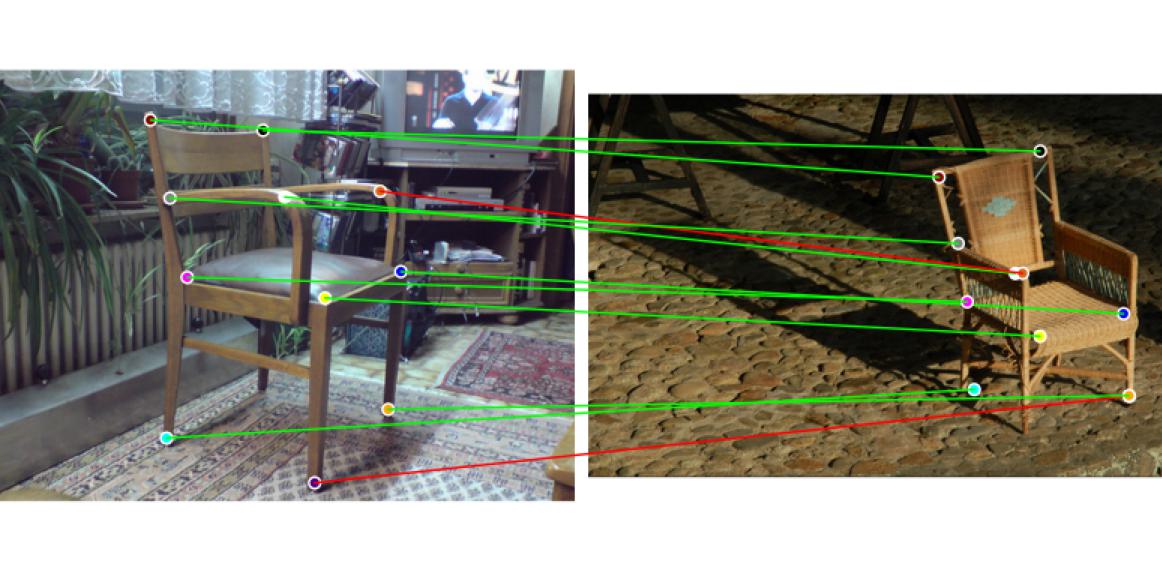} \\
    \includegraphics[width=0.22\textwidth]{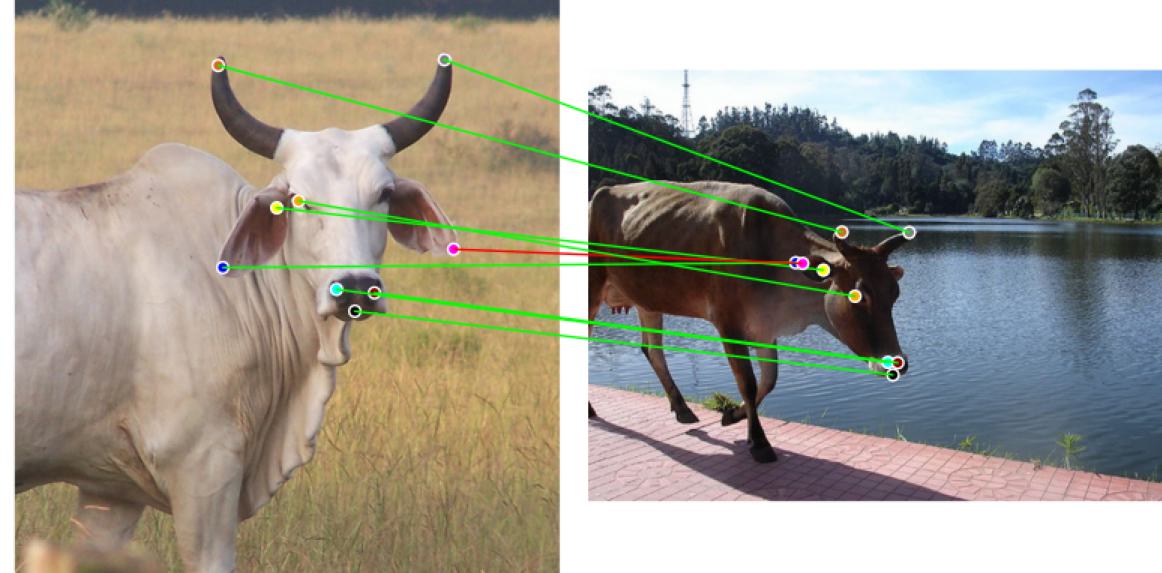} & 
    \includegraphics[width=0.22\textwidth]{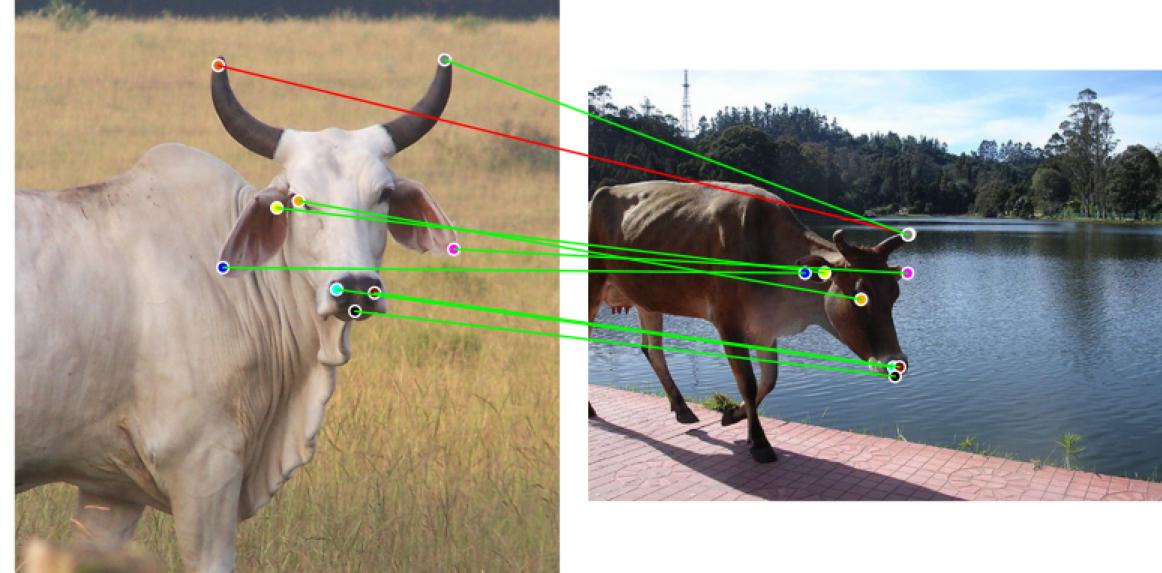} & 
    \includegraphics[width=0.22\textwidth]{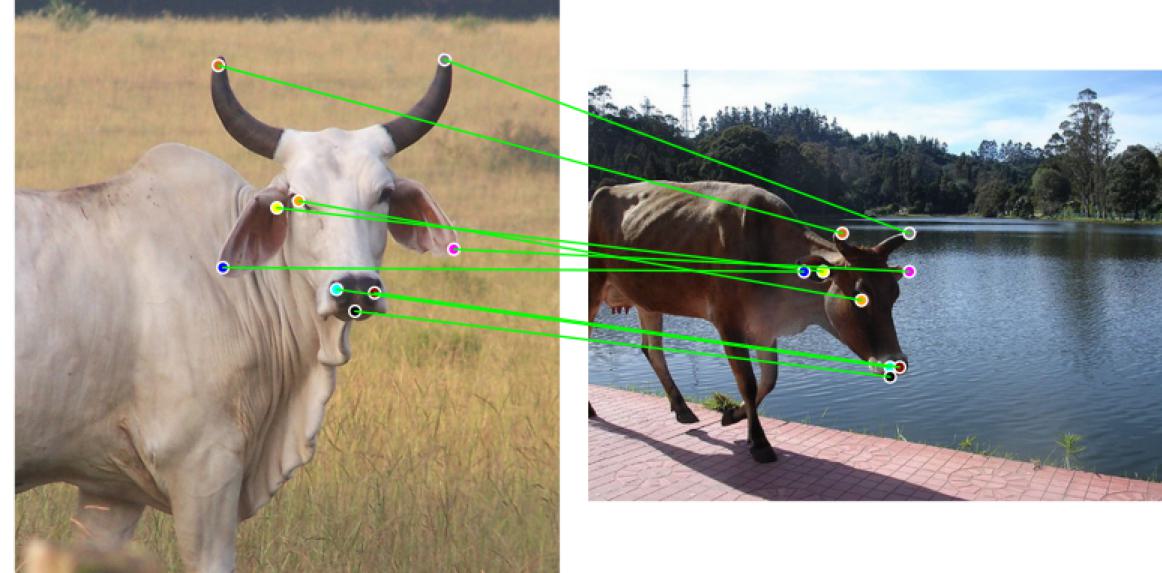} & 
    \includegraphics[width=0.22\textwidth]{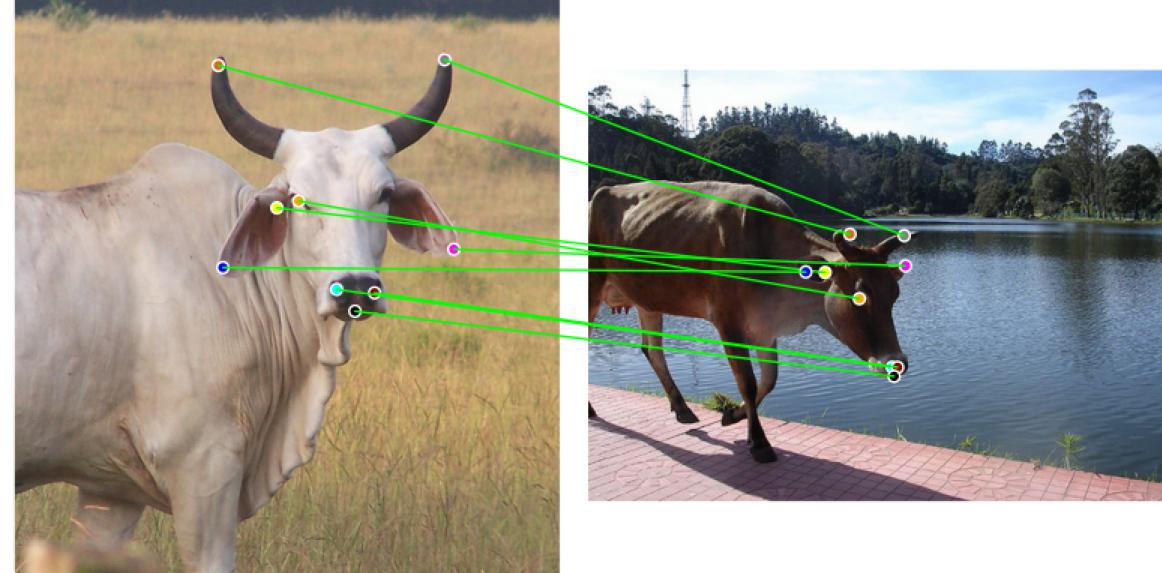} \\
    \includegraphics[width=0.22\textwidth]{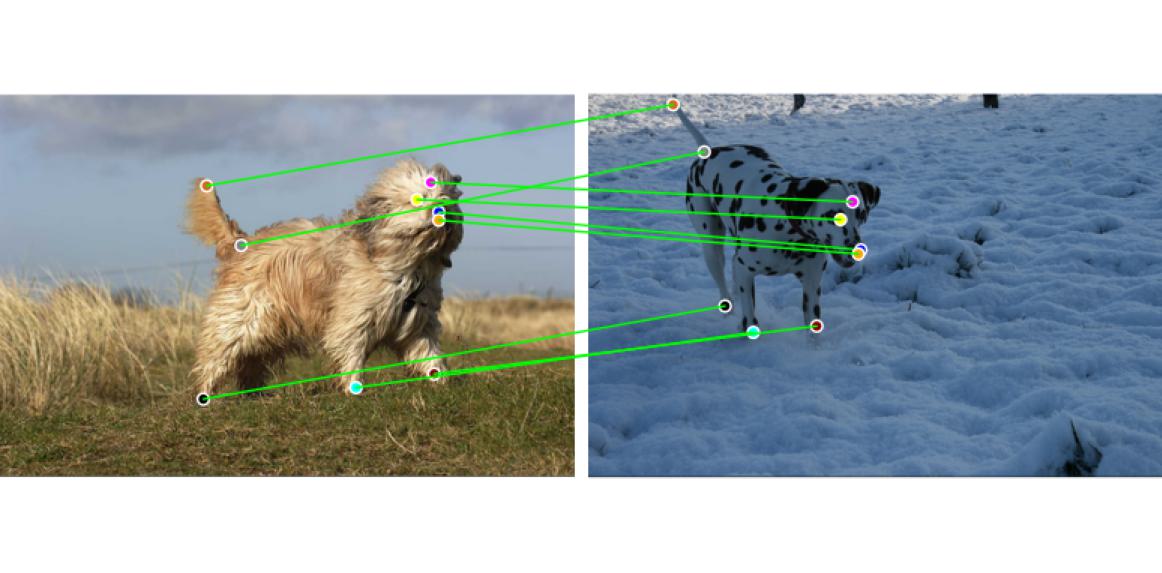} & 
    \includegraphics[width=0.22\textwidth]{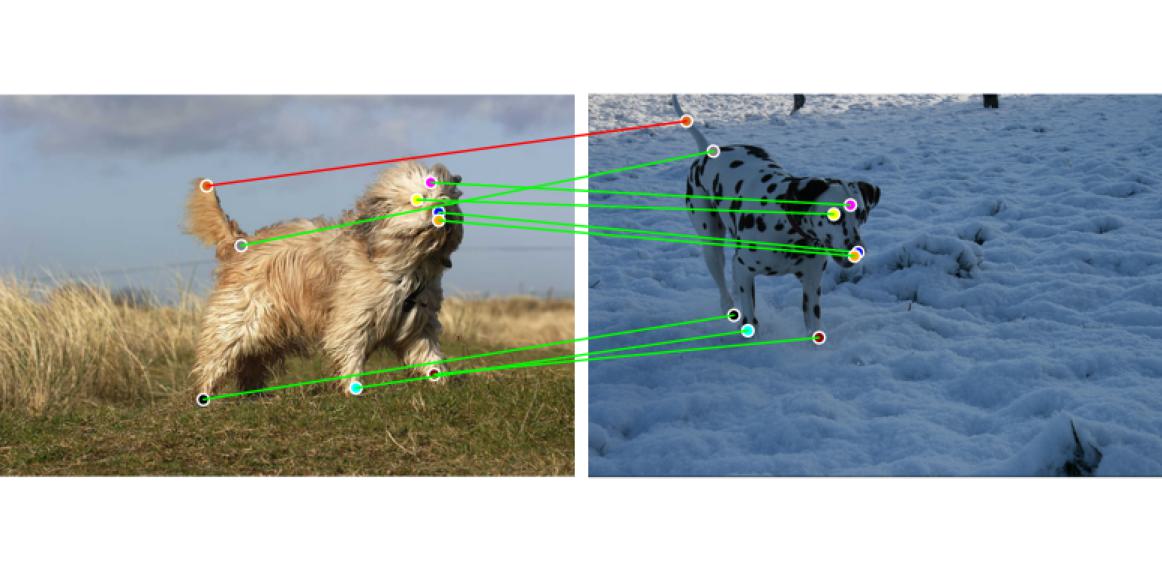} & 
    \includegraphics[width=0.22\textwidth]{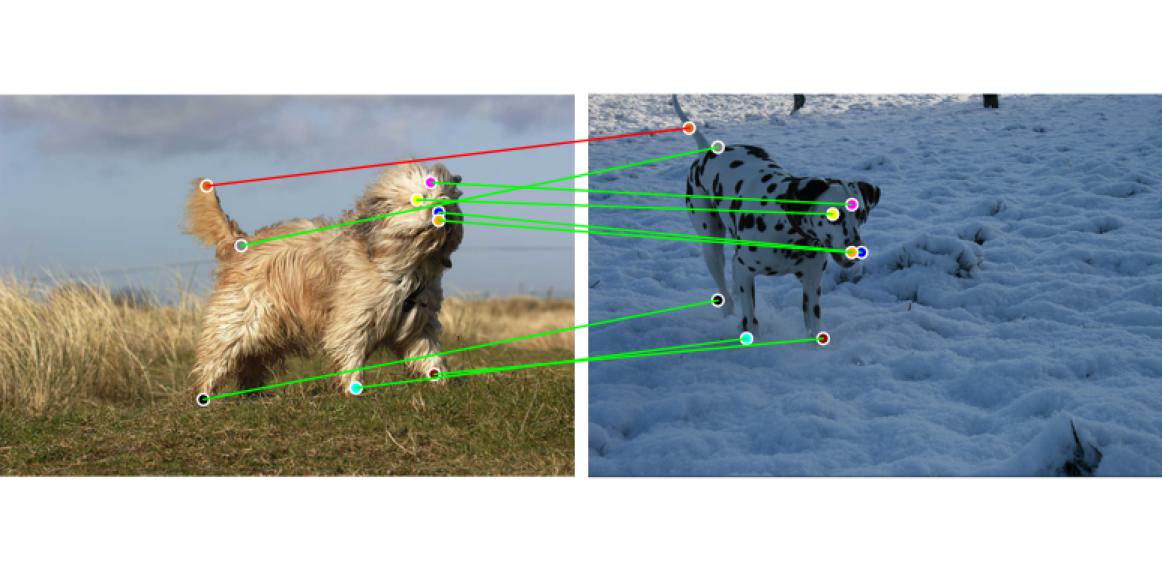} & 
    \includegraphics[width=0.22\textwidth]{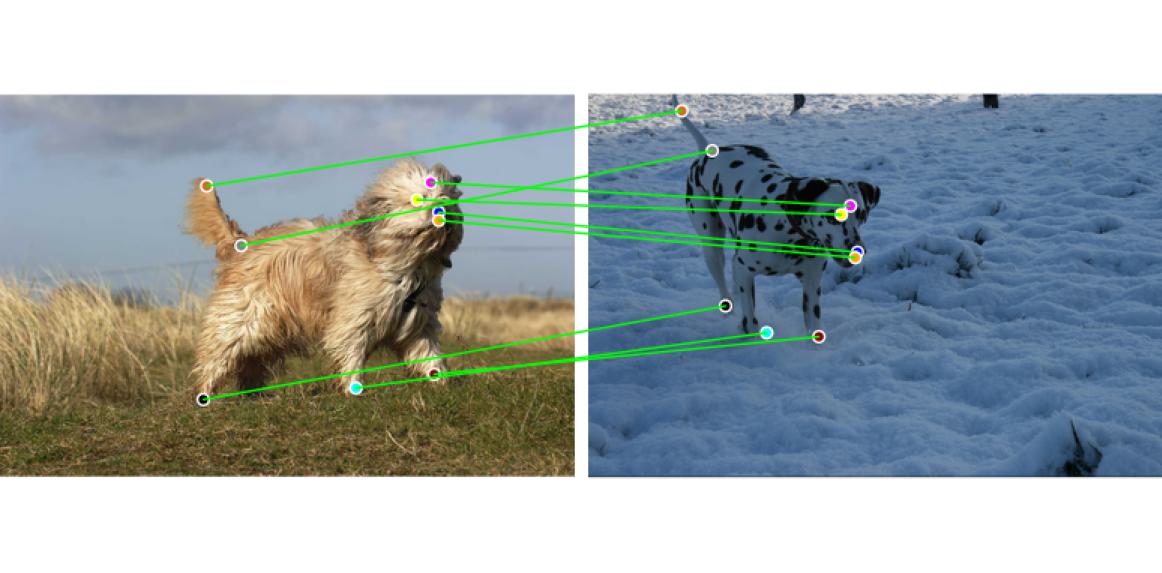} \\
    \includegraphics[width=0.22\textwidth]{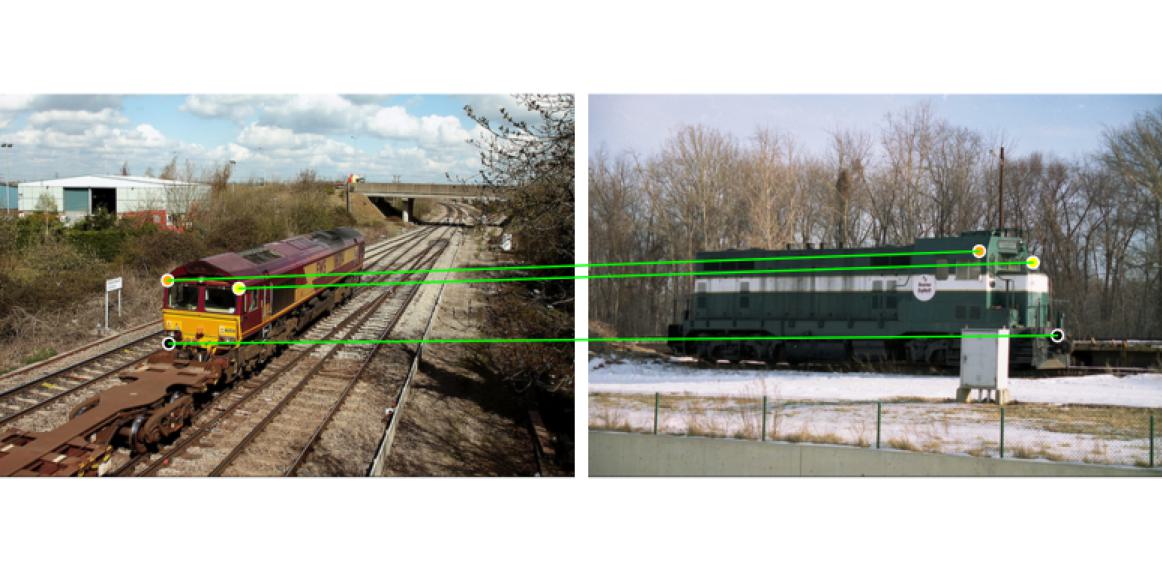} & 
    \includegraphics[width=0.22\textwidth]{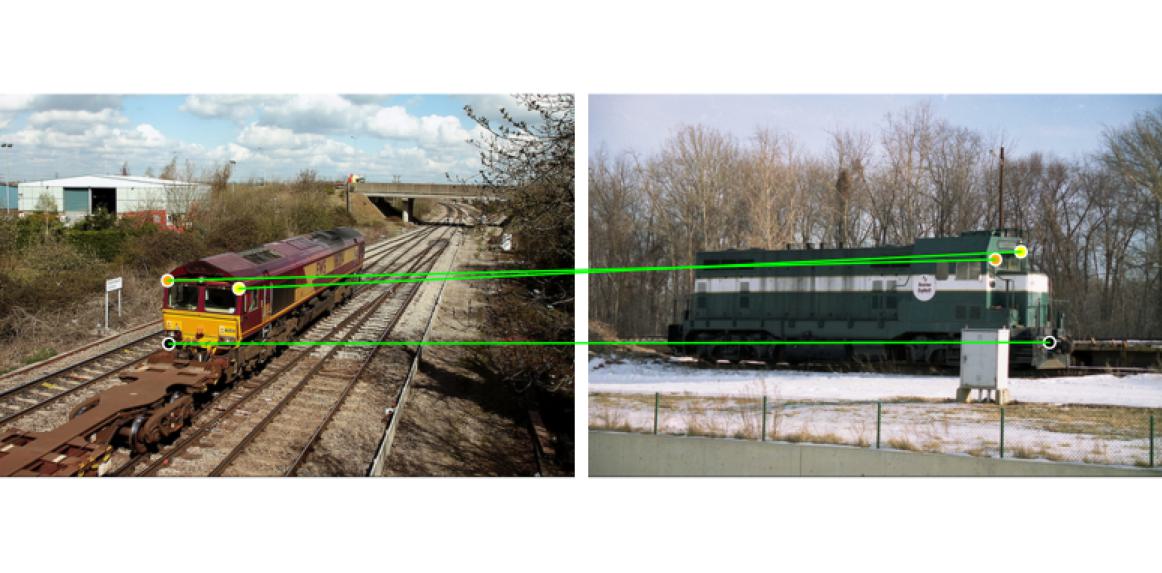} & 
    \includegraphics[width=0.22\textwidth]{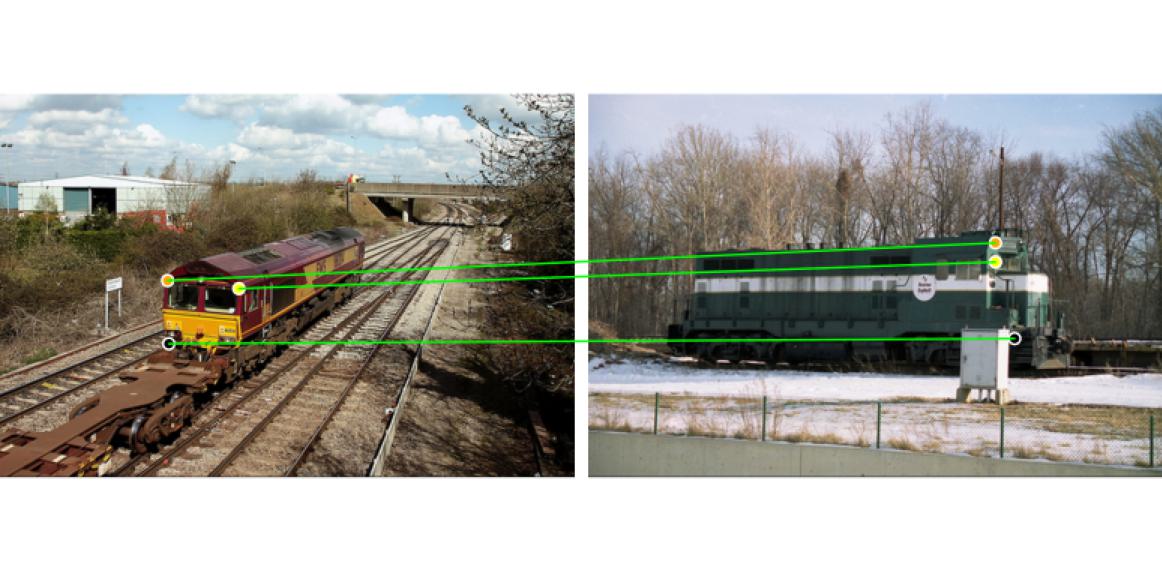} & 
    \includegraphics[width=0.22\textwidth]{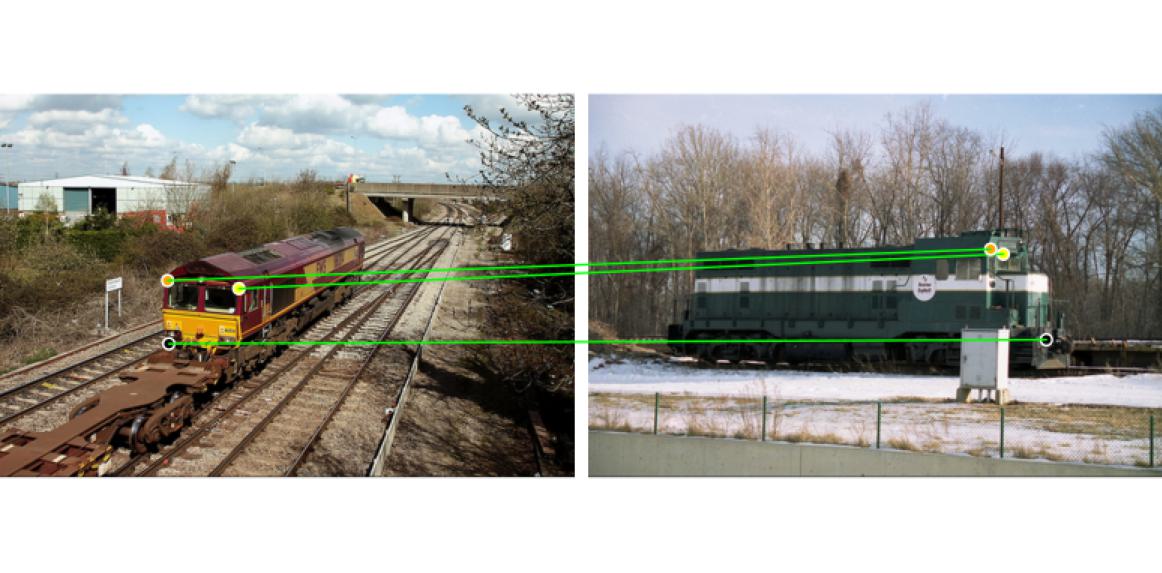} \\
    \includegraphics[width=0.22\textwidth]{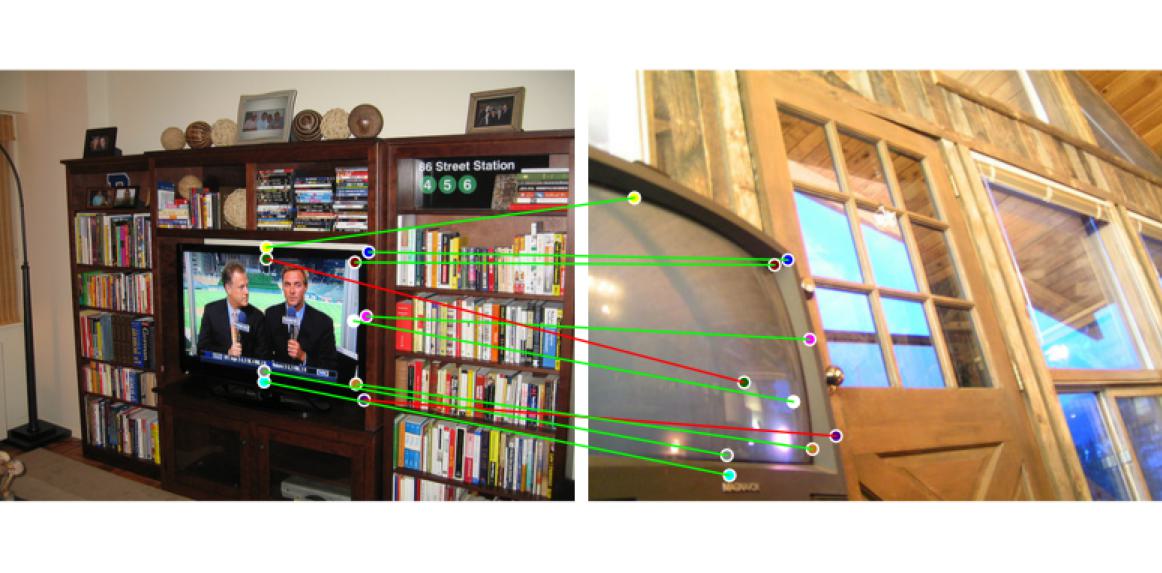} & 
    \includegraphics[width=0.22\textwidth]{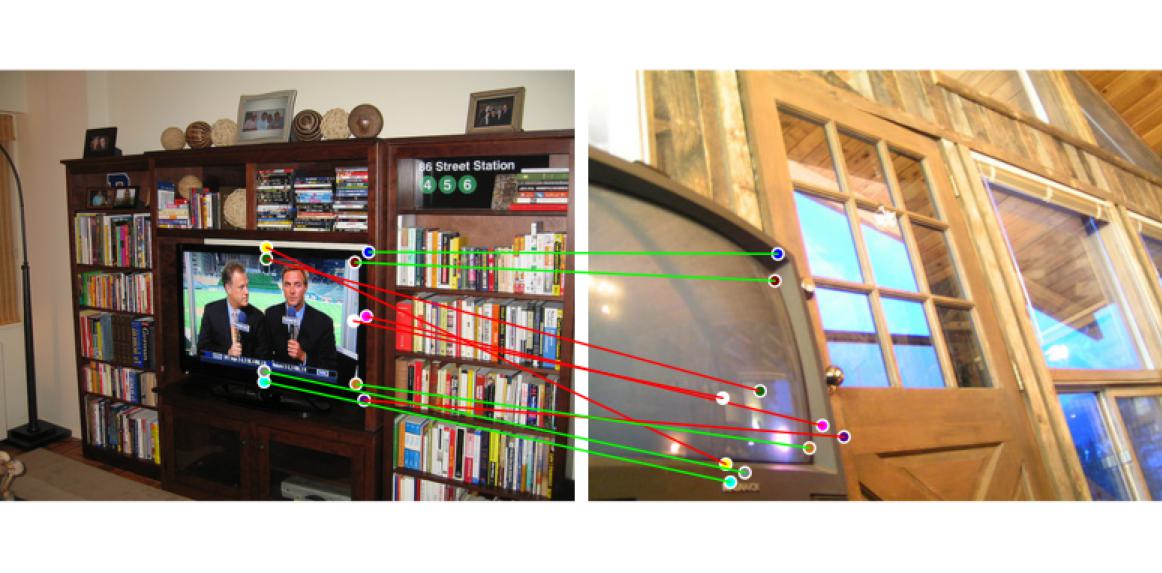} & 
    \includegraphics[width=0.22\textwidth]{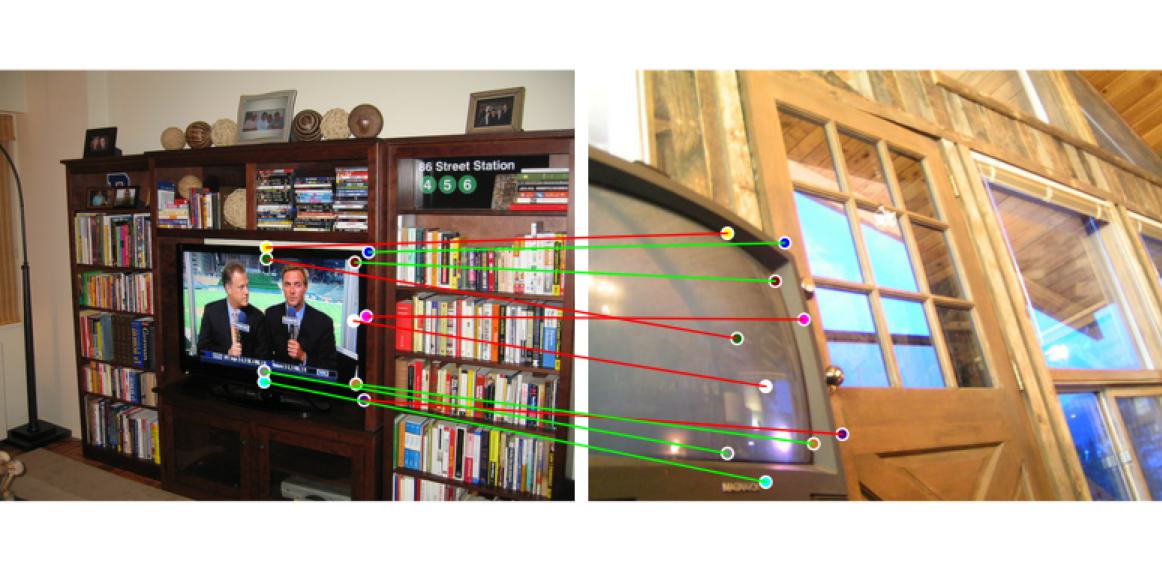} & 
    \includegraphics[width=0.22\textwidth]{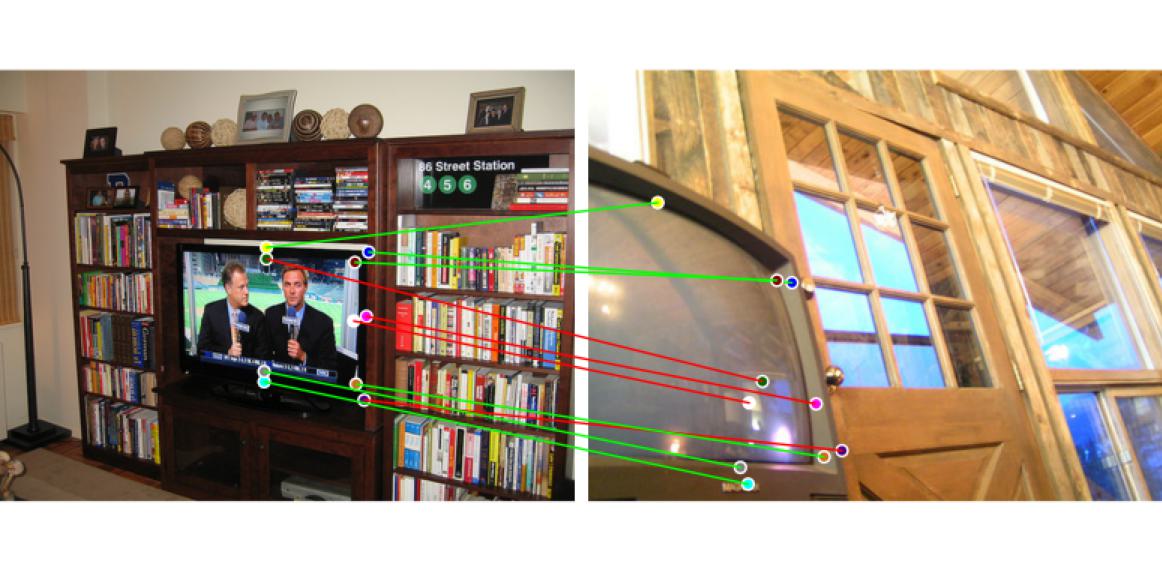} \\
  \end{tabular}
  \caption{Uncurated image pairs of SPair-71K dataset of three SOTA models and ours.}
  \label{fig:uncurated_matches_1}
\end{figure*}

\begin{figure*}[htbp]
  \centering
  \begin{tabular}{cccc}
    \multicolumn{1}{c}{DistillDIFT} & \multicolumn{1}{c}{TLR} & \multicolumn{1}{c}{SphMap} & \multicolumn{1}{c}{Ours} \\[5pt]
    % Bicycle row (bicycle__00931)
    \includegraphics[width=0.22\textwidth]{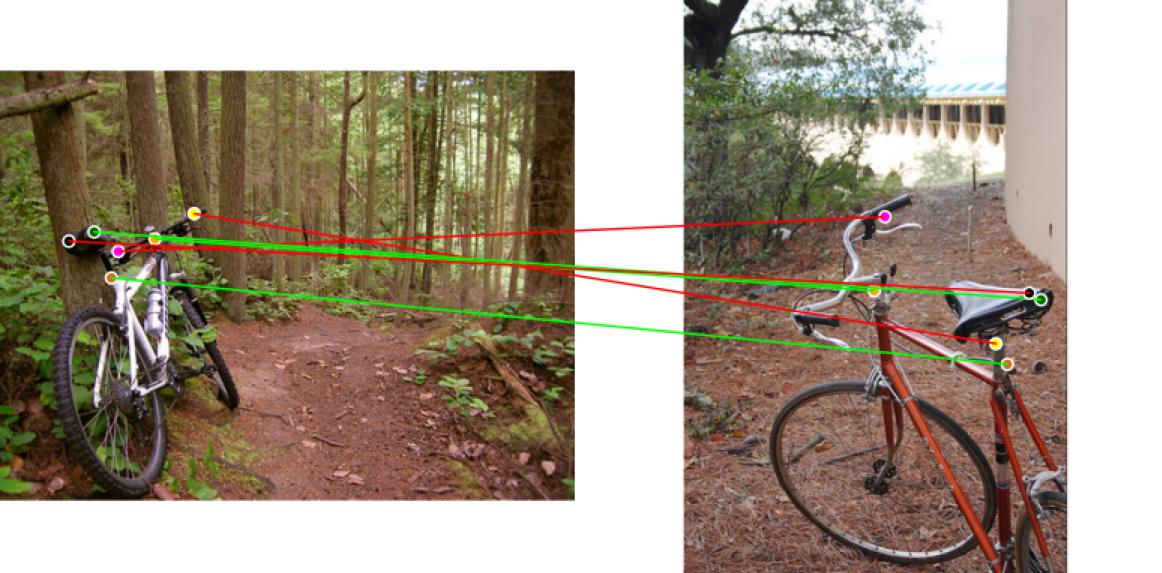} & 
    \includegraphics[width=0.22\textwidth]{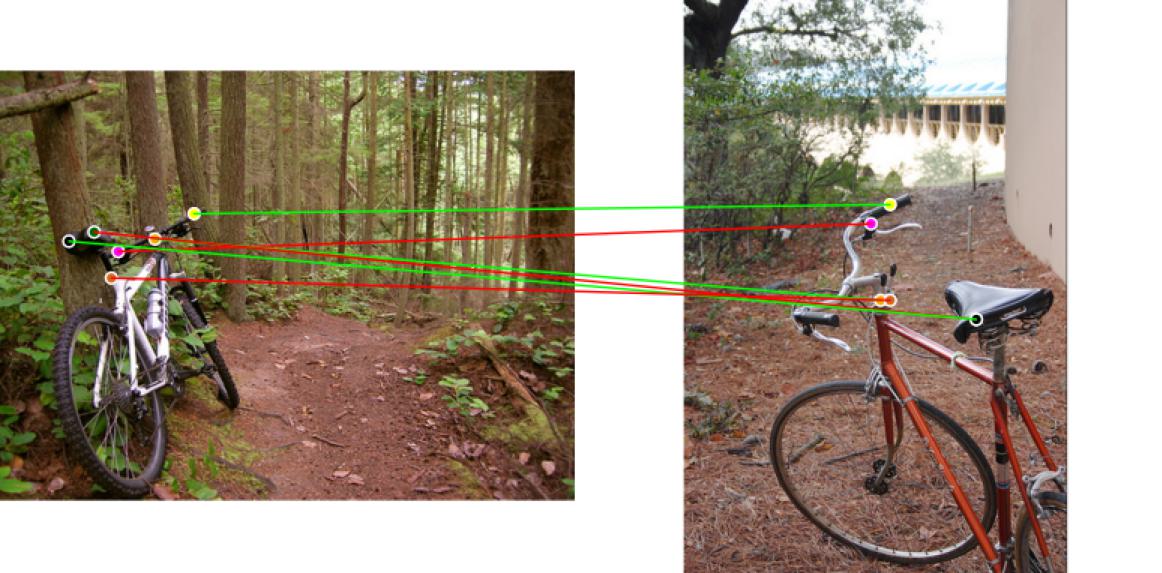} & 
    \includegraphics[width=0.22\textwidth]{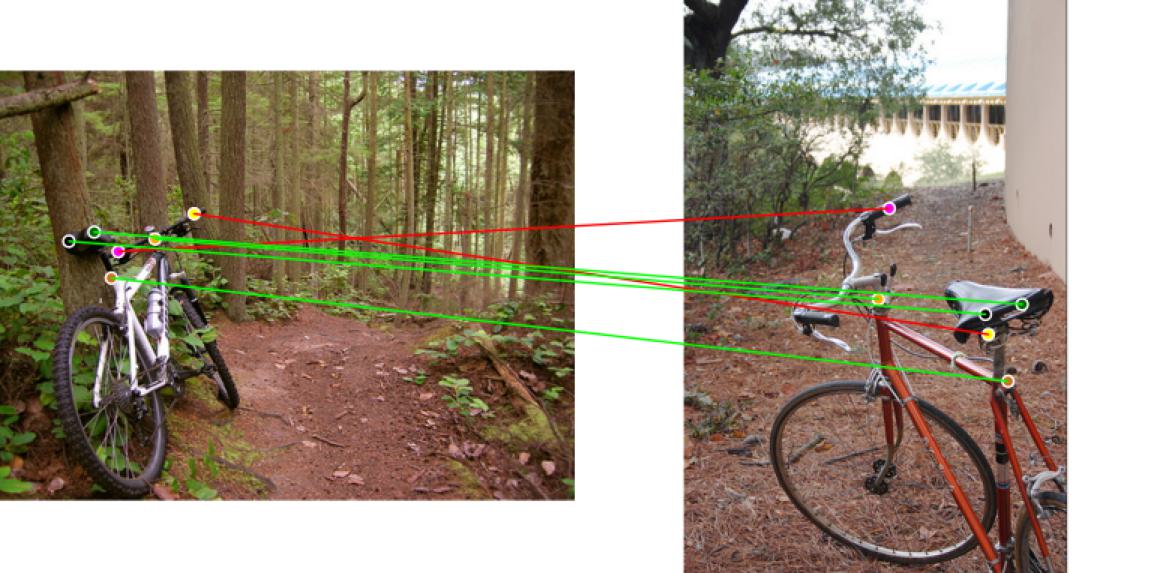} & 
    \includegraphics[width=0.22\textwidth]{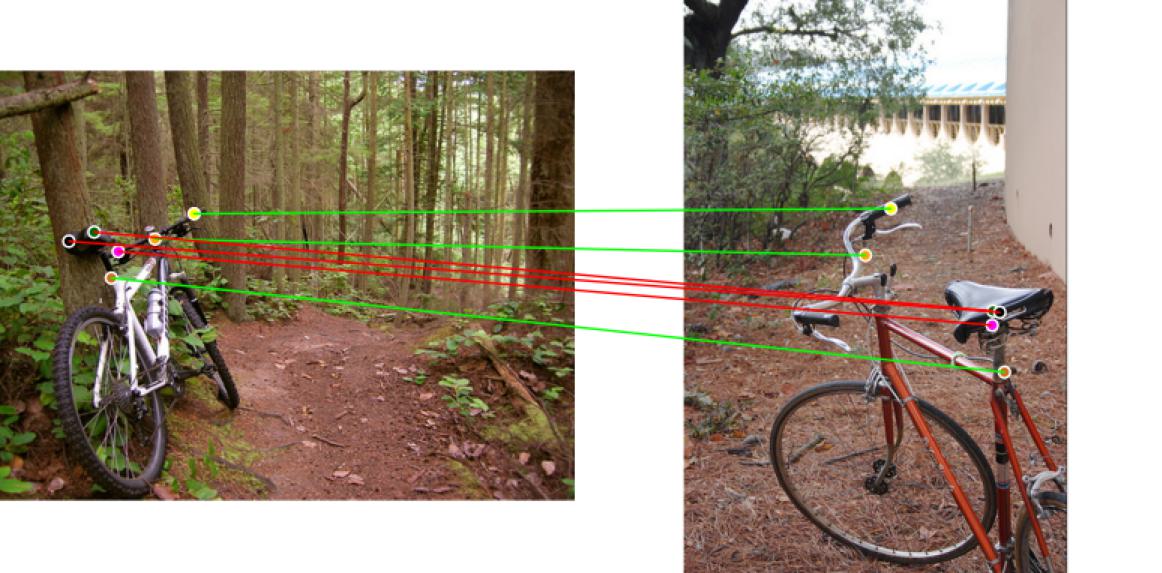} \\
    % Bird row (bird__01625)
    \includegraphics[width=0.22\textwidth]{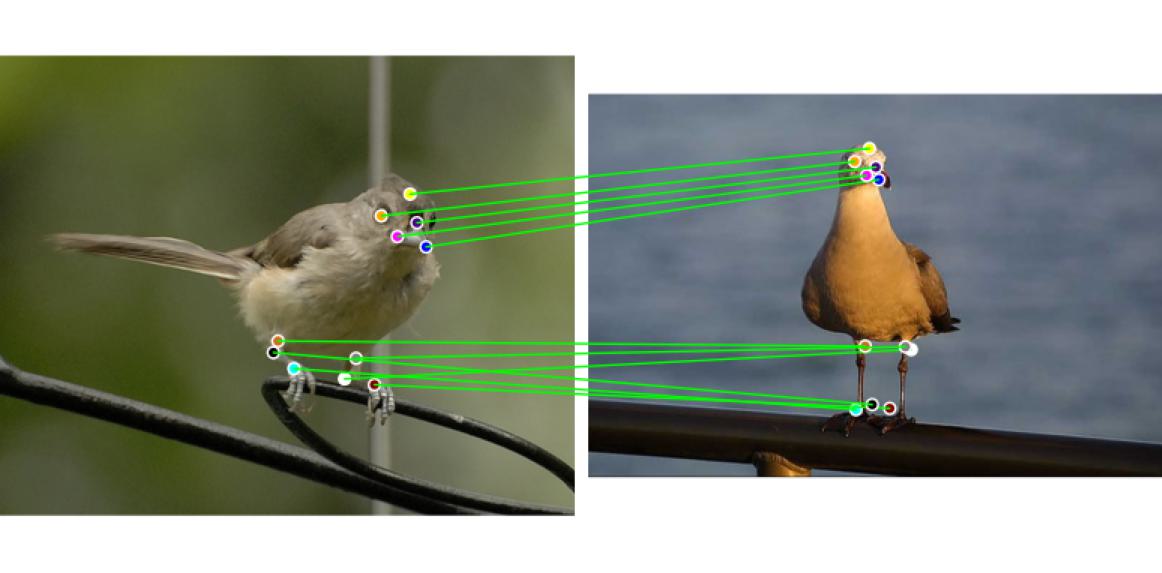} & 
    \includegraphics[width=0.22\textwidth]{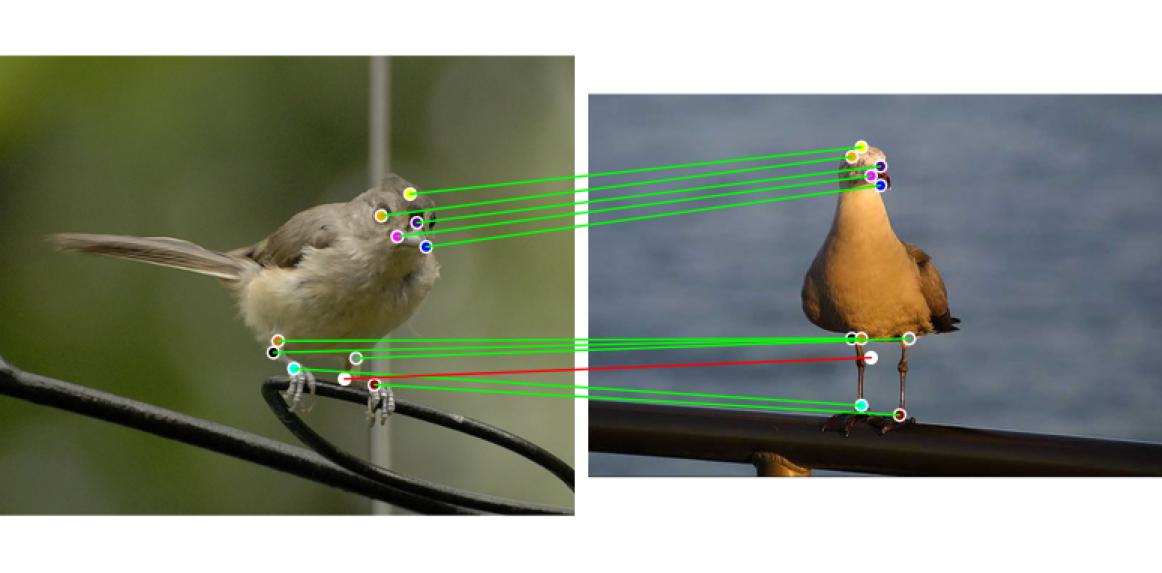} & 
    \includegraphics[width=0.22\textwidth]{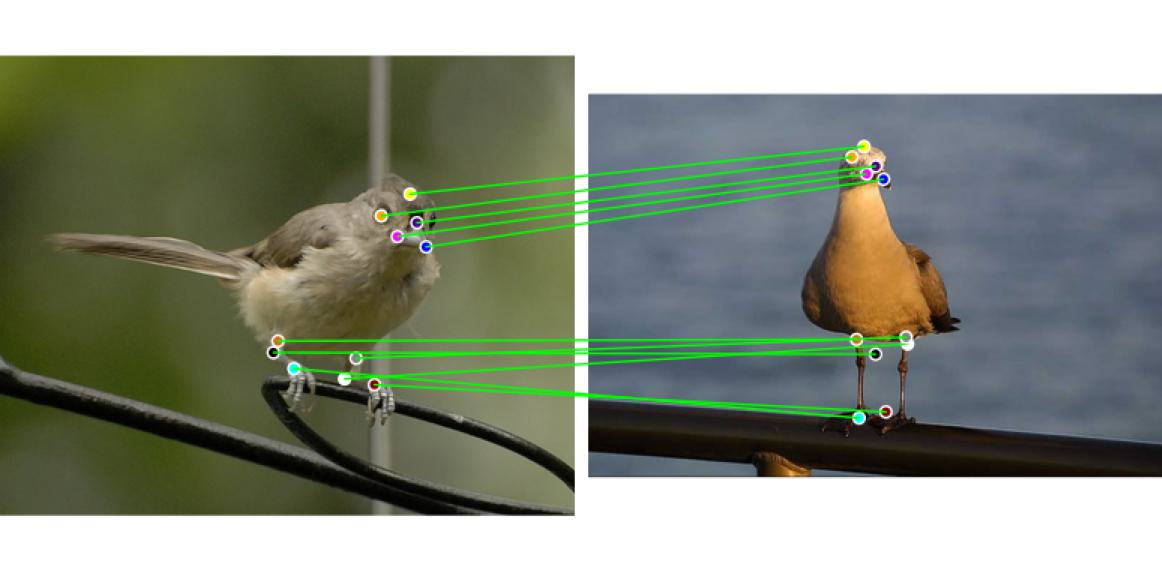} & 
    \includegraphics[width=0.22\textwidth]{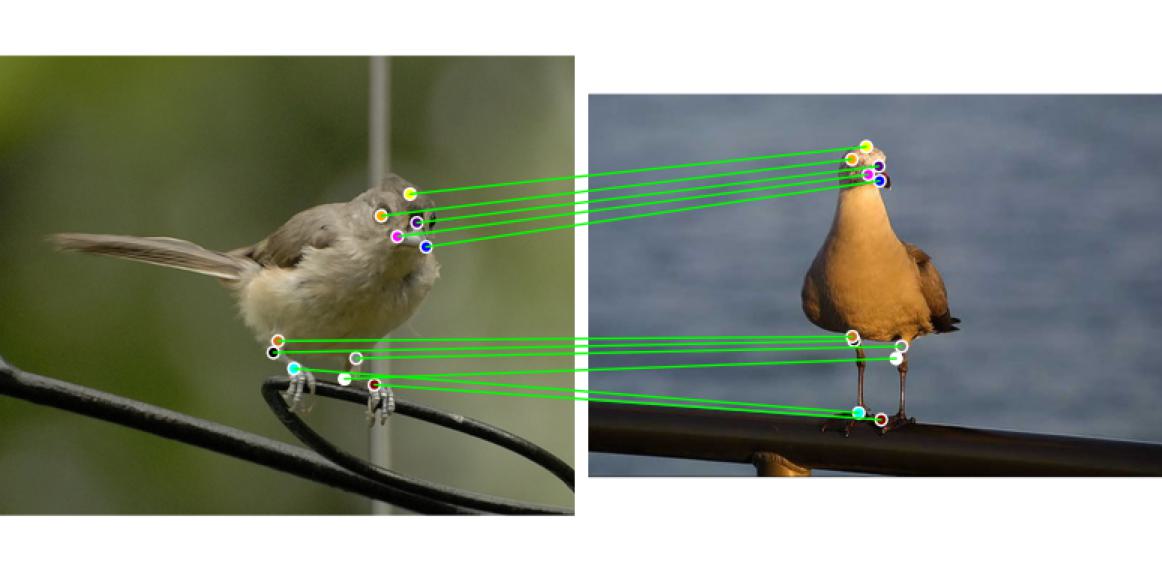} \\
    % Bird row (bird__01984)
    \includegraphics[width=0.22\textwidth]{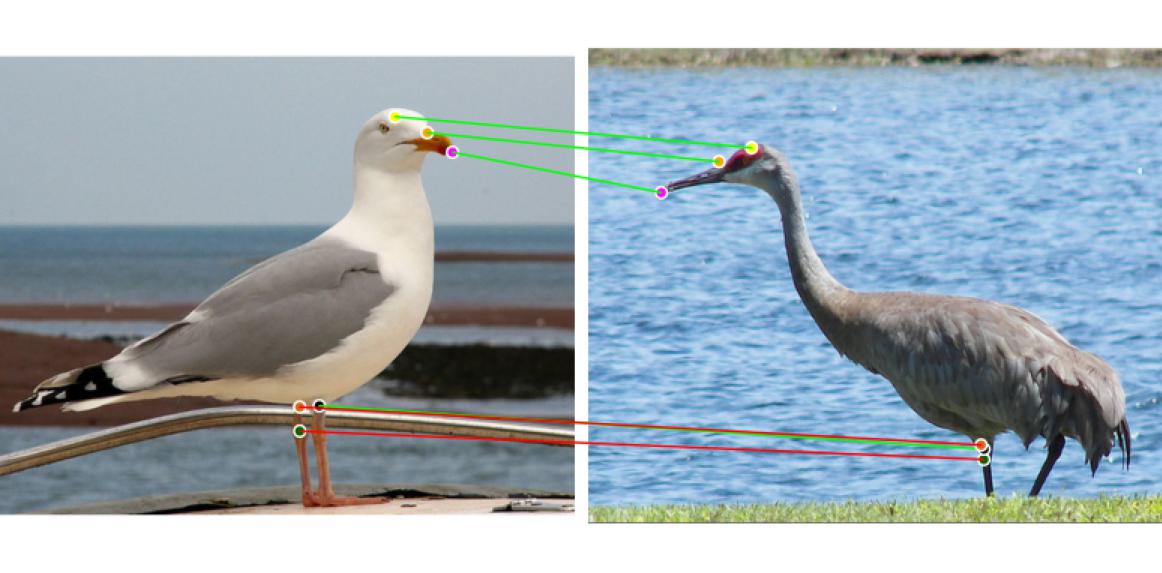} & 
    \includegraphics[width=0.22\textwidth]{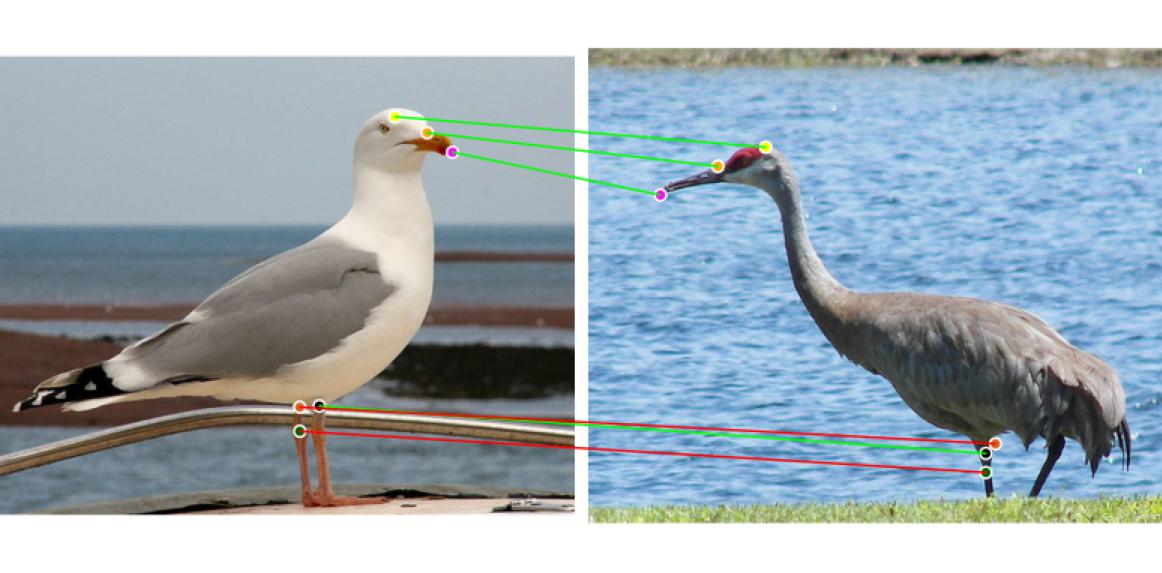} & 
    \includegraphics[width=0.22\textwidth]{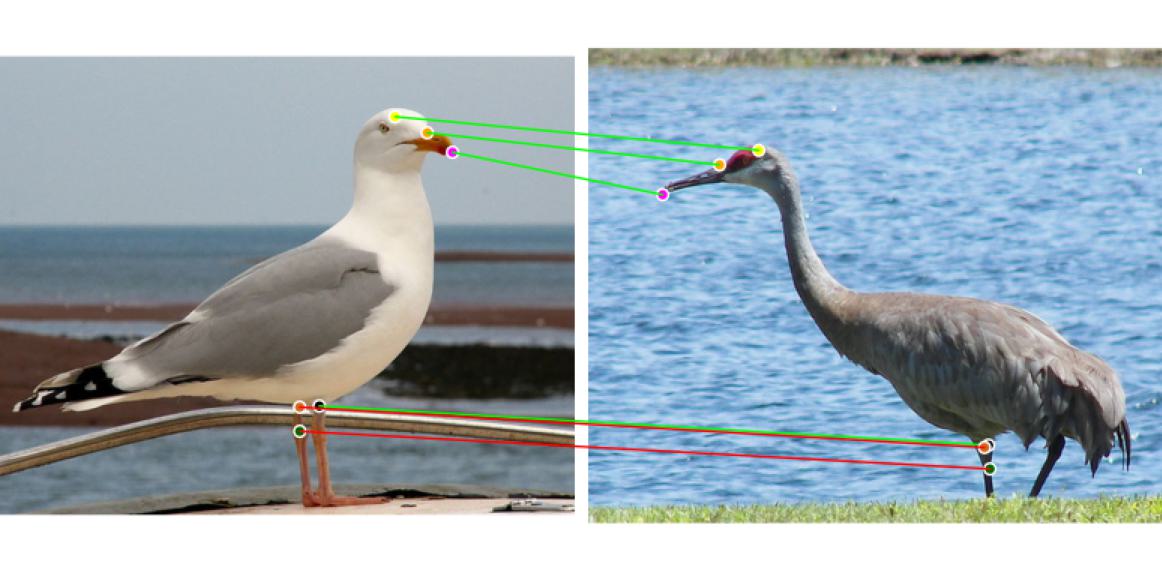} & 
    \includegraphics[width=0.22\textwidth]{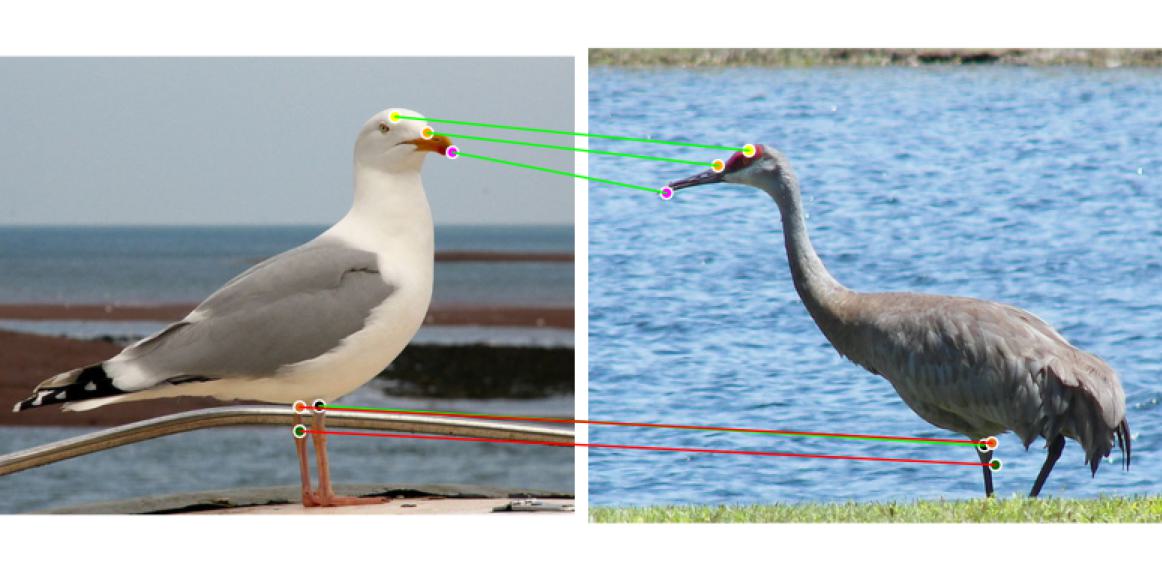} \\
    % Bus row (bus__04066)
    \includegraphics[width=0.22\textwidth]{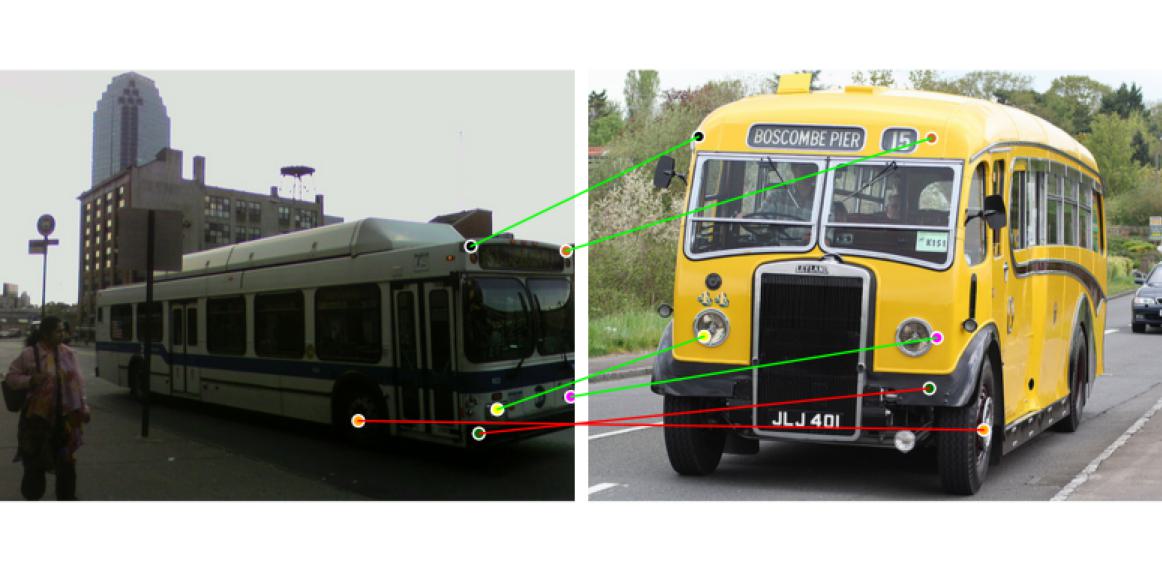} & 
    \includegraphics[width=0.22\textwidth]{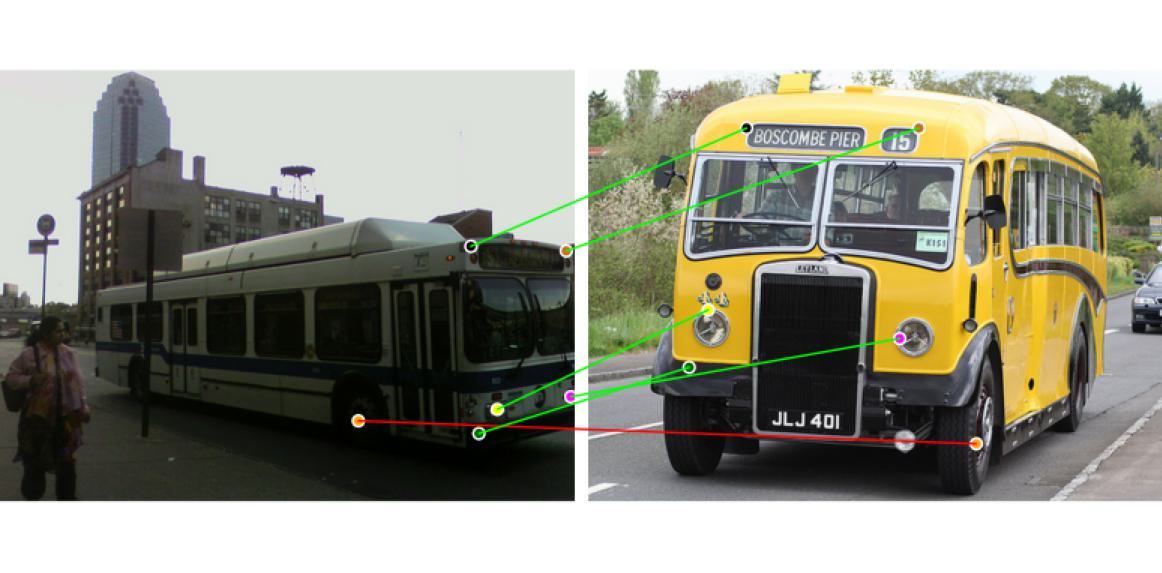} & 
    \includegraphics[width=0.22\textwidth]{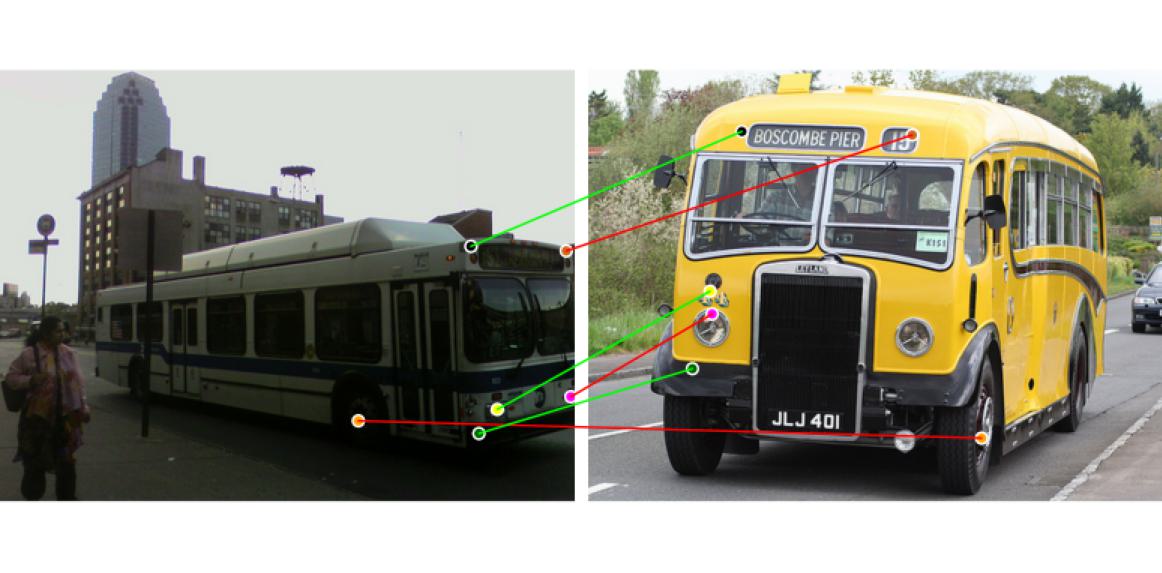} & 
    \includegraphics[width=0.22\textwidth]{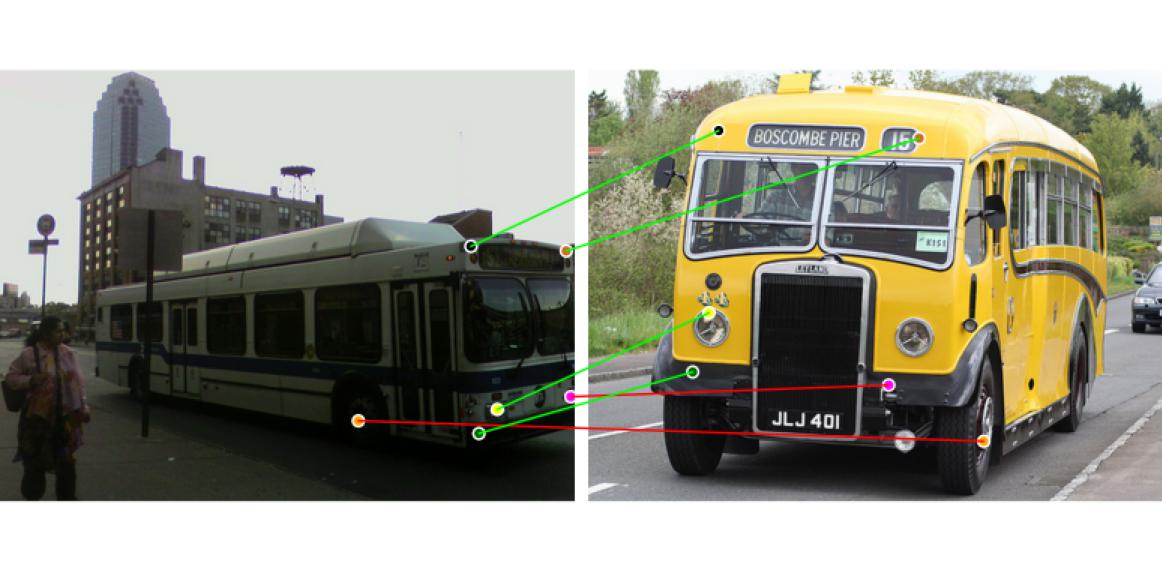} \\
    % Car row (car__04339)
    \includegraphics[width=0.22\textwidth]{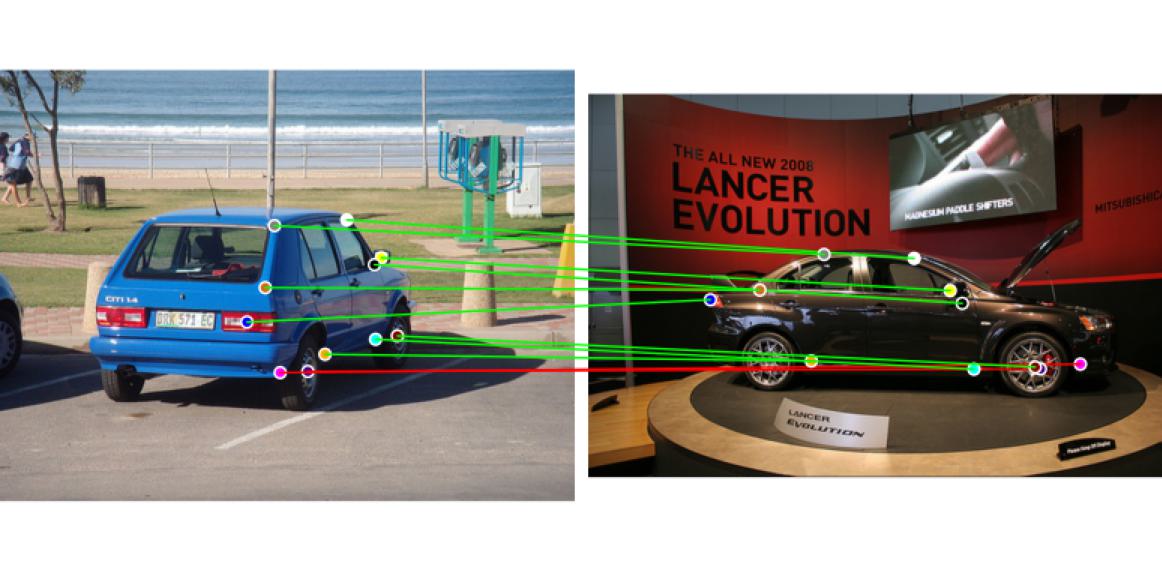} & 
    \includegraphics[width=0.22\textwidth]{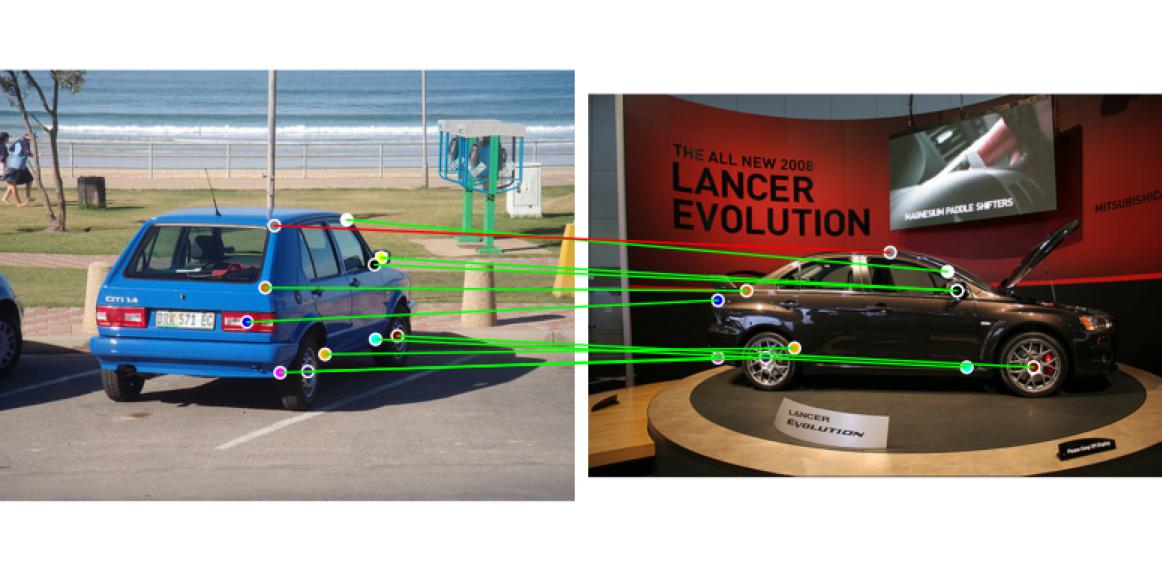} & 
    \includegraphics[width=0.22\textwidth]{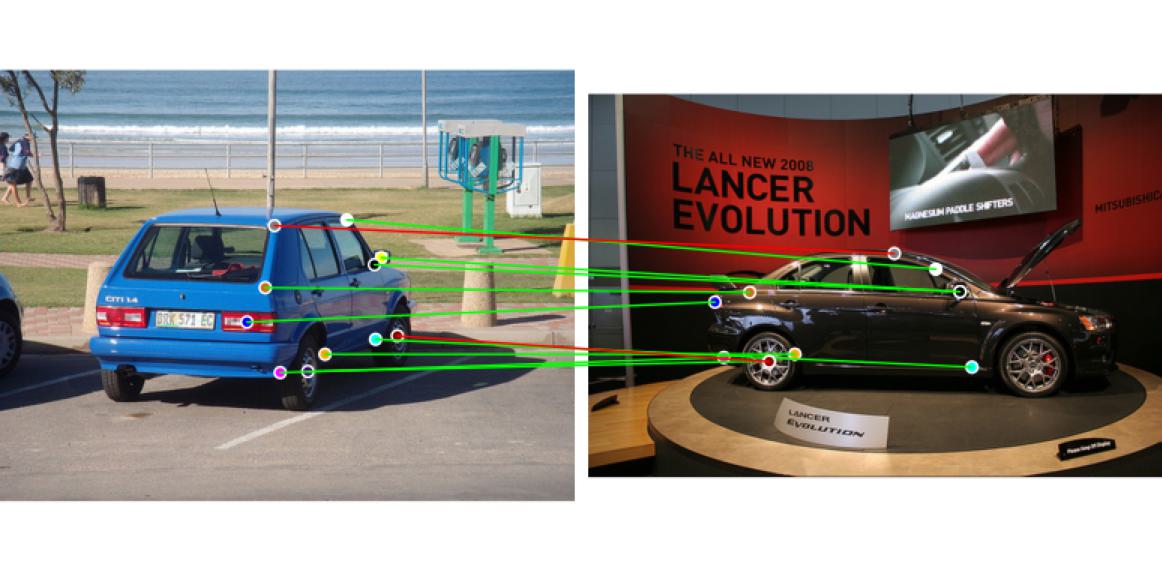} & 
    \includegraphics[width=0.22\textwidth]{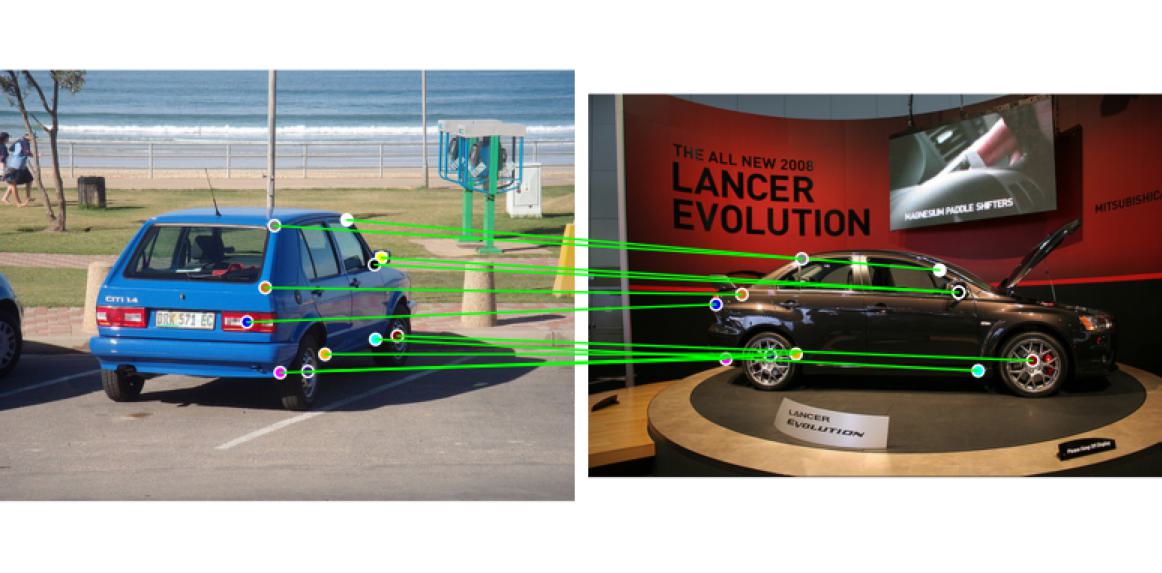} \\
    % Cat row (cat__05200)
    \includegraphics[width=0.22\textwidth]{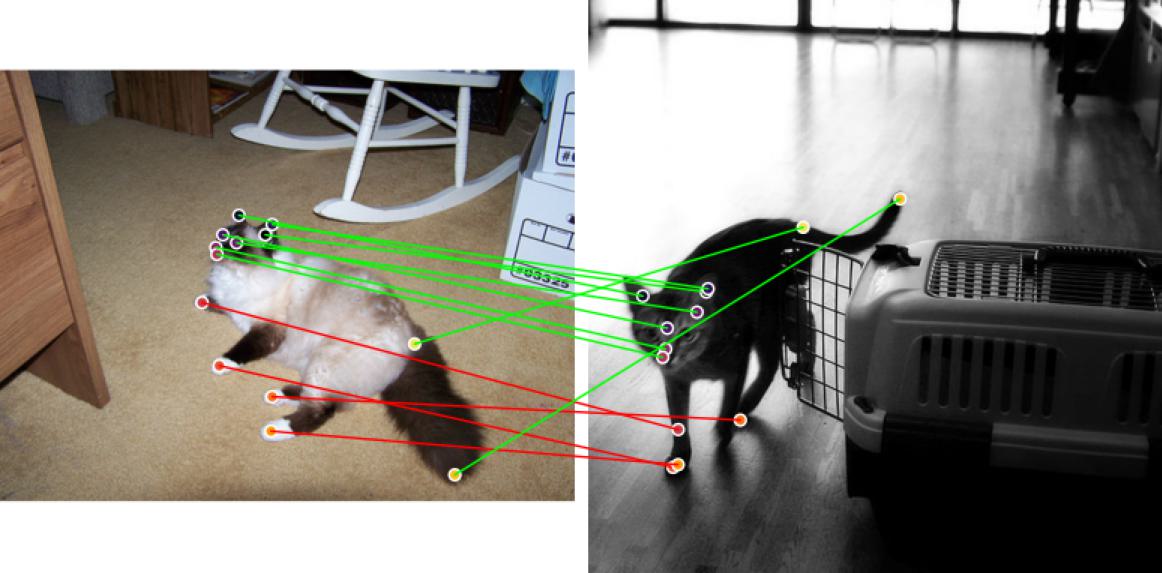} & 
    \includegraphics[width=0.22\textwidth]{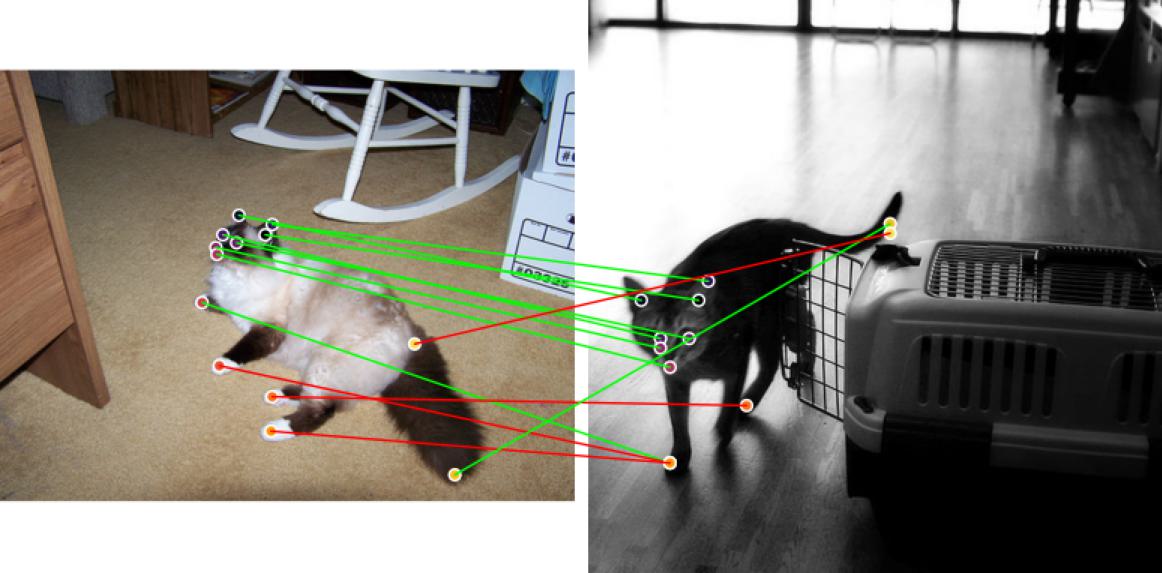} & 
    \includegraphics[width=0.22\textwidth]{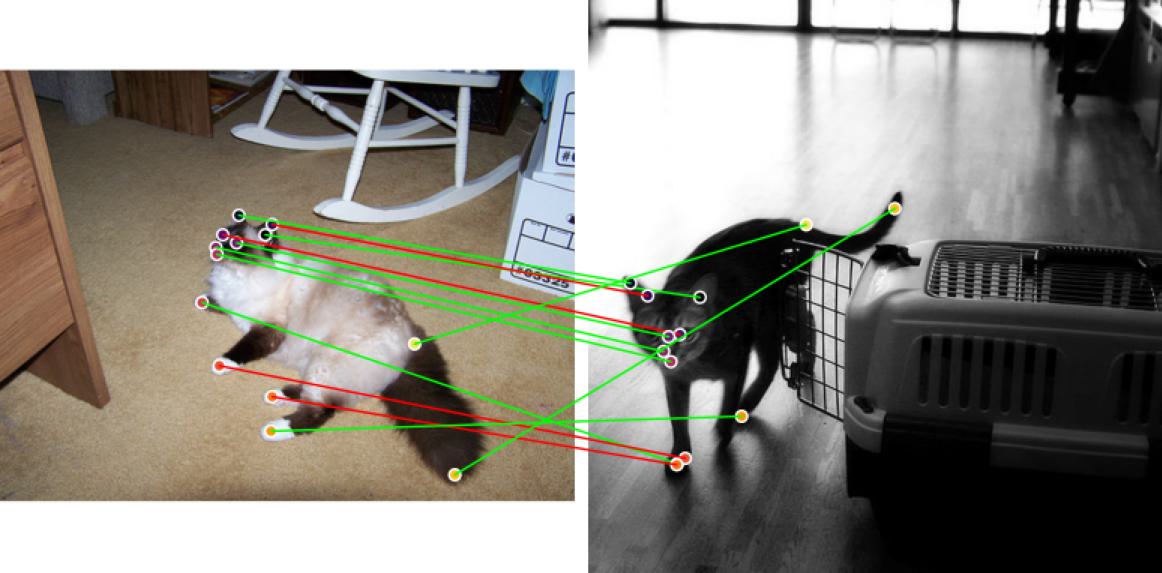} & 
    \includegraphics[width=0.22\textwidth]{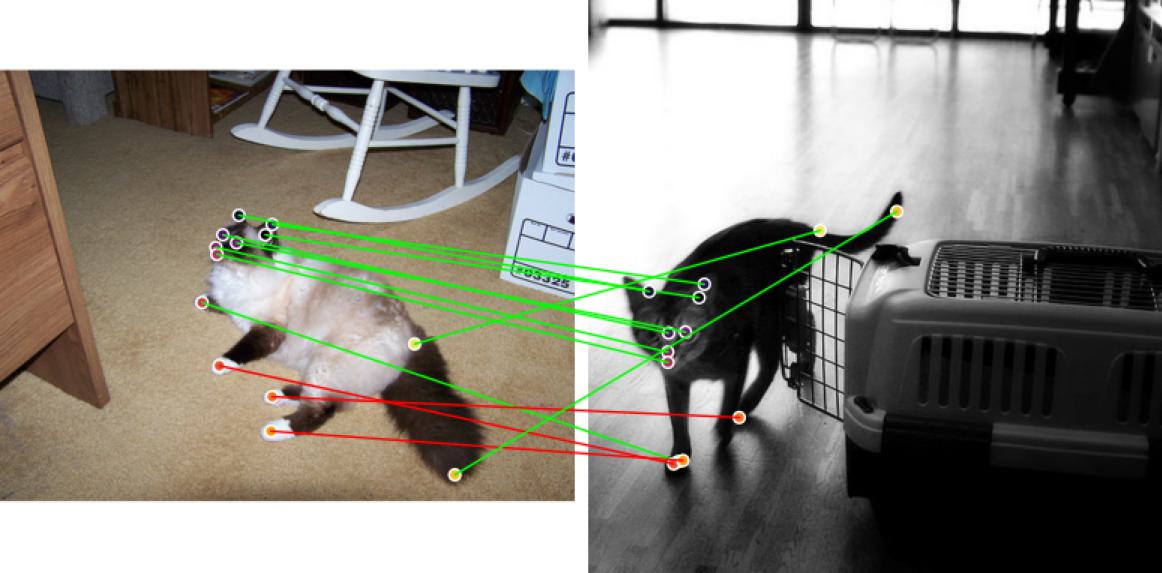} \\
    % Cat row (cat__05356)
    \includegraphics[width=0.22\textwidth]{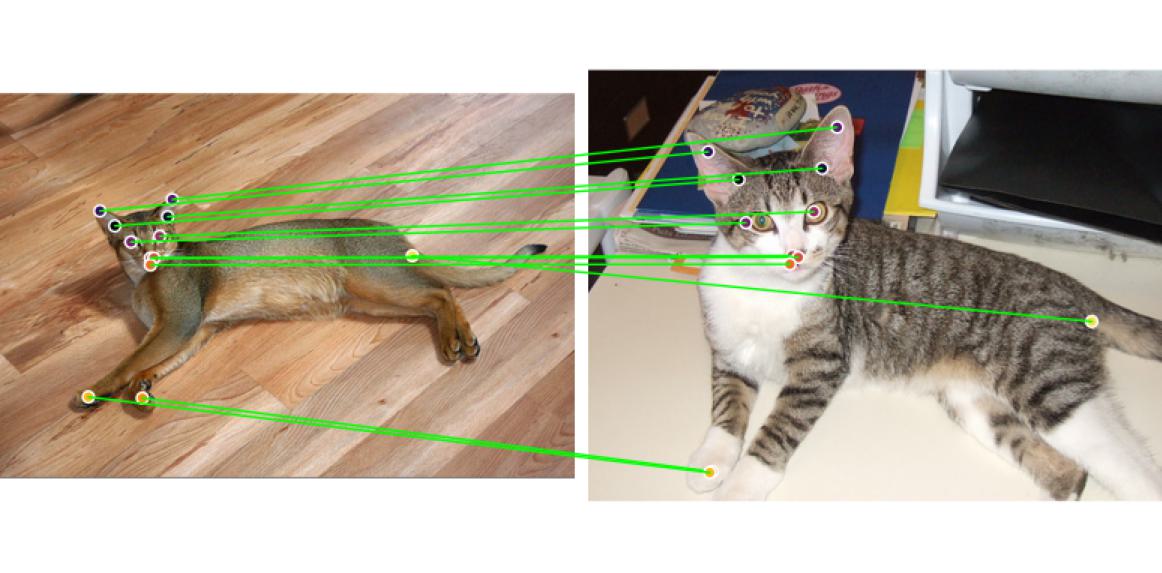} & 
    \includegraphics[width=0.22\textwidth]{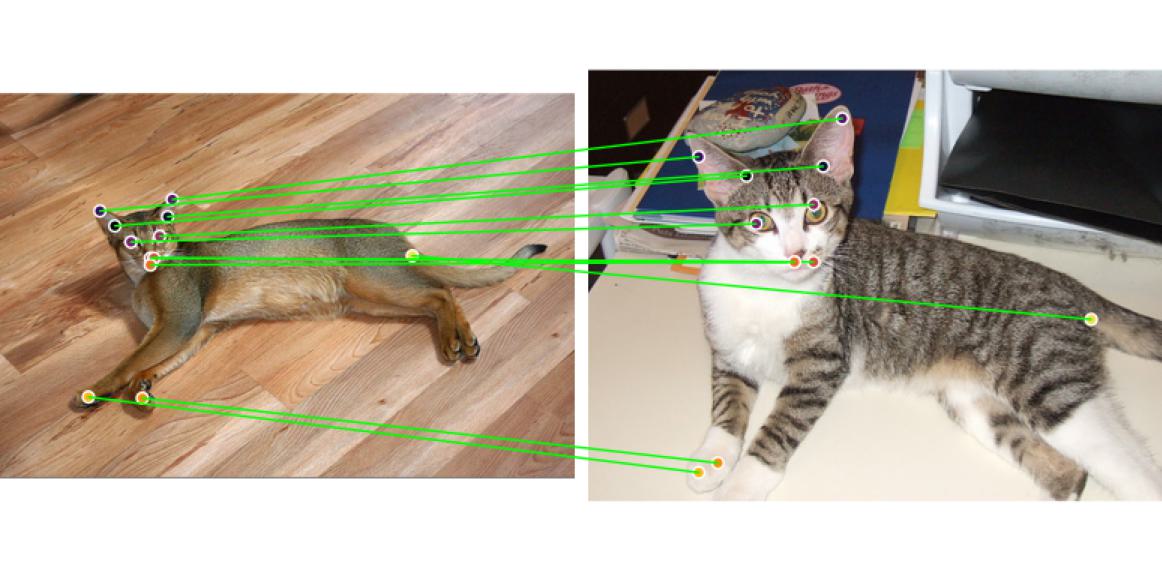} & 
    \includegraphics[width=0.22\textwidth]{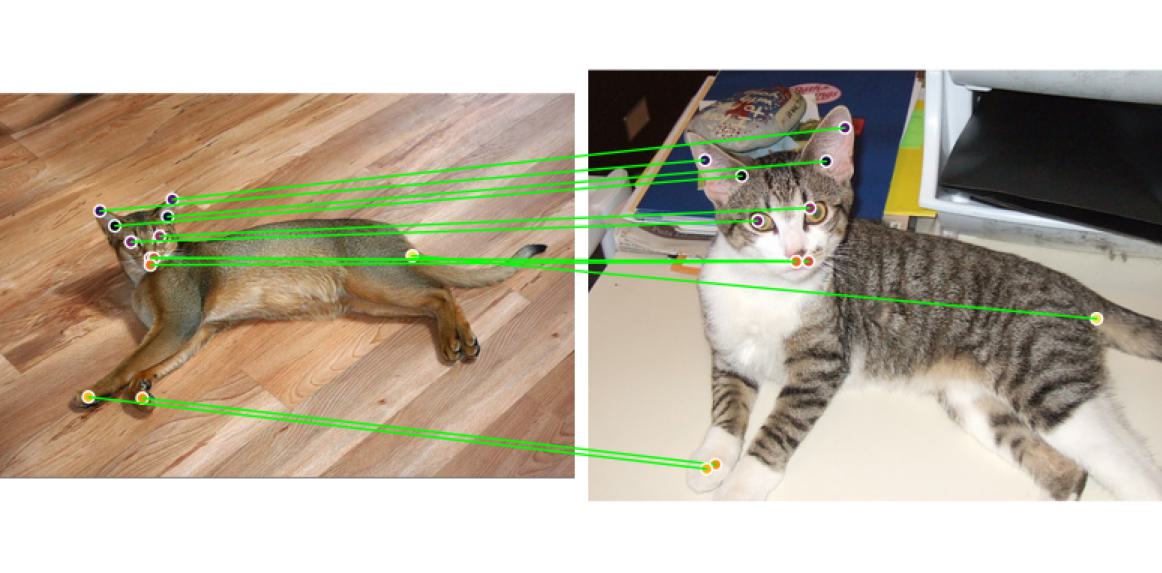} & 
    \includegraphics[width=0.22\textwidth]{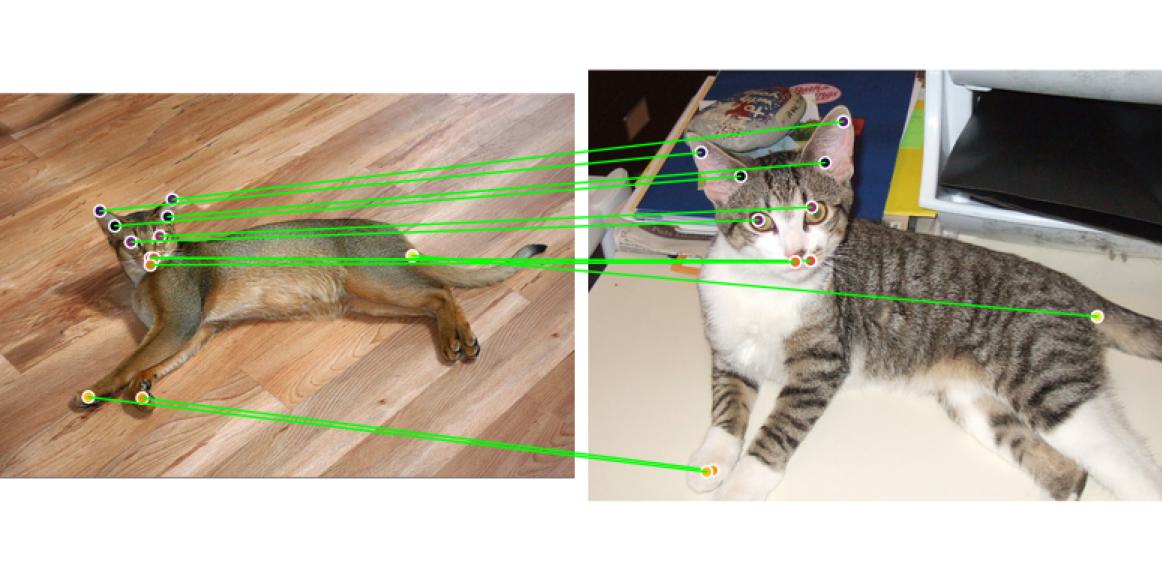} \\
    % Motorbike row (motorbike__08513)
    \includegraphics[width=0.22\textwidth]{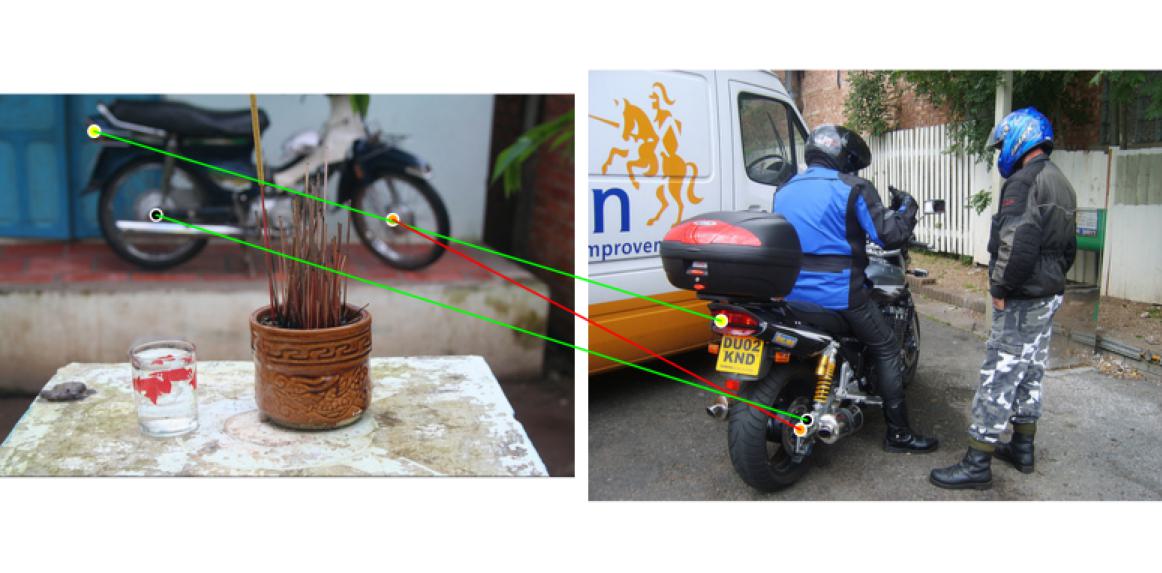} & 
    \includegraphics[width=0.22\textwidth]{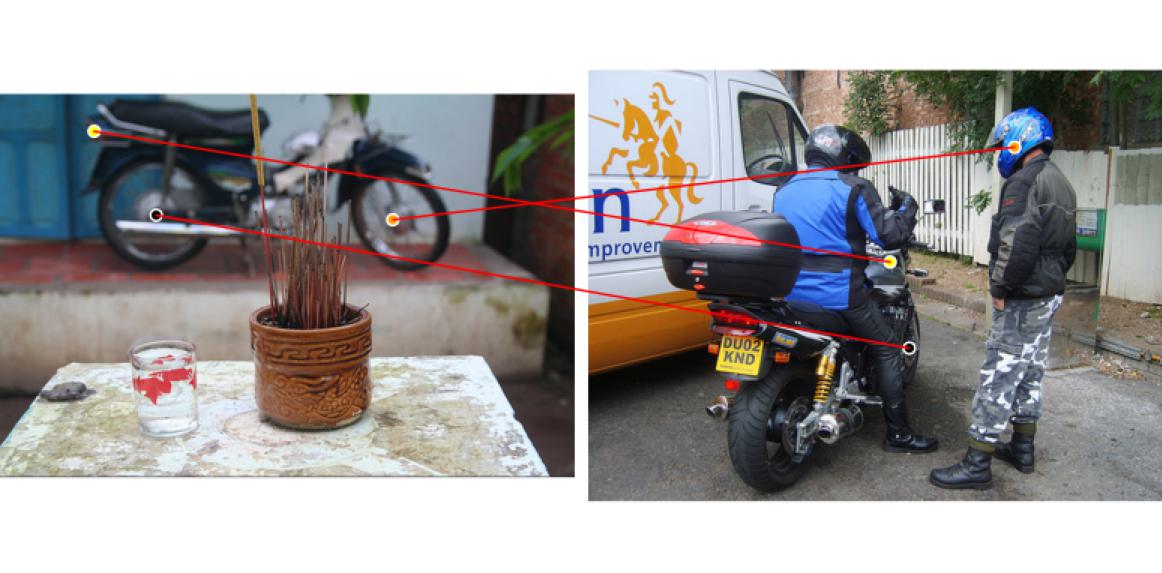} & 
    \includegraphics[width=0.22\textwidth]{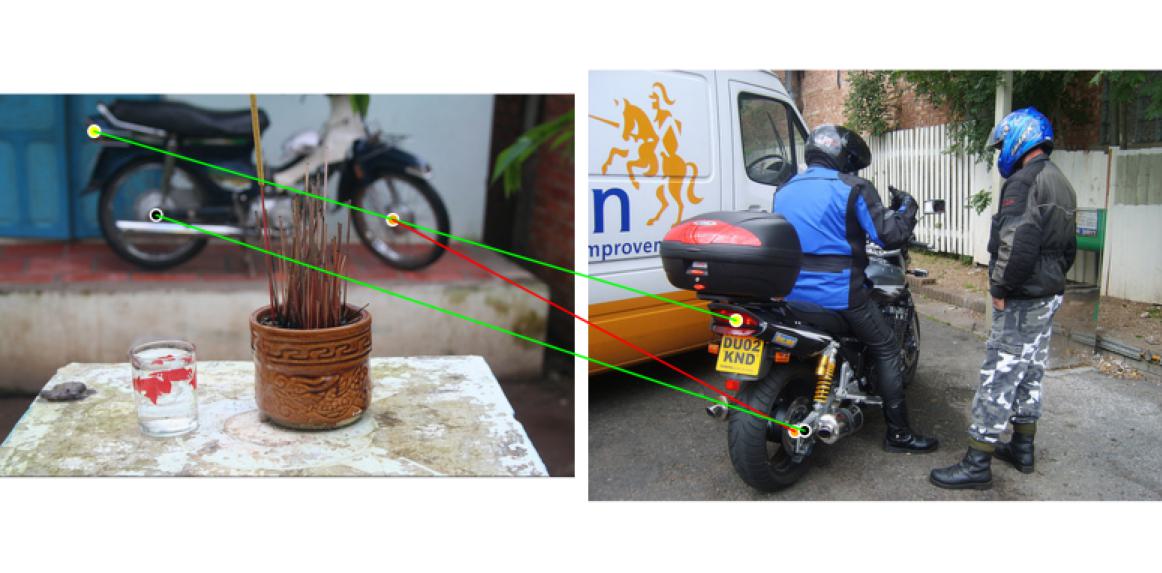} & 
    \includegraphics[width=0.22\textwidth]{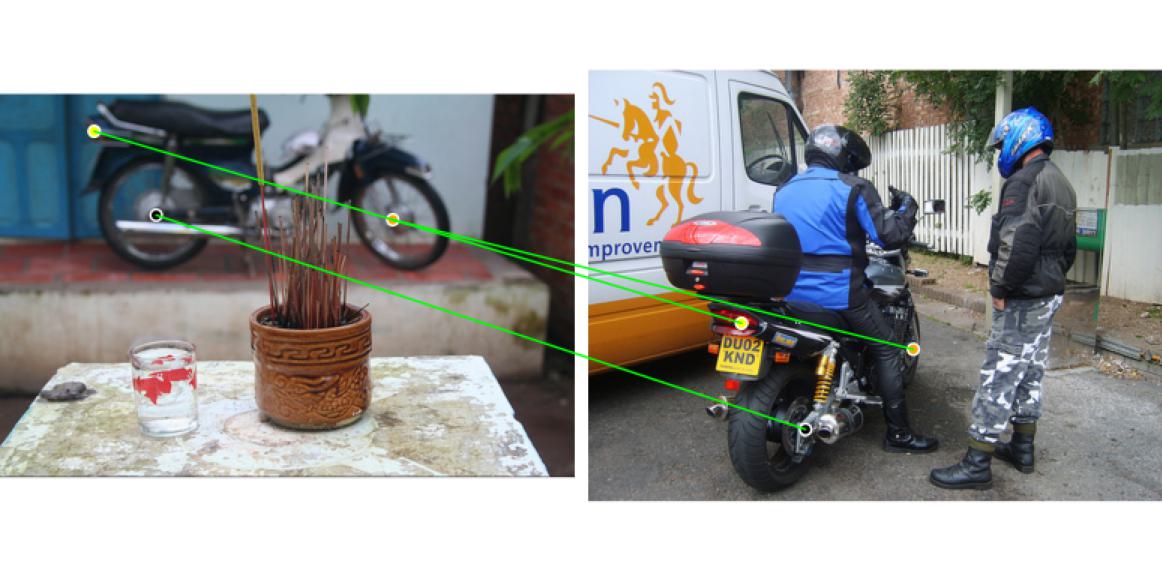} \\
    % Sheep row (sheep__10680)
    \includegraphics[width=0.22\textwidth]{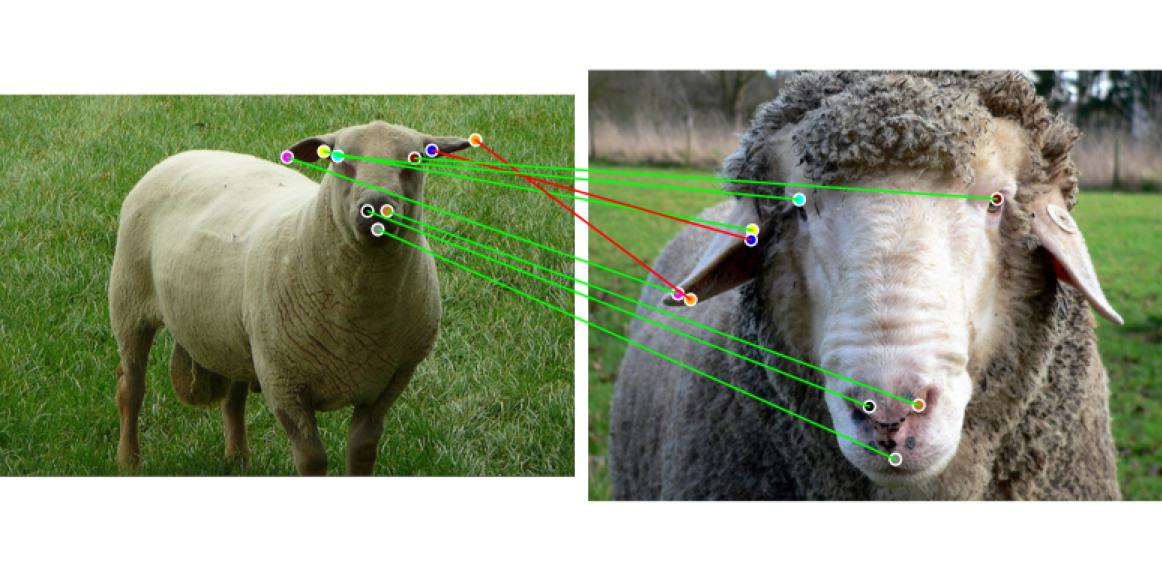} & 
    \includegraphics[width=0.22\textwidth]{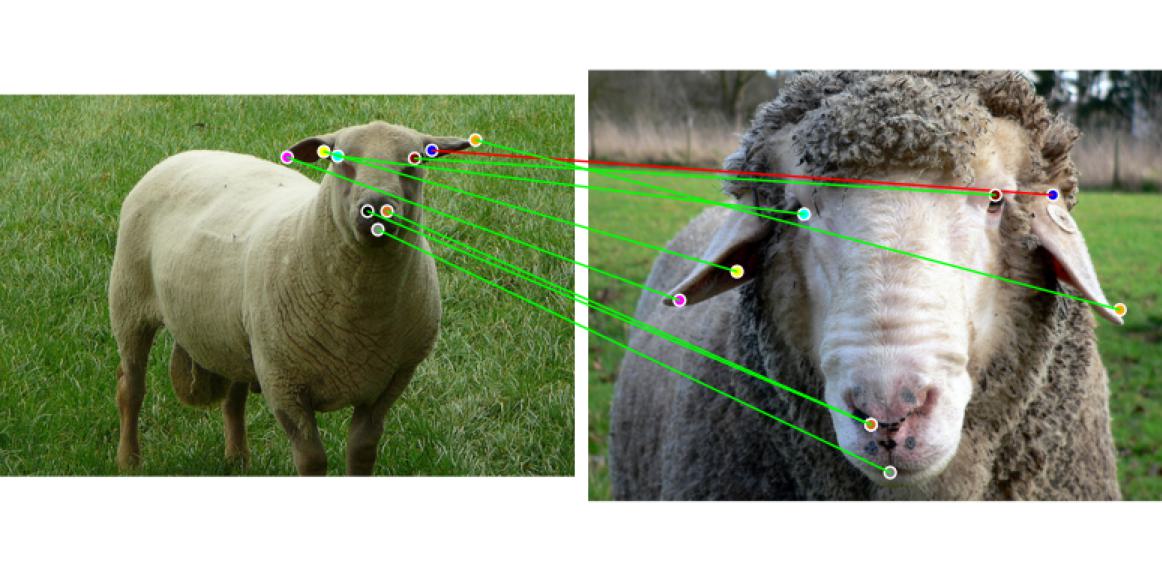} & 
    \includegraphics[width=0.22\textwidth]{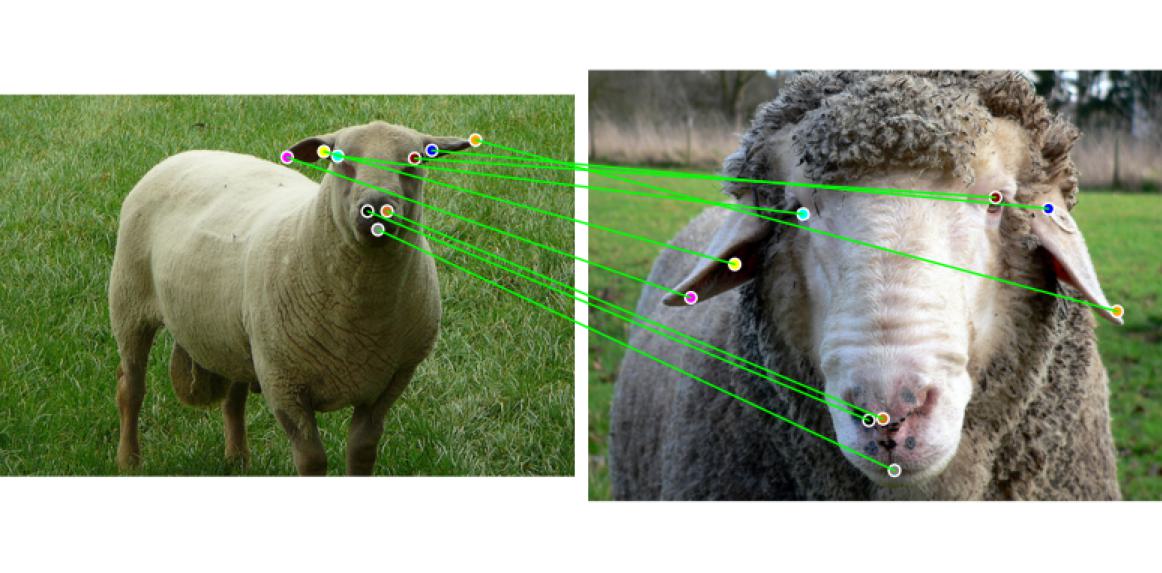} & 
    \includegraphics[width=0.22\textwidth]{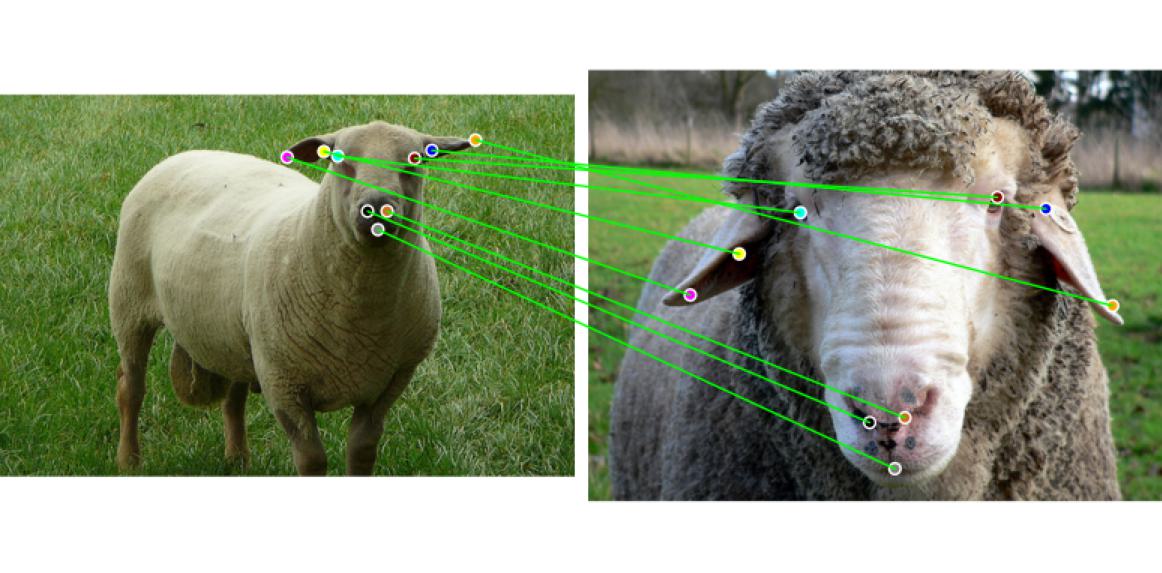} \\
    % TVMonitor row (tvmonitor__12172)
    \includegraphics[width=0.22\textwidth]{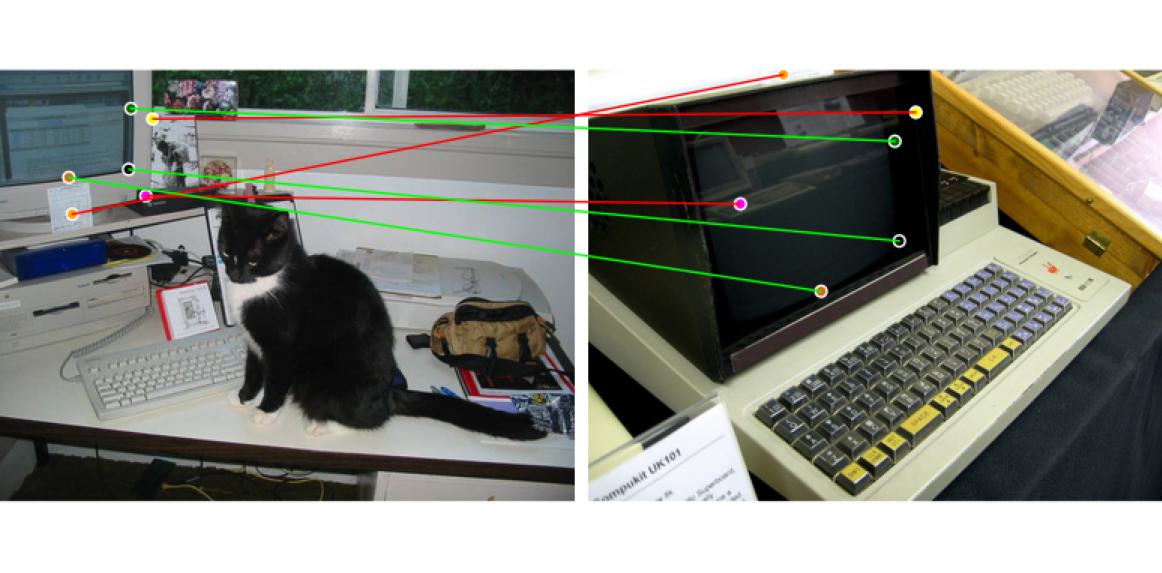} & 
    \includegraphics[width=0.22\textwidth]{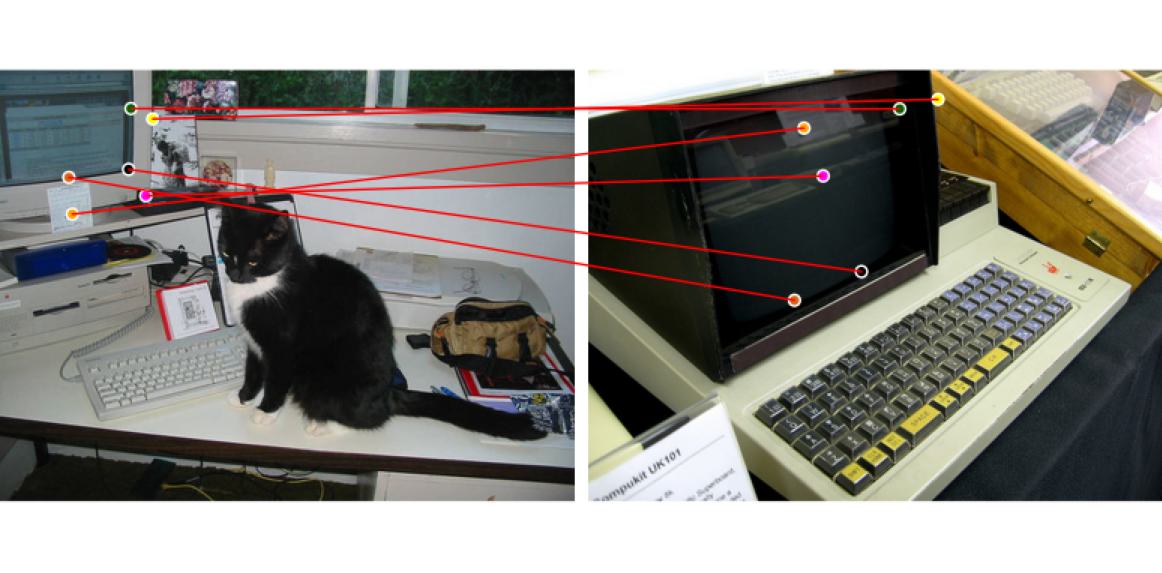} & 
    \includegraphics[width=0.22\textwidth]{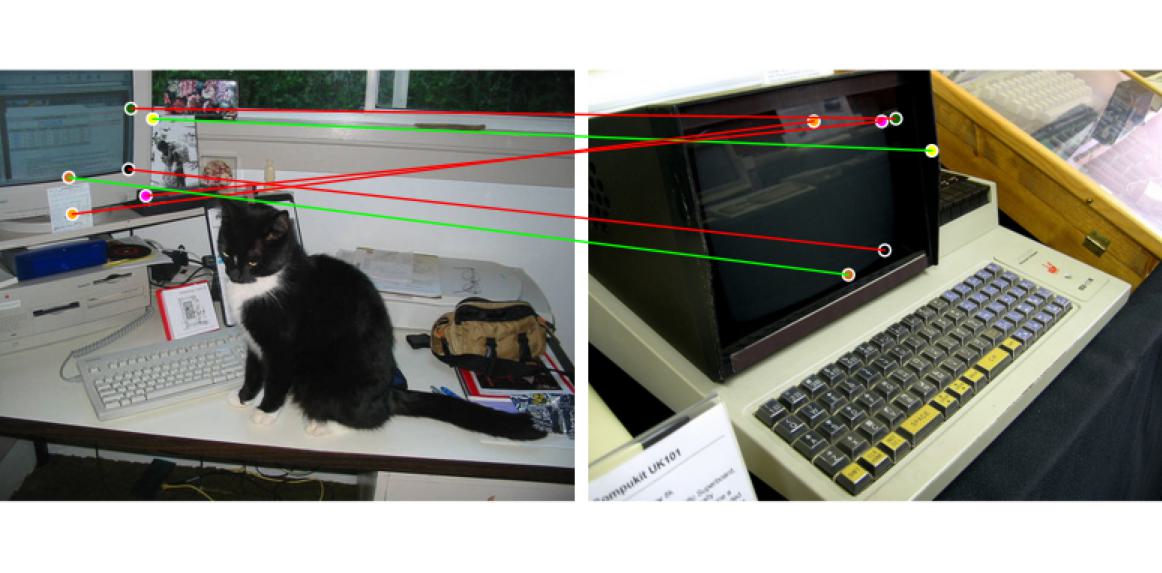} & 
    \includegraphics[width=0.22\textwidth]{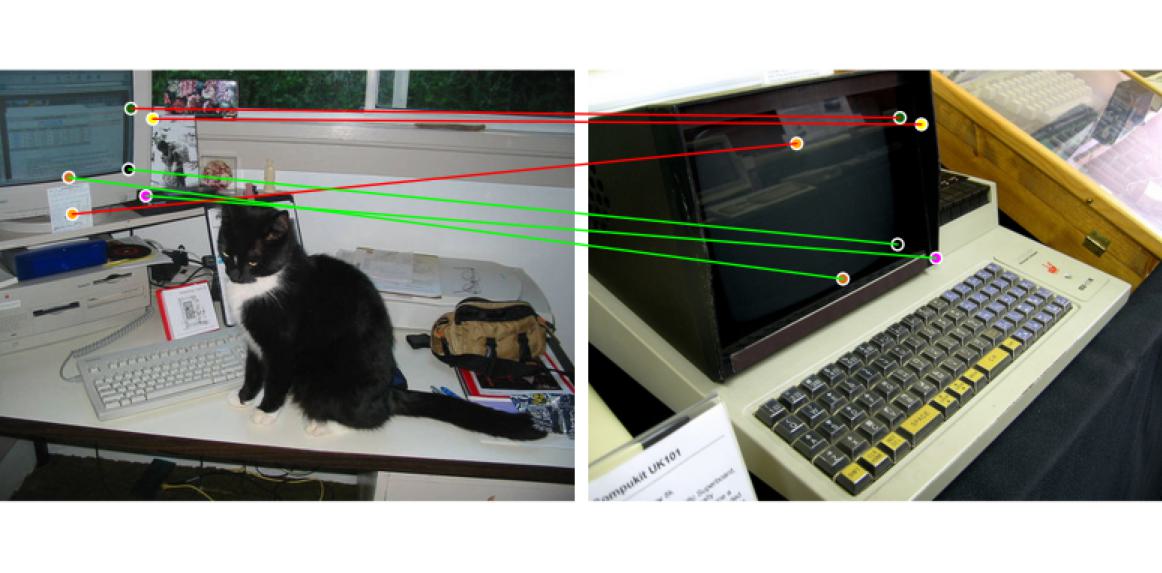} \\
  \end{tabular}
  \caption{Uncurated image pairs of SPair-71K dataset of three SOTA models and ours.}
  \label{fig:uncurated_matches_2}
\end{figure*}